\documentclass[nohyperref]{article}

\usepackage{microtype}
\usepackage{graphicx}
\usepackage{caption}
\usepackage{subcaption}
\usepackage{booktabs} % for professional tables
\usepackage{xcolor}
\usepackage{tikz}
\usepackage{pgfplots}
\usepackage{amsmath}
\usepackage{chemarrow}
\usepackage{rotating}
\usepackage{anyfontsize}

\usetikzlibrary{positioning}
\usetikzlibrary{shapes.geometric}
\usetikzlibrary{arrows.meta}
\usetikzlibrary{shapes.misc}
\pgfplotsset{width=\columnwidth, height=4cm, compat=1.9}

\usepackage{hyperref}

\usepackage[accepted]{icml2023}

\usepackage{amsmath}
\usepackage{amssymb}
\usepackage{mathtools}
\usepackage{amsthm}

\DeclareMathOperator*{\argmin}{arg\,min}

\usepackage[capitalize,noabbrev]{cleveref}

\theoremstyle{plain}

\theoremstyle{definition}

\theoremstyle{remark}

\usepackage[textsize=tiny]{todonotes}

\icmltitlerunning{TR0N: Translator Networks for 0-Shot Plug-and-Play Conditional Generation}

\begin{document}

\twocolumn[
\icmltitle{{\color{violet} TR0N}: {\color{violet}Tr}anslator Networks for {\color{violet}0}-Shot Plug-and-Play Conditional Generatio{\color{violet}n}}



\icmlsetsymbol{equal}{*}

\begin{icmlauthorlist}
\icmlauthor{Zhaoyan Liu}{equal,l6,uoft}
\icmlauthor{No\"{e}l Vouitsis}{equal,l6}
\icmlauthor{Satya Krishna Gorti}{l6}
\icmlauthor{Jimmy Ba}{uoft,vec}
\icmlauthor{Gabriel Loaiza-Ganem}{l6}
\end{icmlauthorlist}

\icmlaffiliation{uoft}{University of Toronto, Toronto, Canada}
\icmlaffiliation{l6}{Layer 6 AI, Toronto, Canada}
\icmlaffiliation{vec}{Vector Institute, Toronto, Canada}

\icmlcorrespondingauthor{Zhaoyan Liu}{zhaoyan@layer6.ai}
\icmlcorrespondingauthor{No\"{e}l Vouitsis}{noel@layer6.ai}
\icmlcorrespondingauthor{Satya Krishna Gorti}{satya@layer6.ai}
\icmlcorrespondingauthor{Jimmy Ba}{jba@cs.toronto.edu}
\icmlcorrespondingauthor{Gabriel Loaiza-Ganem}{gabriel@layer6.ai}

\icmlkeywords{Machine Learning, ICML}

\vskip 0.3in
]



\printAffiliationsAndNotice{\icmlEqualContribution} % otherwise use the standard text.

\begin{abstract}
We propose TR0N, a highly general framework to turn pre-trained unconditional generative models, such as GANs and VAEs, into conditional models. The conditioning can be highly arbitrary, and requires only a pre-trained auxiliary model. For example, we show how to turn unconditional models into class-conditional ones with the help of a classifier, and also into text-to-image models by leveraging CLIP. TR0N learns a lightweight stochastic mapping which ``translates'' between the space of conditions and the latent space of the generative model, in such a way that the generated latent corresponds to a data sample satisfying the desired condition. The translated latent samples are then further improved upon through Langevin dynamics, enabling us to obtain higher-quality data samples. TR0N requires no training data nor fine-tuning, 
yet can achieve a zero-shot FID of $10.9$ on MS-COCO, outperforming competing alternatives not only on this metric, but also in sampling speed -- all while retaining a much higher level of generality. 
% and we empirically verify that it outperforms competing alternatives -- both in FID score and sampling speed -- while retaining a much higher level of generality.
Our code is available at \url{https://github.com/layer6ai-labs/tr0n}.
\end{abstract}

\section{Introduction}
Large machine learning models have recently achieved remarkable success across various tasks \citep{brown2020language, jia2021scaling, nichol2021glide, chowdhery2022palm, rombach2022high, yu2022scaling, ramesh2022hierarchical, saharia2022photorealistic, reed2022a}. 
Nonetheless, training such models requires massive computational resources. 
Properly and efficiently leveraging existing large pre-trained models is thus of paramount importance. 
Yet, tractably combining the capabilities of these models in a plug-and-play manner remains a generally open problem. 
Mechanisms to achieve this task should ideally be modular and model-agnostic, such that one can easily swap out a model component for one of its counterparts (e.g.\ interchanging a GAN \citep{goodfellow2014generative} for a VAE \citep{kingma2013auto, rezende2014stochastic}, or swapping CLIP \citep{radford2021learning} for a new state-of-the-art text/image model). 

\begin{figure} [t!]
\centering
\fontsize{7.5}{9}
\selectfont
\begin{tabular}
{p{0.28\linewidth}p{0.28\linewidth}p{0.28\linewidth}}
   \includegraphics[width=2.45cm]{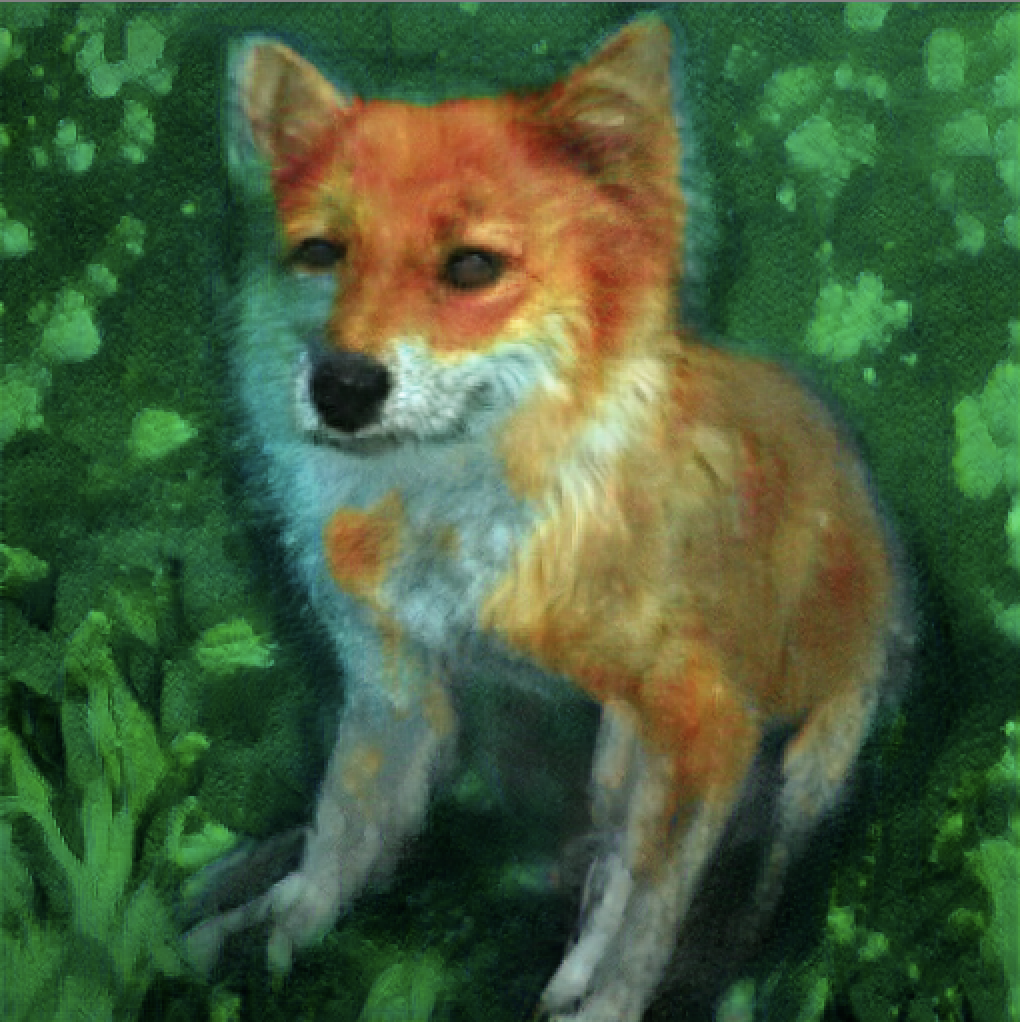} & \includegraphics[width=2.45cm]{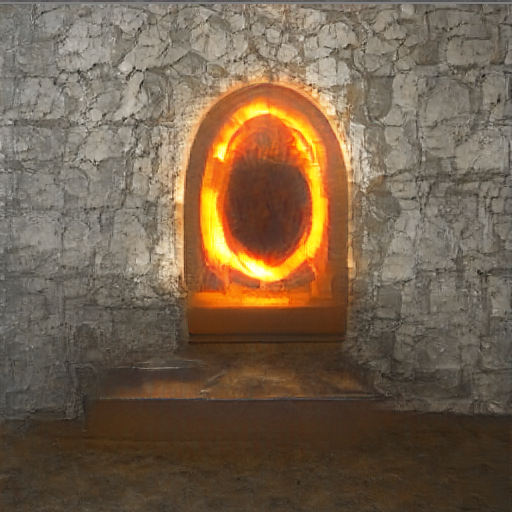}
   & \includegraphics[width=2.45cm]{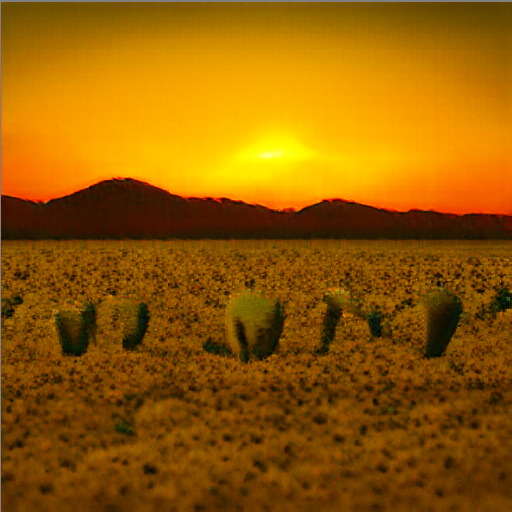}\\
    A painting of a fox in impressionist style & A photo of a flaming portal to an ancient place rendered in unreal engine &   A photo of a sunset over a desert landscape with sand dunes and cacti\\
   \includegraphics[width=2.45cm]{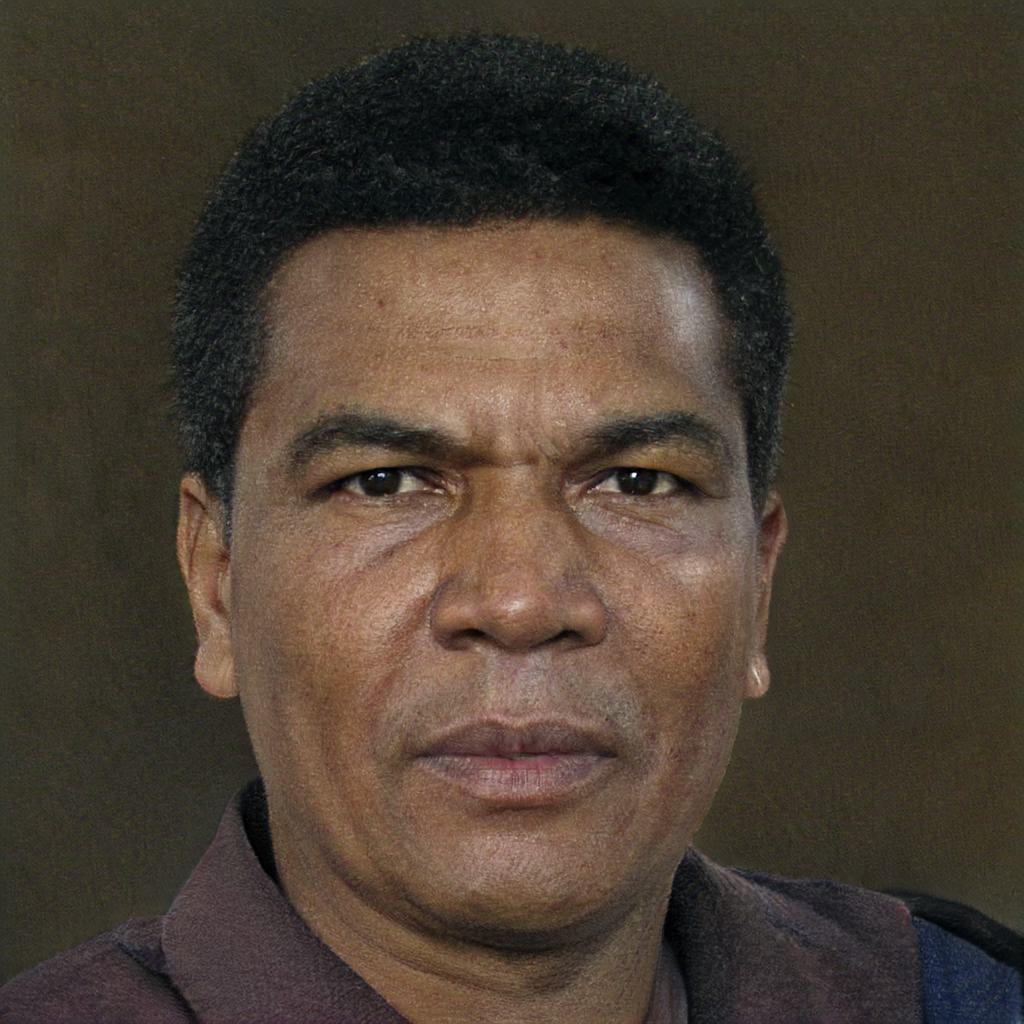} & 
   \includegraphics[width=2.45cm]{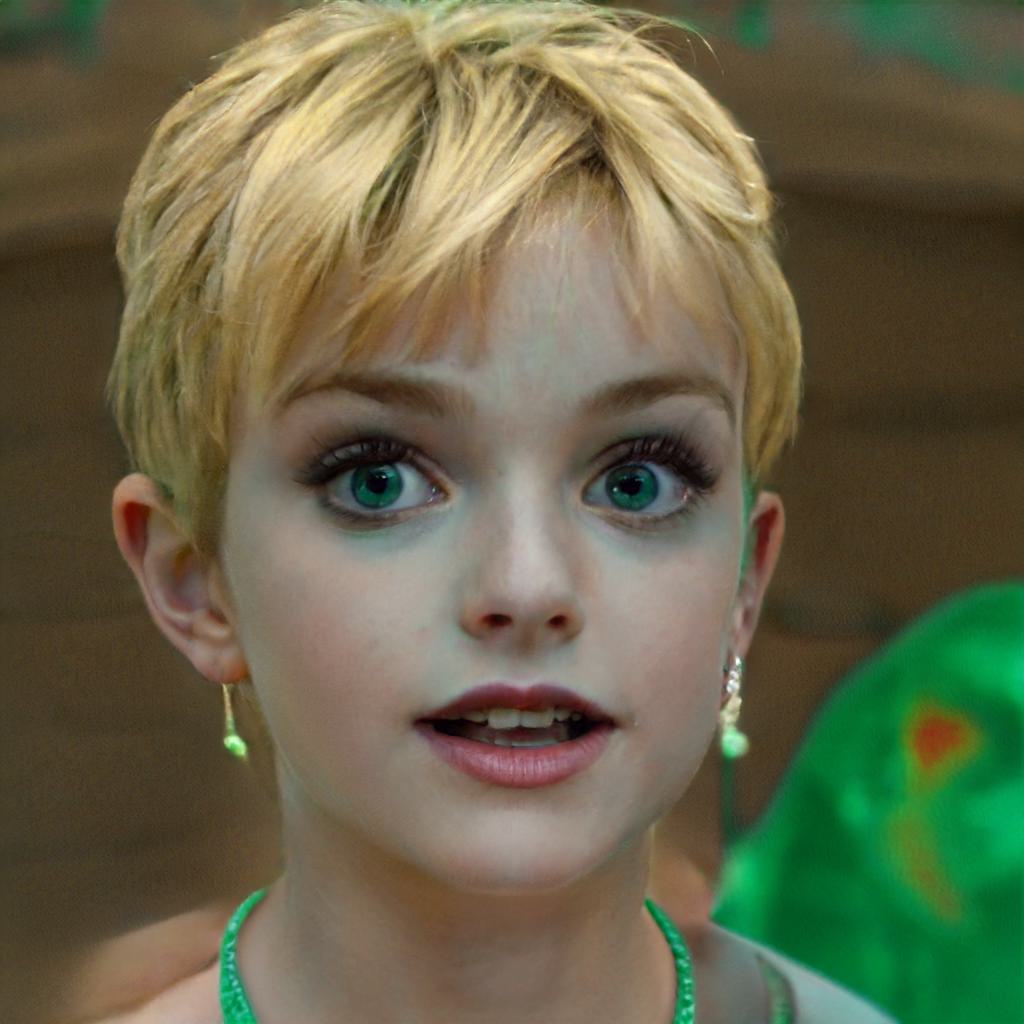}
   &\includegraphics[width=2.45cm]{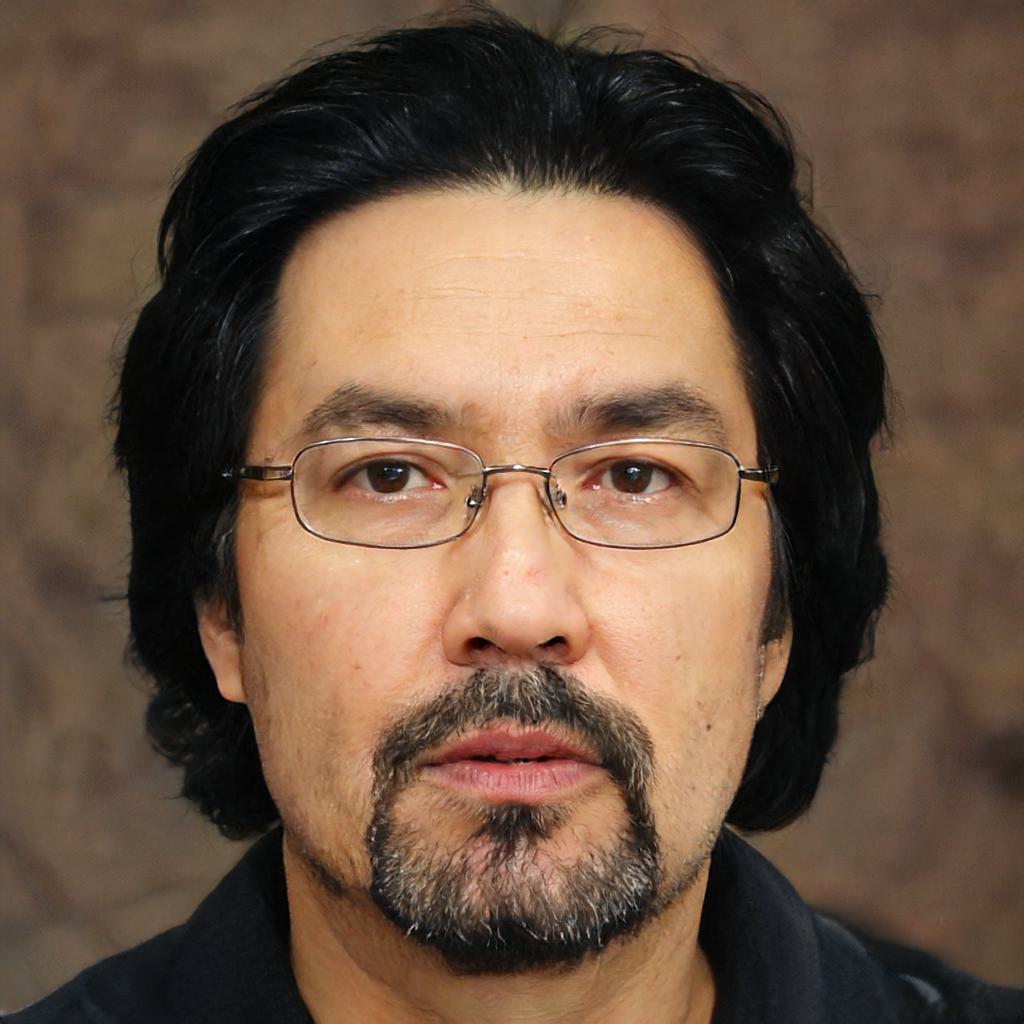}\\
  \hspace{3ex}Muhammad Ali &   \hspace{5ex}Tinker Bell & A man with glasses, long black hair with sideburns and a goatee
\end{tabular}
\vspace{-10pt}
\caption{Images generated by TR0N from corresponding text captions, obtained by finding adequate points on the latent space of a pre-trained GAN. Neither fine-tuning nor training data are used. \textbf{Top row}: BigGAN pre-trained on ImageNet. \textbf{Bottom row}: StyleGAN2 pre-trained on FFHQ.}
\label{fig:fig1}
\vspace{-10pt}
\end{figure}

\begin{figure*}[th!]
    \centering
    \hspace{-0.75cm}
    \begin{subfigure}[b]{0.18\textwidth}
        \centering
        \resizebox{!}{5cm}{
            \begin{tikzpicture}[
blue/.style={rectangle, rounded corners=2, draw=blue!60, fill=blue!5, very thick, minimum size=5mm},
red/.style={trapezium, rounded corners=2, trapezium angle=60, draw=red!60, fill=red!5, very thick, minimum size=5mm},
red_inv/.style={trapezium, rounded corners=2, trapezium angle=-60, draw=red!60, fill=red!5, very thick, minimum size=5mm},
font=\large]

%Nodes
\node[anchor=north] (latent) {$z \sim p(z)$};
\node[red_inv] (generator) [above=of latent] {\color{red}$G$};
\node[anchor=center] (data) [above=0.2cm of generator] {$x$};
\node[red] (aux) [above=of generator] {\color{red}$f$};
\node[anchor=center] (condition) [above=0.2cm of aux] {$c$};
\node[blue] (translator) [above=of aux] {\color{blue}Translator};
\node[anchor=south] (loss) [above=of translator] {Loss$(z)$};

%Lines
\draw[-{Stealth[scale=1.1]}] (latent.north) -- (generator.south);
\draw[-{Stealth[scale=1.1]}] ([xshift=-0.7em]latent.north) to [out=135,in=215] ([xshift=-0.7em]loss.south);
\draw[-] (generator.north) -- (data.south);
\draw[-{Stealth[scale=1.1]}] (data.north) -- (aux.south);
\draw[-] (aux.north) -- (condition.south);
\draw[-{Stealth[scale=1.1]}] (condition.north) -- (translator.south);
\draw[-{Stealth[scale=1.1]}] (translator.north) -- (loss.south);
\draw[dashed, -{Stealth[scale=1.1]}] ([xshift=0.7em]loss.south) -- ([xshift=0.7em]translator.north);

\end{tikzpicture}
        }
    \end{subfigure}
    \hspace{-0.5cm}
    \begin{subfigure}[b]{0.18\textwidth}
        \centering
        \resizebox{!}{5cm}{
            \begin{tikzpicture}[
blue/.style={rectangle, rounded corners=2, draw=blue!60, fill=blue!5, ultra thick, minimum size=5mm},
red/.style={rectangle, rounded corners=2, draw=red!60, fill=red!5, ultra thick, minimum size=5mm},
font=\Large]
\makeatletter
\newcommand{\xdashrightarrow}[2][]{\ext@arrow 0359\rightarrowfill@@{#1}{#2}}
\def\rightarrowfill@@{\arrowfill@@\relax\relbar\chemarrow}
\def\arrowfill@@#1#2#3#4{%
  $\m@th\thickmuskip0mu\medmuskip\thickmuskip\thinmuskip\thickmuskip
   \relax#4#1
   \xleaders\hbox{$#4#2$}\hfill
   #3$%
}
\makeatother

%Nodes
\matrix [draw] at (0,0) (legend) {
  \node [red,label={[label distance=0.35cm]right: Frozen weights}] {}; \\
  \\[0.075cm]
  \node [blue,label={[label distance=0.35cm]right: Trainable weights}] {}; \\
  \\[0.075cm]
  \node [label=right:Gradient update] {\Large$\xdashrightarrow{\hspace{0.75cm}}$}; \\
};

\node[anchor=center] (tron_left) [above=2.5cm of legend] {\LARGE \hspace{-15ex} {\textbf{TR0N}}};
\node[anchor=center] (tron_right) [above=2.5cm of legend] {\LARGE \hspace{15ex} {\textbf{TR0N}}};
\node[anchor=center] (title_left) [above=1.75cm of legend] {\LARGE \hspace{-15ex} {\textbf{training}}};
\node[anchor=center] (title_right) [above=1.75cm of legend] {\LARGE \hspace{15ex} {\textbf{sampling}}};

% \matrix [draw] at (-5.5, 2.75) {
%   \node [red,label={[label distance=0.32cm]right: Frozen}] {}; \\
%   \\[0.075cm]
%   \node [blue,label={[label distance=0.32cm]right: Trainable}] {}; \\
%   \\[0.075cm]
%   \node [label={[label distance=0.4cm]right:Latent}] {$z$}; \\
%   \\[0.075cm]
%   \node [label={[label distance=0.18cm]right:Prior}] {$p(z)$}; \\
%   \\[0.075cm]
%   \node [label={[label distance=0.35cm]right:Generator/decoder model}] {$G$}; \\
%   \\[0.075cm]
%   \node [label={[label distance=0.4cm]right:Data}] {$x$}; \\
%   \\[0.075cm]
%   \node [label={[label distance=0.38cm]right:Auxiliary model}] {$f$}; \\
%   \\[0.075cm]
%    \node [label={[label distance=0.4cm]right:Condition}] {$c$}; \\
%   \\[0.075cm]
%   \node [label={[label distance=0.18cm]right:Translator loss}] {Loss}; \\
%   \\[0.075cm]
%   \node [label={[label distance=0.38cm]right:Discrepancy measure}] {$U$}; \\
%   \\[0.075cm]
%   \node [label=right:Gradient update] {$\xdashrightarrow{\hspace{0.75cm}}$}; \\
%   \\[0.075cm]
%   \node [label={[label distance=0.38cm]right:Gradient update steps}] {$T$}; \\
% };

% Lines
\draw[-] ([yshift=-3.5cm]legend.south) -- ([yshift=-0.25cm]legend.south);
\draw[-] ([yshift=3.5cm]legend.north) -- ([yshift=0.25cm]legend.north);
\end{tikzpicture}
        }
    \end{subfigure}
    \hspace{0.5cm}
    \begin{subfigure}[b]{0.62\textwidth}
        \centering
        \resizebox{!}{5cm}{
            \input{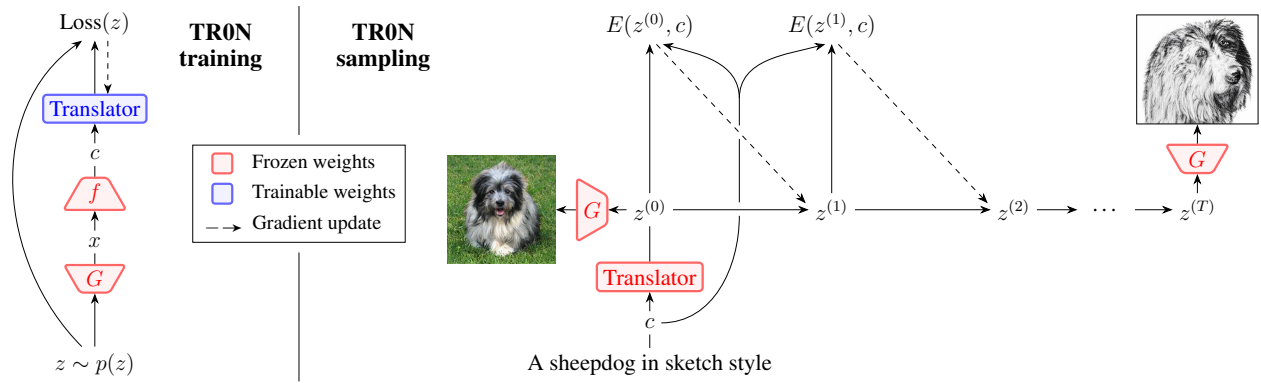}
        }
    \end{subfigure}   
    \caption{\textbf{Left panel}: The stochastic translator network learns to recover $z$ from $c=f(G(z))$. \textbf{Right panel}: The (stochastic) output $z^{(0)}$ of the trained translator -- which is such that $G(z^{(0)})$ ``roughly satisfies'' condition $c$ -- initializes Langevin dynamics over $E(z,c)$ which further improves the sample so as to better match $c$. In the depicted example, $G$ is a GAN trained on ImageNet, $f$ the CLIP image encoder, $c$ the CLIP text embedding corresponding to the given text prompt, and $E(z,c)$ the negative cosine similarity between $f(G(z))$ and $c$.}
    % \caption{\textbf{Left panel}: given $c=f(G(z))$, the  translator network is trained to recover $z$. \textbf{Right panel}: once the translator is trained, its output $z^{(0)}$ -- which is such that $G(z^{(0)})$ ``roughly satisfies'' condition $c$ -- is used to initialize Langevin dynamics over $E(z,c)$, which further improves the sample so as to better match $c$. In the depicted example, $G$ is a GAN trained on ImageNet, $f$ the CLIP image encoder, $c$ the CLIP text embedding corresponding to the given text prompt, and $E(z,c)$ the negative cosine similarity between $f(G(z))$ and $c$.}
    \label{fig:tron}
\end{figure*}

In this work, we study conditional generation through the lens of combining pre-trained models. 
Conditional generative models aim to learn a conditional distribution of data given some conditioning variable $c$. 
They are typically trained from scratch on pairs of data with corresponding $c$ (e.g.\ images $x$, with corresponding class labels or text prompts fed through a language model $c$) \citep{mirza2014conditional, sohn2015learning}. Our goal is to take an arbitrary pre-trained unconditional pushforward generative model \citep{salmona2022can, ross2022neural} --  i.e.\ a model $G$ which transforms latent variables $z$ sampled from a prior $p(z)$ to data samples $x = G(z)$ -- and turn it into a conditional model. To this end, we propose TR0N, a highly general framework to make pre-trained unconditional generative models conditional. TR0N assumes access to a pre-trained auxiliary model $f$ that maps each data point $x$ to its corresponding condition $c=f(x)$, e.g.\ $f$ could be a classifier, or a CLIP encoder. TR0N also assumes access to a function $E(z, c)$ such that latents $z$ for which $G(z)$ ``better satisfies'' a condition $c$ are assigned smaller values. Using this function, for a given $c$, TR0N performs $T$ steps of gradient minimization of $E(z,c)$ over $z$ to find latents that, after applying $G$, will generate desired conditional data samples.

However, we show that na\"ively initializing the optimization of $E$ is highly suboptimal. With this in mind, TR0N starts by learning a network that we use to better initialize the optimization process. We refer to this network as the translator network since it ``translates" from a condition $c$ to a corresponding latent $z$ such that $E(z,c)$ is small, essentially amortizing the optimization problem. Importantly, the translator network is trained \emph{without fine-tuning $G$ or $f$ nor using a provided dataset}. In this sense, TR0N is a zero-shot method wherein the only trainable component is a lightweight translator network. Importantly, TR0N avoids the highly expensive training of a conditional model from scratch and is model-agnostic: we can use any $G$ and any $f$, which also makes it straightforward to update any of these components whenever a newer state-of-the-art version becomes available. We outline the procedure to train the translator network on the left panel of \autoref{fig:tron}.

Once the translator network is trained, we use its output to initialize the optimization of $E$. This reclaims any performance lost due to the amortization gap \citep{cremer2018inference,kim2018semi}, resulting in better local optima and faster convergence than na\"ive initialization. In reality, TR0N is a stochastic method: the translator network is a conditional distribution $q_\theta(z|c)$ that assigns high density to latents $z$ such that $E(z,c)$ is small, and we add noise during the gradient optimization of $E$, which allows us to interpret TR0N as sampling with Langevin dynamics \citep{welling2011bayesian} using an efficient initialization scheme. We exemplify how to sample with TR0N on the right panel of \autoref{fig:tron}.

Our contributions are: $(i)$ introducing translator networks and a particularly efficient parameterization of them, allowing for various ways to initialize Langevin dynamics; $(ii)$ framing TR0N as a highly general framework, whereas previous related works mostly focus on a single task with specific choices of $G$ and $f$; and $(iii)$ showing that TR0N empirically outperforms competing alternatives across tasks in image quality and computational tractability, while producing diverse samples; and that it can achieve an FID \citep{Heusel2017GANsTB} of $10.9$ on MS-COCO \citep{lin2014microsoft}.
% both in image quality -- achieving an FID \citep{Heusel2017GANsTB} of $10.9$ on MS-COCO \citep{lin2014microsoft} -- and computational tractability, while producing diverse samples.

\section{Background}
\paragraph{Joint text/image models} In this work, we leverage pre-trained joint text/image models as a particular choice for both the auxiliary model $f$ and to construct $E$, enabling TR0N to be conditioned on either free-form text prompts or on image semantics. 
Recent joint text/image models such as CLIP learn a joint representation space $\mathcal{C}_\text{CLIP}$ for images and texts. 
CLIP includes an image encoder $f^{\text{img}}: \mathcal{X} \rightarrow \mathcal{C}_\text{CLIP}$ and a text encoder $f^{\text{txt}}:\mathcal{T} \rightarrow \mathcal{C}_\text{CLIP}$, where $\mathcal{X}$ is the space of images and $\mathcal{T}$ is the space of text prompts, which are trained in such a way that images and texts that are semantically aligned are mapped to similar representations. 
More specifically, CLIP is such that the negative cosine similarity $U_{\text{sim}}(f^{\text{img}}(x), f^{\text{txt}}(y))$ is small for semantically aligned image/text pairs $(x, y) \in \mathcal{X} \times \mathcal{T}$, and large for semantically unaligned pairs, where $U_{\text{sim}}(c',c) = -c^\top c' / (\Vert c'\Vert_2 \Vert c \Vert_2)$.
% This is usually achieved by minimizing the negative cosine similarity $U_{\text{sim}}(f^{\text{img}}(x), f^{\text{txt}}(y))$ for semantically aligned image/text pairs $(x, y) \in \mathcal{X} \times \mathcal{T}$, and maximizing it for semantically unaligned pairs, where $U_{\text{sim}}(c',c) = -c^\top c' / (\Vert c'\Vert_2 \Vert c \Vert_2)$.

\paragraph{Pushforward models} We use the term \emph{pushforward model} to refer to any generative model whose samples $x \in \mathcal{X}$ can be obtained as $x=G(z)$, where $z \in \mathcal{Z}$ is a latent variable sampled from some (typically not trainable) prior $p(z)$, and $G : \mathcal{Z} \rightarrow \mathcal{X}$ is a neural network. 
Many models fall into this category, including generative adversarial networks (GANs), variational autoencoders (VAEs), normalizing flows \citep{dinh2016density, durkan2019neural} and variants thereof \citep{brehmer2020flows, caterini2021rectangular, ross2021tractable}, and more \citep{tolstikhin2017wasserstein,loaiza2022diagnosing}. We focus on GANs and VAEs since they use a low-dimensional latent space $\mathcal{Z}$, which will later make the translator network's task easier. Our main goal is to turn a pre-trained unconditional pushforward model $(p(z), G)$ into a conditional model $(p(z|c), G)$.

\paragraph{EBMs and Langevin dynamics} We will later formalize the goal of TR0N as sampling from a distribution $p(z|c)$ defined only up to proportionality, i.e.\ $p(z|c) \propto e^{-\beta E(z,c)}$, where $E:\mathcal{Z} \times \mathcal{C} \rightarrow \mathbb{R}$ is called the energy function, and the hyperparameter $\beta > 0$ controls the degree to which small values of $E(z,c)$ correspond to large values of $p(z|c)$, and vice-versa. We hereafter refer to this formulation as an energy-based model (EBM). While the energy function in EBMs is typically learnable \citep{xie2016theory,du2019implicit}, in our work we define and fix an energy function that allows us to enforce the requirement that ``applying $G$ to a sample from $p(z|c)$ satisfies condition $c$''. 
Langevin dynamics is a method that allows us to sample from EBMs by constructing a Markov chain $(z^{(0)}, z^{(1)}, \dots)$ given by
\begin{equation}\label{eq:langevin_dynamics}
    z^{(t+1)} = z^{(t)} - \dfrac{\beta \lambda^{(t)}}{2} \nabla_z E\left(z^{(t)}, c\right) + \sqrt{\lambda^{(t)}}\epsilon^{(t)},
\end{equation}
where the sequence $(\lambda^{(0)}, \lambda^{(1)}, \dots)$ is a hyperparameter, and $\epsilon^{(t)} \sim \mathcal{N}(\epsilon; 0, I)$. Under mild conditions and by sending $\lambda^{(t)}$ to $0$ at an appropriate rate, 
the limiting distribution of this Markov chain as $t\to\infty$ is $p(z|c)$. 
% the stationary distribution of this Markov chain (i.e.\ its limiting distribution as $t\to\infty$) is $p(z|c)$. 
Langevin dynamics can be interpreted as gradient descent on $E$ with added noise, and has been successfully applied to sample and train deep EBMs, where in practice it is common to deviate from theory and set $\lambda^{(t)} = \lambda > 0$ for all $t$ (i.e.\ a single scalar hyperparameter $\lambda$ is used) for improved empirical performance. 
Also, while in theory convergence does not depend upon the starting point $z^{(0)}$, in practice this choice can greatly speed up convergence \citep{hinton2002training, nijkamp2020anatomy, yoon2021autoencoding}, just as with gradient descent \citep{boyd2004convex, glorot2010understanding}.

\section{TR0N}
\subsection{Plug-and-play components of TR0N}

TR0N requires three key components to ensure that it can operate as a plug-and-play framework. First, TR0N takes an arbitrary pre-trained pushforward model $(p(z), G)$. 
 % that it aims to turn into a conditional one $(p(z|c), G)$. 
 TR0N also assumes access to a pre-trained auxiliary model $f: \mathcal{X} \rightarrow \mathcal{C}$ that maps data to its corresponding condition. For example, if our goal is to condition on class labels, $f$ would be a classifier, and $\mathcal{C}$ the space of probability vectors of appropriate length. If we aim to condition on text, $f$ could be given by the CLIP image encoder $f^{\text{img}}$ -- although we will see later that a different choice of $f$ led us to improved empirical performance in this setting -- and $\mathcal{C}$ the latent space of CLIP, $\mathcal{C}_\text{CLIP}$.  
The final component of TR0N is a function $E: \mathcal{Z} \times \mathcal{C} \rightarrow \mathbb{R}$ which measures how much $G(z)$ satisfies condition $c$, an intuitive choice being 
\begin{equation}\label{eq:score}
    E(z,c) = U(f(G(z)), c),
\end{equation}
where $U: \mathcal{C} \times \mathcal{C} \rightarrow \mathbb{R}$ measures discrepancy between conditions, for example: when $f$ is a classifier, $U$ could be the categorical cross entropy; and when $f$ is the image encoder from CLIP, $U$ could be the negative cosine similarity, $U_{\text{sim}}$. However, other choices of $E$ are possible, as we will show in our experiments.

\subsection{Overview of TR0N}

\paragraph{Translator networks} TR0N uses the aforementioned components to train the translator network which, given $c$, aims to output a $z$ with small $E(z, c)$. This can be intuitively understood as amortizing the minimization of $E$ with a neural network so as to not have to run a minimizer from scratch for every $c$. Since there can be many latents $z$ for which $G(z)$ satisfies $c$ (i.e. $E(z, c)$ is small), we propose to have the translator be a distribution $q_\theta(z|c)$, parameterized by $\theta$. This way, the translator can assign high density to all the latents $z$ such that $E(z,c)$ is small. We will detail how we instantiate $q_\theta(z|c)$ with a neural network in \autoref{sec:gmm}, but highlight that any choice of conditional density is valid. Importantly, since we have access to the unconditional model $(p(z), G)$, we can generate synthetic data $G(z)$ with $z \sim p(z)$; and since we have access to $f$, we can obtain the condition corresponding to $G(z)$, namely $c=f(G(z))$. Together, this means that the translator can be trained through maximum likelihood \emph{without the need for a provided training dataset}, through
\begin{equation}\label{eq:ml}
\theta^* = \argmin_\theta \mathbb{E}_{p(z)}\left[-\log q_\theta \left(z|c=f(G(z))\right)\right].
\end{equation}
We summarize the above objective in Algorithm \ref{alg:training}.

\begin{algorithm}[t]
   \caption{TR0N training}
   \label{alg:training}
\begin{algorithmic}
   \STATE {\bfseries Input:} $p(z)$, $G$, $f$, $q_\theta(z|c)$, and batch size $B$
   \WHILE{not converged}
   \STATE Sample $z_i \sim p(z)$ for $i=1,\dots,B$
   \STATE $c_i \leftarrow f(G(z_i))$ for $i=1,\dots,B$
   \STATE $\Delta \leftarrow \nabla_\theta \dfrac{1}{B}\displaystyle \sum_{i=1}^B -\log q_\theta(z_i|c_i)$
   \STATE Use $\Delta$ to update $\theta$, e.g.\ with ADAM \citep{kingma2015adam}
   \ENDWHILE
\end{algorithmic}
\end{algorithm}

\paragraph{Error correction} The translator is trained to stochastically recover $z$ from $c=f(G(z))$, so that intuitively it places high densities on latents which have low $E(z, c)$ values. Yet, the translator is not directly trained to minimize $E$, and thus having an error correction step, over which $E$ is explicitly optimized, is beneficial to further improve its output. Thus, for a given $c$, we run $T$ steps of gradient descent on $E(z, c)$ over $z$, which we initialize with the help of the translator. Initializing optimization with $q_{\theta^*}(z|c)$ rather than na\"ively (e.g\ Gaussian noise) significantly speeds up convergence, and as we will see in our experiments, can also lead to better local optima. Importantly, we can use the translator in various ways to initialize optimization. For example, we can sample $M$ times from $q_{\theta^*}(z|c)$, and use the sample with the lowest $E(z, c)$ value (which would be impossible if the translator was deterministic). We will detail another way to leverage the translator network to initialize optimization in \autoref{sec:gmm}. In practice, we add Gaussian noise to gradient descent. Together with the stochasticity of the translator, this ensures diverse samples. Lastly, we transform the final latent, $z^{(T)}$, through $G$ to obtain a conditional sample from TR0N. We summarize this procedure in Algorithm \ref{alg:sampling}.
% which we will shortly formalize as sampling from a conditional pushforward model.

\subsection{TR0N as an EBM sampler}

TR0N can be formalized as sampling from an EBM with Langevin dynamics. Defining the distribution $p(z|c)$, which we call the \emph{conditional prior}, as $p(z|c) \propto e^{-\beta E(z, c)}$, Algorithm \ref{alg:sampling} uses Langevin dynamics \eqref{eq:langevin_dynamics} to sample from $p(z|c)$, initialized with the help of $q_{\theta^*}(z|c)$. Thus, TR0N can be interpreted as a sampling algorithm for the conditional pushforward model $(p(z|c), G)$. Again, $G$ remains fixed throughout, and conditioning is achieved only through the prior $p(z|c)$. 
In this view, the translator network $q_{\theta^*}(z|c)$ can be understood as a rough approximation to $p(z|c)$, as both of these distributions assign large densities to latents $z$ for which $E(z, c)$ is small. This is precisely why the translator provides a good initialization for Langevin dynamics: the more $z^{(0)}$ ``comes from $p(z|c)$'', the faster \eqref{eq:langevin_dynamics} will converge.

\paragraph{Why maximum-likelihood?} If our goal is for the translator to be close to the conditional prior, i.e.\ $q_{\theta^*}(z|c) \approx p(z|c)$, then a natural question is why train the translator through \eqref{eq:ml}, which does not involve $p(z|c)$, rather than by minimizing some discrepancy between these two distributions? The answer is that, since the target $p(z|c)$ is specified only up to proportionality and true samples from it are not readily available (better sampling from $p(z|c)$ is in fact what we designed TR0N to achieve), minimizing commonly-used discrepancies such as the KL divergence or the Wasserstein distance is not tractable. The only discrepancy we are aware of that could be used in this setting is the Stein discrepancy, which has also been used to train EBMs \citep{grathwohl2020learning}. However, in preliminary experiments we observed very poor results by attempting to minimize this discrepancy. In contrast, the maximum-likelihood objective \eqref{eq:ml} is straightforward to optimize, and obtained strong empirical performance in our experiments.

\begin{algorithm}[t]
   \caption{TR0N sampling}
   \label{alg:sampling}
\begin{algorithmic}
   \STATE {\bfseries Input:} $c$, $q_{\theta^*}(z|c)$, $M$, $E$, $T$, $\beta$, $\lambda$, $G$
   \STATE {\bfseries Output:} $G(z^{(T)})$
   \STATE Initialize $z^{(0)}$ with $q_{\theta^*}(z|c)$, e.g.\ $z_m \sim q_{\theta^*}(z|c)$ for $m=1,\dots, M$, and $z^{(0)} = \displaystyle \argmin_{z_m} E(z_m, c)$
   \FOR{$t=0$ {\bfseries to} $T-1$}
   \STATE Sample $\epsilon^{(t)} \sim \mathcal{N}(\epsilon; 0, I)$
   \STATE $z^{(t+1)} = z^{(t)} - \dfrac{\beta \lambda}{2} \nabla_z E \left(z^{(t)}, c\right) + \sqrt{\lambda}\epsilon^{(t)}$
   \ENDFOR
\end{algorithmic}
\end{algorithm}

\subsection{GMMs to parameterize translator networks}\label{sec:gmm}

While clearly any choice of conditional density model $q_\theta(z|c)$ can be used in TR0N, we choose a Gaussian mixture model (GMM), as it has several advantages that we will discuss shortly. More specifically, we use a neural network, parameterized by $\eta$, which maps conditions $c \in \mathcal{C}$ to the mean $(\mu_{\eta, k}(c))_{k=1}^K \in \mathcal{Z}^K$ and weight $w_\eta(c) \in \mathbb{R}^K$ parameters of a Gaussian mixture, i.e.\
\begin{equation}
    q_\theta(z|c) = \sum_{k=1}^K w_{\eta, k}(c)\mathcal{N}(z; \mu_{\eta, k}(c), \text{diag}(\sigma^2)),
\end{equation}
where $w_\eta (c)$ has positive entries which add up to one (enforced with a softmax), and $\theta = (\eta, \sigma)$, i.e.\ $\sigma$ is learnable. 
We use a simple multilayer perceptron with multiple heads to parameterize this neural network.

Our GMM choice for the stochastic translator has four important benefits: $(i)$ It is a very lightweight model, and thus achieves our goal of being much more tractable to train than any of the pre-trained components $G$ and $f$, which we once again highlight remain fixed throughout. $(ii)$ Sampling from a GMM is very straightforward and can be done very quickly. $(iii)$ Empirically, we found that using more complicated density models $q_\theta(z|c)$ such as normalizing flows did not result in improved performance. We hypothesize that, since Langevin dynamics acts as an error correction step, $q_{\theta^*}(z|c)$ just needs to approximate, rather than perfectly recover, $p(z|c)$. $(iv)$ Finally, taking $q_\theta(z|c)$ as a GMM allows using the translator to initialize Langevin dynamics in ways that are not straightforward to extend to a non-GMM setting. In particular, we found that sometimes (when diversity is not as paramount), rather than initializing \eqref{eq:langevin_dynamics} as described in Algorithm \ref{alg:sampling}, better performance could be achieved by directly using the GMM parameters. That is, we initialize at the GMM mean, $z^{(0)} = \sum_k w_{\eta^*, k}(c) \mu_{\eta^*, k} (c)$. Note that the mean of more complex distributions might not be so easily computable. Further, we found that when initializing this way, optimizing the weights and means directly yielded better performance, i.e\ we write $z^{(t)}$ as $z^{(t)} = \sum_k w_k^{(t)} \mu_k^{(t)}$, and perform Langevin dynamics as
\begin{align}\label{eq:langevin_params}
&(w^{(t+1)}, \mu^{(t+1)}) = \\
& (w^{(t)}, \mu^{(t)}) -\dfrac{\beta \lambda}{2} \nabla_{(w,\mu)}E\left(z^{(t)}, c\right) + \sqrt{\lambda}\epsilon^{(t)},\nonumber
\end{align}
where $w^{(0)} = w_{\eta^*}(c)$, $\mu_k^{(0)} = \mu_{\eta^*, k}(c)$ for $k=1,\dots,K$, and the size of $\epsilon^{(t)}$ is appropriately changed from \eqref{eq:langevin_dynamics}.

\subsection{TR0N for Bayesian inference}\label{sec:bayes}

In some settings, the auxiliary model $f$ might provide a probabilistic model $p(c|x)$. For example, when $f$ is a classifier, $p(c|x)=f_c(x)$.\footnote{We slightly abuse notation here and use $c$ interchangeably as either a one-hot vector, or as the corresponding integer index.}  Combined with the pushforward model, this provides a latent/data/condition joint distribution $p(z,x,c)=p(z)\delta_{G(z)}(x)p(c|x)$, where $\delta_{G(z)}(x)$ denotes a point mass on $x$ at $G(z)$. For Bayesian inference, it might be of interest to sample from the corresponding posterior $p(x|c)$, which is equivalent to sampling from $p(z|c)$ and transforming the result through $G$. That is, in this scenario, the conditional prior $p(z|c)$ is a proper posterior distribution of latents given a condition. TR0N can sample from this posterior by using specific choices of $\beta$ and $E$. While these choices provide a probabilistically principled way of combining $(p(z), G)$ and $f$ into a conditional model, we find that non-Bayesian choices obtain stronger empirical results. We nonetheless believe that TR0N being compatible with Bayesian inference is worth highlighting. Due to space constraints, we include additional details in Appendix \ref{app:bayes}.

\section{Related Work}
\begin{figure*}[t!]
    \centering
    \includegraphics[width=0.2375\linewidth]{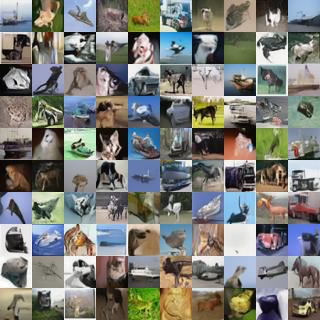} \hspace{0.05cm}
    \includegraphics[width=0.2375\linewidth]{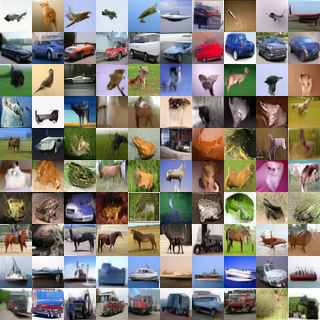}
    \hspace{0.05cm}
    \includegraphics[width=0.2375\linewidth]{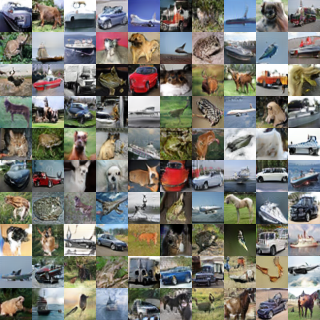}
    \hspace{0.05cm}
    \includegraphics[width=0.2375\linewidth]{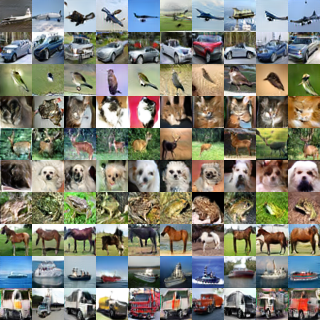}
    \caption{Samples from NVAE (\textbf{first panel}), TR0N:NVAE+ResNet50 (\textbf{second panel}), AutoGAN (\textbf{third panel}), and TR0N:AutoGAN+ResNet50 (\textbf{fourth panel}). Rows on the second and fourth panels correspond to classes: TR0N  learns to diversely sample in a class-conditional way, while retaining the image quality of the underlying unconditional model. Best viewed while zoomed-in.}
    \label{fig:cifar_conditioning}
\end{figure*}

Several methods aim to obtain a conditional generative model by combining pre-trained models, although none of them shares all of the advantages of TR0N. 
Notably, almost all the works we discuss below are shown to work for a single task, unlike TR0N which is widely applicable.

\paragraph{Non-zero-shot methods} \citet{Zhou2021TowardsLT} and \citet{Wang2022CLIPGENLT} leverage CLIP to train text-to-image models without text data, but unlike TR0N, still require a training dataset of images and relatively longer training times. \citet{wang2022traditional} propose a method to turn a classifier into a conditional generative model which also requires training data to train a masked autoencoder. \citet{nie2021controllable} condition GANs through a similar EBM as us, but use data to train $f$, do not condition on text, and do not use translator networks. \citet{zhang2023adding} add conditioning to pre-trained diffusion models \citep{ho2020denoising}, but require training data to do so.

\paragraph{Deterministic optimization} The works of \citet{nguyen2016synthesizing}, \citet{liu2021fusedream}, \citet{patashnik2021styleclip}, and \citet{li2022composing} can be thought of as deterministic versions of our EBM, where rather than sampling from $p(z|c) \propto e^{-\beta E(z,c)}$, the energy $E(z, c)$ is directly minimized over $z$. These methods do not account for the fact that there can be many latents $z$ such that $G(z)$ satisfies condition $c$, and thus can be less diverse than TR0N. Additionally, these methods do not have a translator network, and with the exception of FuseDream \citep{liu2021fusedream}, na\"ively initialize optimization, resulting in reduced empirical performance and needing more gradient steps for optimization to converge. We also note that FuseDream's initialization scheme -- which we detail in Appendix \ref{app:exp_details} for completeness -- requires many forward passes through $G$ and $f$, and remains much more computationally demanding than TR0N's.

\paragraph{Stochastic methods} \citet{ansari2021refining} apply Langevin dynamics on the latent space of a GAN, but do so to iteratively refine its samples, rather than for conditional sampling. \citet{nguyen2017plug} use a similar EBM to ours, but do not use a translator network and initialize Langevin dynamics na\"{i}vely, once again resulting in significantly decreased empirical performance as compared to TR0N. \citet{promptgen2022} also define a similar EBM to ours, which is approximated with a normalizing flow for each different $c$, meaning that a different model has to be trained for each condition, resulting in a method that is far less scalable than TR0N. Finally, \citet{Pinkney2022clip2latent} propose clip2latent, which can be understood as using a diffusion model instead of a GMM as the translator network, making clip2latent more expensive to train than TR0N. Importantly, they perform no error correction step whatsoever, and thus do not leverage important information contained in the gradient of $E$.

\section{Experiments}
All our experimental details -- including which translator-based initialization we used for each experiment -- are provided in Appendix \ref{app:exp_details}.

\subsection{Conditioning on class labels}

We demonstrate TR0N's ability to make an unconditional model on CIFAR-10 \citep{krizhevsky2009learning} into a class-conditional one. To highlight the flexibile plug-and-play nature of TR0N, we use two different pushforward models $G$: an NVAE \citep{Vahdat2020NVAEAD}, and an AutoGAN \citep{Gong2019AutoGANNA} -- we use this somewhat non-standard choice of GAN since most publicly available GANs pre-trained on CIFAR-10 are class-conditional. 
Here, $\mathcal{C}$ is the space of probability vectors of length $10$, we take $f$ as a ResNet50 classifier \cite{He2015DeepRL}, and use $E$ as in \eqref{eq:score} with $U$ given by the cross-entropy loss, $U_{\text{ent}}(c', c) \coloneqq -\sum_j c_j \log c'_j$.

\autoref{fig:cifar_conditioning} shows qualitative results: we can see that, for both pushforward models, TR0N not only obtains samples from each of the 10 classes, but that it achieves this without sacrificing neither image quality nor diversity. 

We also make quantitative comparisons between each unconditional model (i.e.\ NVAE and AutoGAN) and the resulting conditional models provided by TR0N. To make the comparison equitable, we sample unconditionally from TR0N models by first sampling one of the $10$ classes uniformly at random, and then sampling from the corresponding conditional. Results are shown in \autoref{table:cifar-table}, by measuring image quality and diversity through both the FID score and the inception score \citep{salimans2016improved}, and the quality of conditioning through the average probability that the ResNet50 assigns to the intended class of TR0N samples. TR0N not only makes the models conditional as these probabilities are very close to $1$, especially for the AutoGAN-based model, but it also improves their FID and inception scores (IS): TR0N leverages the classifier $f$ not only to make a conditional model, but also to improve upon its underlying pre-trained pushforward model.

\begin{figure*} [h!]
\hspace*{-0.2cm}
\centering
\fontsize{8.5}{10}
\selectfont
\begin{tabular}
% {cccccccc}
{p{0.145\linewidth}p{0.145\linewidth}p{0.145\linewidth}p{0.145\linewidth}p{0.145\linewidth}p{0.145\linewidth}}
   \includegraphics[width=2.6cm]{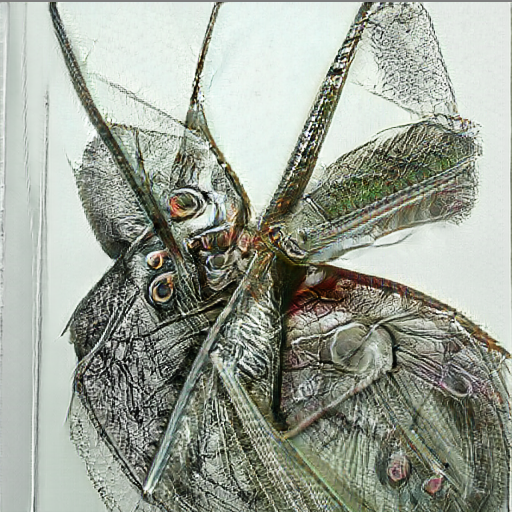} & \hspace{-0.25cm}\includegraphics[width=2.6cm]{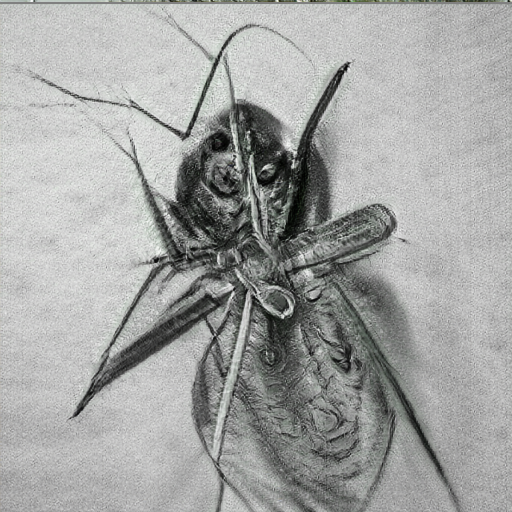} &\includegraphics[width=2.6cm]{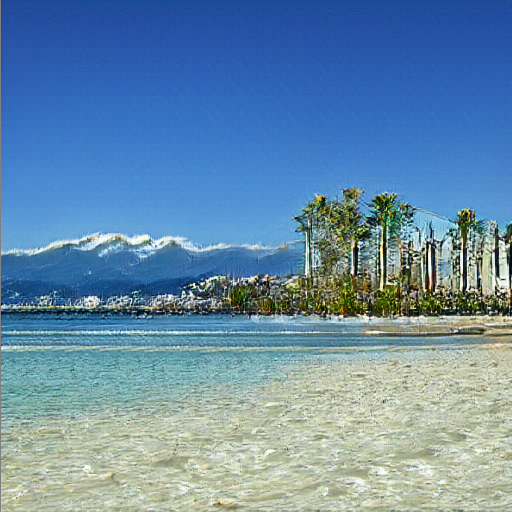} & \hspace{-0.25cm}\includegraphics[width=2.6cm]{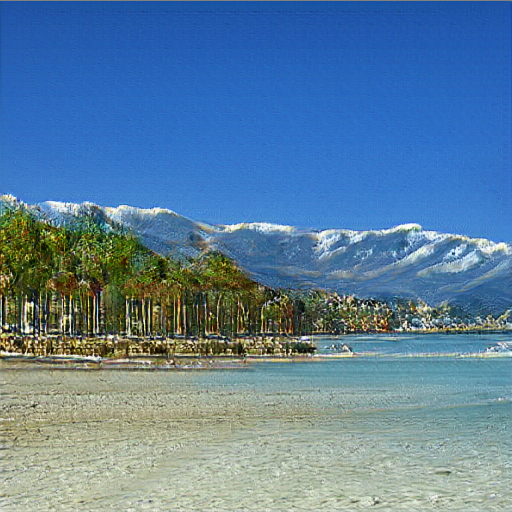} & \includegraphics[width=2.6cm]{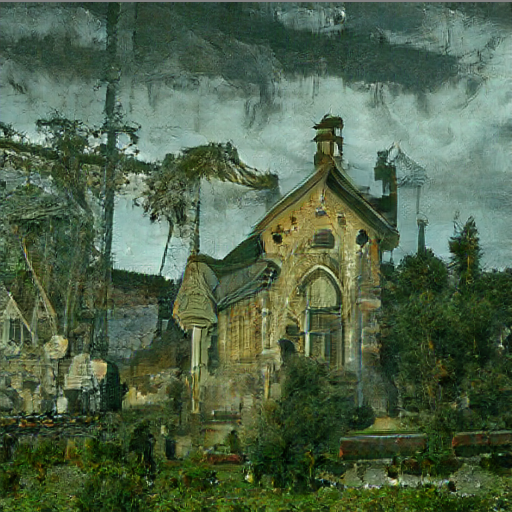}
   & \hspace{-0.25cm}\includegraphics[width=2.6cm]{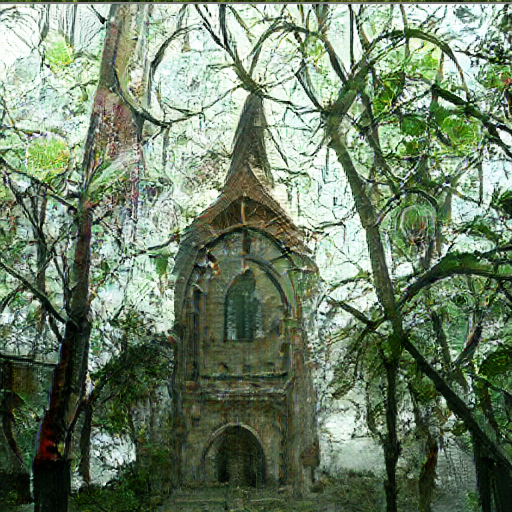}\\[-0.05cm]
   \includegraphics[width=2.6cm]{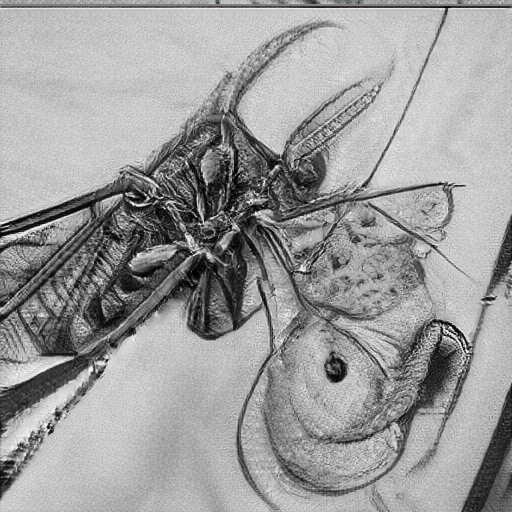} & \hspace{-0.25cm}\includegraphics[width=2.6cm]{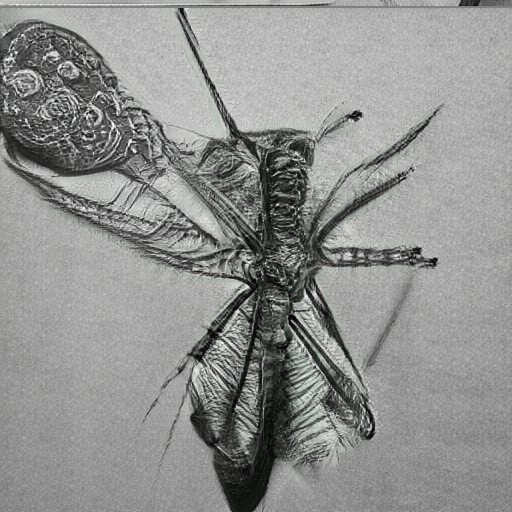} & \includegraphics[width=2.6cm]{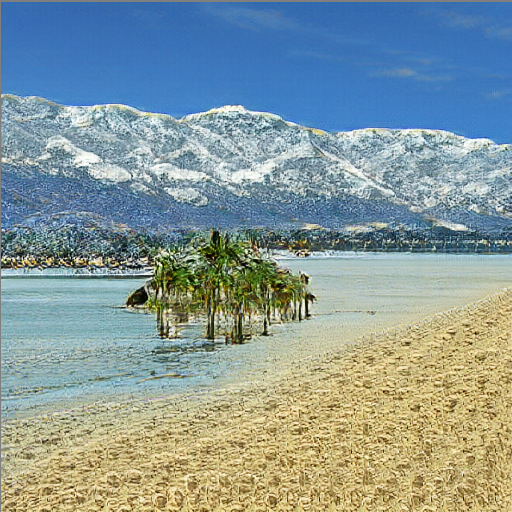} & \hspace{-0.25cm}\includegraphics[width=2.6cm]{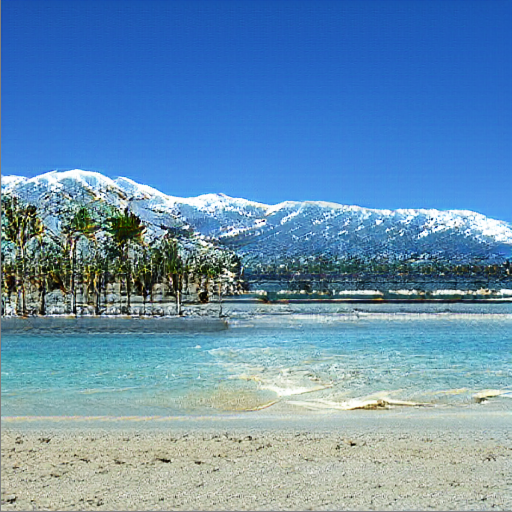} & \includegraphics[width=2.6cm]{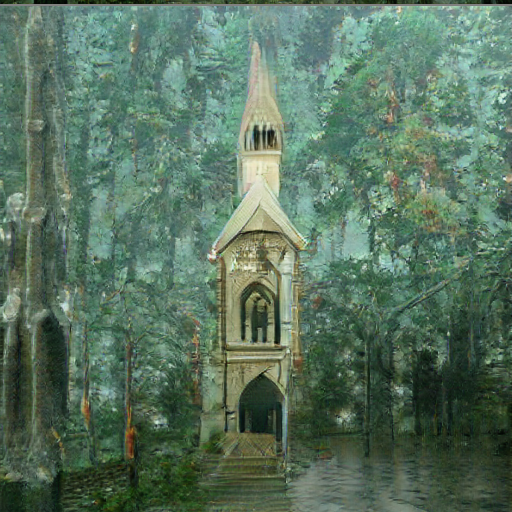}
   & \hspace{-0.25cm}\includegraphics[width=2.6cm]{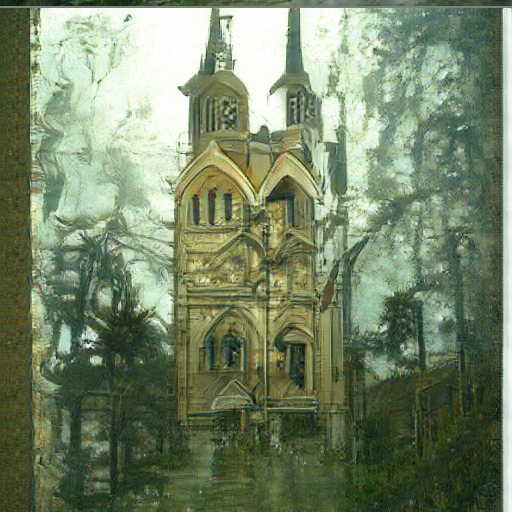} \\
    \multicolumn{2}{p{0.3\linewidth}}{A pencil drawing of an insect, abstract, surrealism} & \multicolumn{2}{p{0.3\linewidth}}{A beach with crystal clear water and palm trees, with snow-capped mountains in the background} &\multicolumn{2}{p{0.3\linewidth}}{A painting of the middle-aged, gothic church surrounded with trees, under the rainy weather} \\

\end{tabular}
\caption{Samples from TR0N:BigGAN+CLIP (BLIP). We can see samples are diverse for all captions.}
\label{fig:TR0N_diversity}
\end{figure*}

\autoref{table:cifar-table} also includes some ablations: $(i)$ removing the error correction (Langevin dynamics) step altogether, which results in heavily degraded FID and IS for the NVAE-based model, and much worse conditioning for both models; $(ii)$ removing the translator, which is equivalent to a stochastic version (i.e.\ with Langevin dynamics instead of gradient descent) of the method of \citet{nguyen2016synthesizing}, and which significantly hurts FID, IS, and conditioning performance, highlighting the relevance of translator networks; $(iii)$ using a deterministic translator rather than a stochastic one (see Appendix \ref{app:exp_details} for details), which significantly hurts FID and IS due to a lack of diversity since Langevin dynamics is always initialized at the same point for a given condition; and $(iv)$ using ADAM instead of gradient descent to update latents in Algorithm \ref{alg:sampling}, which not only removes the formal interpretation of TR0N as an EBM sampler, but also worsens performance across metrics. Finally, we include additional results in Appendix \ref{app:extra_exps} using the Bayesian choice of $\beta$ and $E$ mentioned in \autoref{sec:bayes}.

 \begin{table}[t]
\vskip -0.1in
\caption{FID, IS, and average probability assigned to the intended class of generated samples by a ResNet50 on CIFAR-10. ``no EC'' stands for ``no error correction'', ``no T'' for ``no translator'', ``DT'' for ``deterministic translator'', and ``ADAM'' and for changing the optimizer in Langevin dymanics.}
\begin{center}
\scalebox{0.76}{
\begin{tabular}{lcccr}
\toprule
Model & FID $\downarrow$ & IS $\uparrow$ & Avg. prob. $\uparrow$\\
\midrule
% w/o Augment CE and gamma = 0
NVAE & $41.70$ & $6.95$ & $-$ \\
TR0N:NVAE+ResNet50 & $\mathbf{19.79}$ & $\mathbf{8.64}$ & $\mathbf{0.75}$ \\
% TR0N-NVAE+ResNet50 ($\gamma = 0$) & 19.79 & 0.75 & \\
% TR0N-NVAE+ResNet50 & 19.64 & 0.75 & \\
TR0N:NVAE+ResNet50 (no EC) & $40.80$ & $6.95$ & $0.20$ \\
TR0N:NVAE+ResNet50 (no T) & $36.74$ & $7.75$ & $0.54$ \\
TR0N:NVAE+ResNet50 (DT) & $77.11$ & $7.15$ & $0.48$\\
TR0N:NVAE+ResNet50 (ADAM) & $20.24$ & $8.31$ & $0.51$\\

% with Augment CE
% % TR0N-NVAE+ResNet50 & 18.38 & 0.72 & \\
% TR0N-NVAE+ResNet50 & 19.64 & 0.75 & \\
% TR0N-NVAE+ResNet50 (no EC) & 40.8 & - & \\
% % TR0N-NVAE+ResNet50 (GD) & 18.44 & 0.72 & \\
% % TR0N-NVAE+ResNet50 (no Aug) & 19.64 & 0.75 & \\
% TR0N-NVAE+ResNet50 ($\gamma = 0$) & - & - & \\
% TR0N-NVAE+ResNet50 (no T) & 75.47 & 0.43 & \\
% TR0N-NVAE+ResNet50 (DT) & - & - & \\
\midrule
% w/o Augment CE and gamma = 0
AutoGAN & $12.45$ & $8.53$ & $-$ \\
TR0N:AutoGAN+ResNet50 & $\mathbf{10.69}$ & $\mathbf{8.91}$ & $0.95$\\
% TR0N-AutoGAN+ResNet50 ($\gamma = 0$) & 10.69 & 0.95 & \\
% TR0N-AutoGAN+ResNet50 & 12.08 & 0.95 & \\
TR0N:AutoGAN+ResNet50 (no EC) & $11.00$ & $8.66$ & $0.41$ \\
TR0N:AutoGAN+ResNet50 (no T) & $14.30$ & $8.37$ & $0.68$ \\
TR0N:AutoGAN+ResNet50 (DT) & $123.23$ & $5.48$ & $\mathbf{0.97}$ \\
TR0N:AutoGAN+ResNet50 (ADAM) & $11.08$ & $8.88$ & $0.93$\\

% with Augment CE
% TR0N-AutoGAN+ResNet50 & 12.08 & 0.95 & \\
% % TR0N-AutoGAN+ResNet50 & 12.40 & 0.94 & \\
% TR0N-AutoGAN+ResNet50 (no EC) & 11.0 & 0.41 & \\
% % TR0N-AutoGAN+ResNet50 (GD) & 12.40 & 0.94 & \\
% % TR0N-AutoGAN+ResNet50 (no Aug) & 12.08 & 0.95 & \\
% TR0N-AutoGAN+ResNet50 ($\gamma = 0$) & 10.62 & 0.89 & \\
% TR0N-AutoGAN+ResNet50 (no T) & 14.75 & 0.51 & \\
% TR0N-AutoGAN+ResNet50 (DT) & 149.8 & 0.98 & \\
\bottomrule
\end{tabular}
}
\end{center}
\label{table:cifar-table}
\vskip -0.1in
\end{table}

\subsection{Conditioning on text}

\paragraph{Natural images} We now show TR0N's capability to turn unconditional models into text-to-image models. Here, we use $\mathcal{C}$ as the latent space of CLIP, $\mathcal{C}_\text{CLIP}$, and to condition on a text prompt $y \in \mathcal{T}$, we simply use the text encoder, $c = f^{\text{txt}}(y)$. First, we take $G$ as a BigGAN\footnote{While BigGAN is a class-conditional model, it is not text-conditional. We include the class condition on $\mathcal{Z}$ and think of the GAN as unconditional. See Appendix \ref{app:exp_details} for details.} \citep{brock2018large} pre-trained on ImageNet \citep{deng2009imagenet}, and use two different choices of $f$ leveraging CLIP. The first choice is simply the image encoder of CLIP, $f^{\text{img}}$. We focus our comparisons against FuseDream -- which to the best of our knowledge is the best performing competing method.\footnote{While FuseDream is a deterministic method, its provided implementation (optionally) adds noise during gradient optimization: \url{https://github.com/gnobitab/FuseDream}.}

As our second choice of $f$, we also leverage a pre-trained caption model $h:\mathcal{X} \rightarrow \mathcal{T}$ followed by CLIP's text encoder, i.e.\ $f=f^{\text{txt}} \circ h$, further demonstrating the plug-and-play nature of TR0N. The idea behind this choice is that CLIP's image and text encoders have been shown to not perfectly map images and text to the same regions of $\mathcal{C}_\text{CLIP}$ \citep{liang2022mind}. Adding the caption model $h$ -- which maps images to text descriptions -- allows us to use the text encoder within $f$, i.e.\ the same encoder used to obtain $c$, resulting in better matching latents. This choice of $f$ is a novel empirical contribution for zero-shot text-to-image generation. We use BLIP \citep{li2022blip} for the caption model $h$. For both choices of $f$, we follow FuseDream and use the negative augmented CLIP score $E_{\text{CLIP}}$ as $E$, which is given by $E_{\text{CLIP}}(z, c) \coloneqq \mathbb{E}_{p(\phi)}[U_{\text{sim}}(f^{\text{img}}(\phi[G(z)]), c)]$, where $\phi[x]$ is a differentiable data-augmentation \citep{zhao2020differentiable} of $x$, and $p(\phi)$ a pre-specified distribution over data-augmentations. Like \citet{liu2021fusedream}, we find that using the data augmentations helps avoid adversarial examples with small values of $E(z,c)$ which nonetheless do not satisfy $c$. Note that $E_{\text{CLIP}}$ always uses the image encoder from CLIP, regardless of which $f$ we use to train the translator network. 

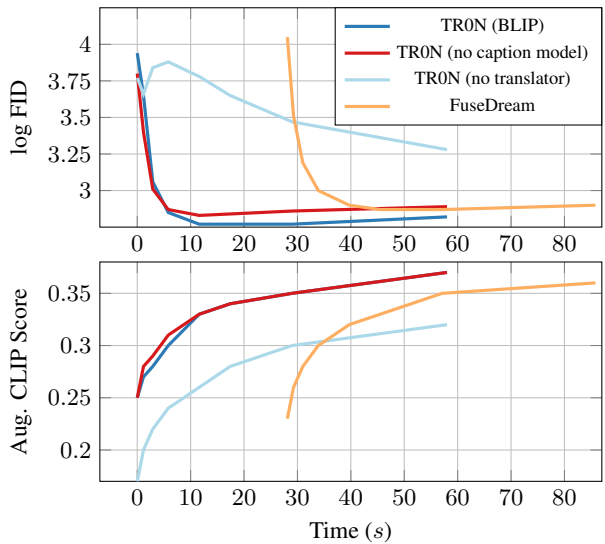
\begin{figure}[t!]
    \centering
    \begin{tikzpicture}
\definecolor{red}{RGB}{215,25,28}
\definecolor{orange}{RGB}{253,174,97}
\definecolor{cyan}{RGB}{171,217,233}
\definecolor{blue}{RGB}{44,123,182}
\footnotesize
\begin{axis}[
    height=4.5cm,
    xlabel near ticks,
    ylabel near ticks,
    ylabel={log FID},
    xmin=-7, xmax=87,
    ymin=2.75, ymax=4.25,
    xtick={0, 10, 20, 30, 40, 50, 60, 70, 80},
    ytick={3, 3.25, 3.5, 3.75, 4},
    legend pos=north east,
    ymajorgrids=true,
    xmajorgrids=true,
    legend style={font=\scriptsize, xshift=0.205cm, yshift=0.095cm},
    every axis plot/.append style={very thick}
]

% %TR0N w caption
% \addplot[
%     color=blue,
%     ]
%     coordinates {
%     (0,3.94)(2,3.65)(5,3.06)(10,2.85)(20,2.77)(30,2.77)(50,2.77)(100,2.82)
%     };
% %TR0N without caption
% \addplot[
%     color=red,
%     ]
%     coordinates {
%     (0,3.80)(2,3.40)(5,3.01)(10,2.87)(20,2.83)(30,2.84)(50,2.86)(100,2.89)
%     };
% %TR0N no translator
% \addplot[
%     color=cyan,
%     ]
%     coordinates {
%     (0,3.77)(2,3.66)(5,3.84)(10,3.88)(20,3.78)(30,3.65)(50,3.47)(100, 3.28)
%     };
% %FuseDream
% \addplot[
%     color=orange,
%     ]
%     coordinates {
%     (40, 4.05)(42,3.51)(45,3.19)(50,3.00)(60,2.90)(70,2.87)(90,2.87)(140,2.90)
%     };

% \legend{TR0N (w/ caption), TR0N (w/o caption), TRON (no T), FuseDream}

%TR0N w caption
\addplot[
    color=blue,
    ]
    coordinates {
    (0.01,3.94)(1.17,3.65)(2.91,3.06)(5.81,2.85)(11.61,2.77)(17.41,2.77)(29.01,2.77)(58.01,2.82)
    };
%TR0N without caption
\addplot[
    color=red,
    ]
    coordinates {
    (0.01,3.80)(1.17,3.40)(2.91,3.01)(5.81,2.87)(11.61,2.83)(17.41,2.84)(29.01,2.86)(58.01,2.89)
    };
%TR0N no translator
\addplot[
    color=cyan,
    ]
    coordinates {
    (0.01,3.77)(1.17,3.66)(2.91,3.84)(5.81,3.88)(11.61,3.78)(17.41,3.65)(29.01,3.47)(58.01, 3.28)
    };
%FuseDream
\addplot[
    color=orange,
    ]
    coordinates {
    (28.11, 4.05)(29.27,3.51)(31.01,3.19)(33.91,3.00)(39.71,2.90)(45.51,2.87)(57.11,2.87)(85.81,2.90)
    };

\legend{TR0N (BLIP), TR0N (no caption model), TR0N (no translator), FuseDream}

% \addplot[
%     color=blue,
%     ]
%     coordinates {
%     (0,51.3)(2,38.3)(5,21.4)(10,17.35)(20,16.00)(30,15.97)(50,16.03)(100,16.72)
%     };
% \addplot[
%     color=red,
%     ]
%     coordinates {
%     (0,44.6)(2,29.9)(5,20.32)(10,17.62)(20,17.02)(30,17.12)(50,17.38)(100,18.00)
%     };
% \addplot[
%     color=orange,
%     ]
%     coordinates {
%     (40, 57.6)(42,33.52)(45,24.3)(50,20.16)(60,18.1)(70,17.58)(90,17.6)(140,18.12)
%     };
% \addplot[
%     color=cyan,
%     ]
%     coordinates {
%     (0,43.21)(2,38.85)(5,46.51)(10,48.49)(20,43.68)(30,38.61)(50,32.07)(100, 26.67)
%     };

\end{axis}
\end{tikzpicture}\\
    \begin{tikzpicture}
\definecolor{red}{RGB}{215,25,28}
\definecolor{orange}{RGB}{253,174,97}
\definecolor{cyan}{RGB}{171,217,233}
\definecolor{blue}{RGB}{44,123,182}
\footnotesize
\begin{axis}[
    height=4.5cm,
    xlabel near ticks,
    xlabel={Time ($s$)},
    ylabel near ticks,
    ylabel={Aug. CLIP Score},
    xmin=-7, xmax=87,
    ymin=0.17, ymax=0.38,
    xtick={0, 10, 20, 30, 40, 50, 60, 70, 80},
    ytick={0.20, 0.25, 0.30, 0.35},
    legend pos=south east,
    ymajorgrids=true,
    xmajorgrids=true,
    every axis plot/.append style={very thick}
]

%FuseDream
% \addplot[
%     color=orange,
%     ]
%     coordinates {
%     (28.11, 4.05)(29.27,3.51)(31.01,3.19)(33.91,3.00)(39.71,2.90)(45.51,2.87)(57.11,2.87)(85.81,2.90)
%TR0N w caption
\addplot[
    color=blue,
    ]
    coordinates {
    (0.01,0.25)(1.17,0.27)(2.91,0.28)(5.81,0.30)(11.61,0.33)(17.41,0.34)(29.01,0.35)(58.01,0.37)
    };
%TR0N without caption
\addplot[
    color=red,
    ]
    coordinates {
    (0.01,0.25)(1.17,0.28)(2.91,0.29)(5.81,0.31)(11.61,0.33)(17.41,0.34)(29.01,0.35)(58.01,0.37)
    };
%TR0N no translator
\addplot[
    color=cyan,
    ]
    coordinates {
    (0.01,0.17)(1.17,0.20)(2.91,0.22)(5.81,0.24)(11.61,0.26)(17.41,0.28)(29.01,0.30)(58.01,0.32)
    };
%FuseDream
\addplot[
color=orange,
]
coordinates {
(28.11, 0.23)(29.27, 0.26)(31.01,0.28)(33.91,0.30)(39.71,0.32)(45.51,0.33)(57.11,0.35)(85.81,0.36)
};
% \legend{TR0N (w/ caption), TR0N (w/o caption), FuseDream, TRON (no T)}

\end{axis}
\end{tikzpicture}
    \\[-0.3cm]
    \caption{Comparisons between TR0N:BigGAN+CLIP and FuseDream on MS-COCO as a function of time required to generate a sample. \textbf{Top panel}: FID score (in log scale), lower is better. \textbf{Bottom panel}: augmented CLIP score, higher is better.}
    \label{fig:fid_vs_iters}
    \vskip -0.1in
\end{figure}

\begin{figure*} [h!]
\newcommand{\centered}[1]{\begin{tabular}{c} #1 \\[1.75cm] \end{tabular}}
\centering
\fontsize{8.5}{10}
\selectfont
\begin{tabular}
{p{0.135\linewidth}p{0.135\linewidth}p{0.135\linewidth}p{0.135\linewidth}p{0.135\linewidth}p{0.135\linewidth}}
    \hspace{2.5ex}0.32 $\pm$ 0.001 & \hspace{2.5ex}0.25 $\pm$ 0.000 & \hspace{2.5ex}0.34 $\pm$ 0.001 & \hspace{2.5ex}0.33 $\pm$ 0.001 & \hspace{2.5ex}0.34 $\pm$ 0.001 & \hspace{2.5ex}0.26 $\pm$ 0.001 \\
    \includegraphics[width=2.25cm]{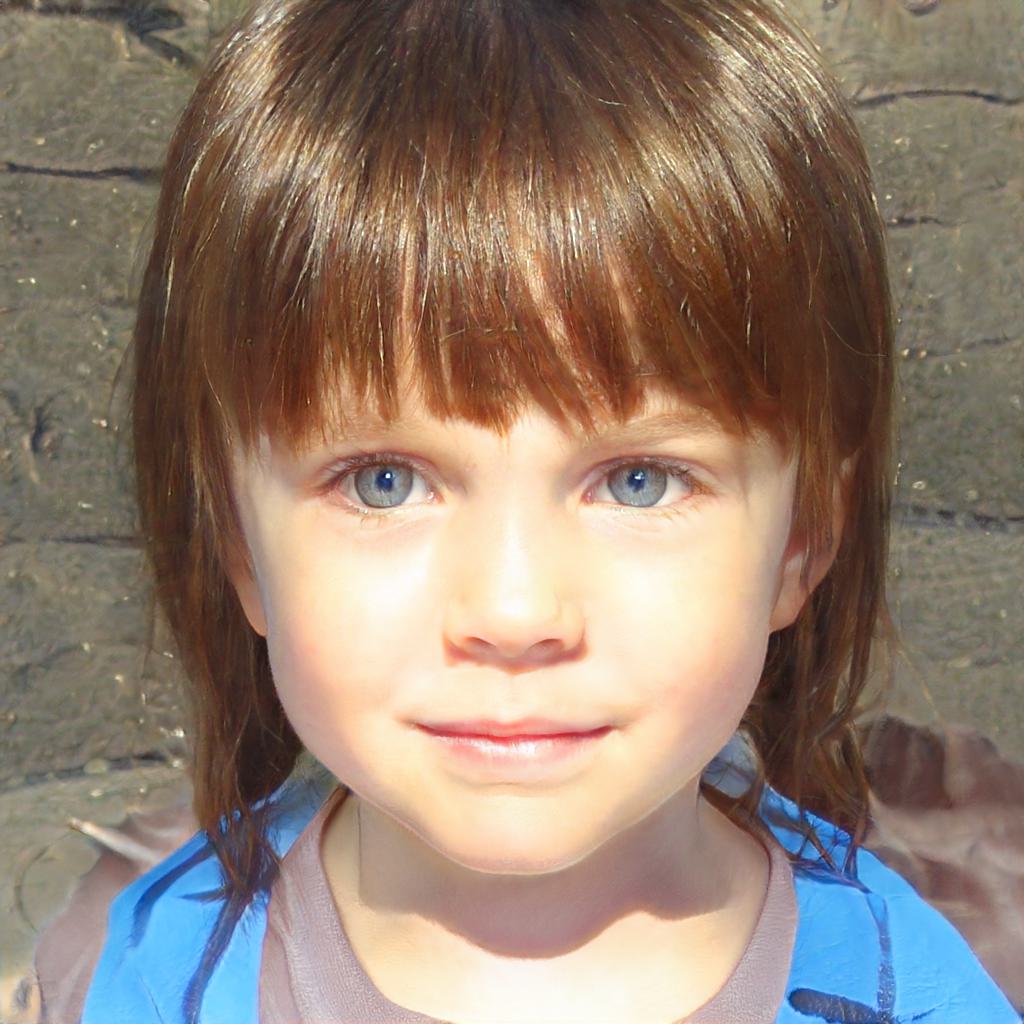} & \includegraphics[width=2.25cm]{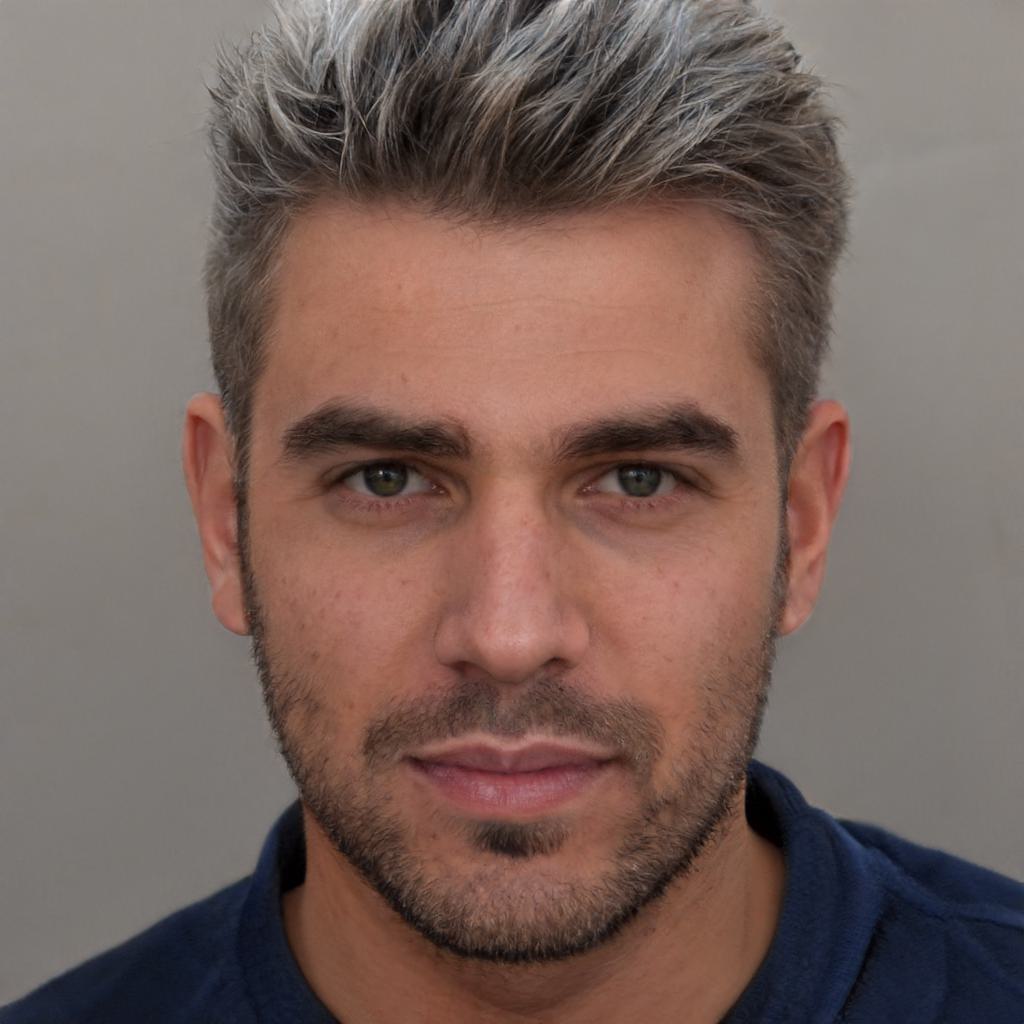} & \includegraphics[width=2.25cm]{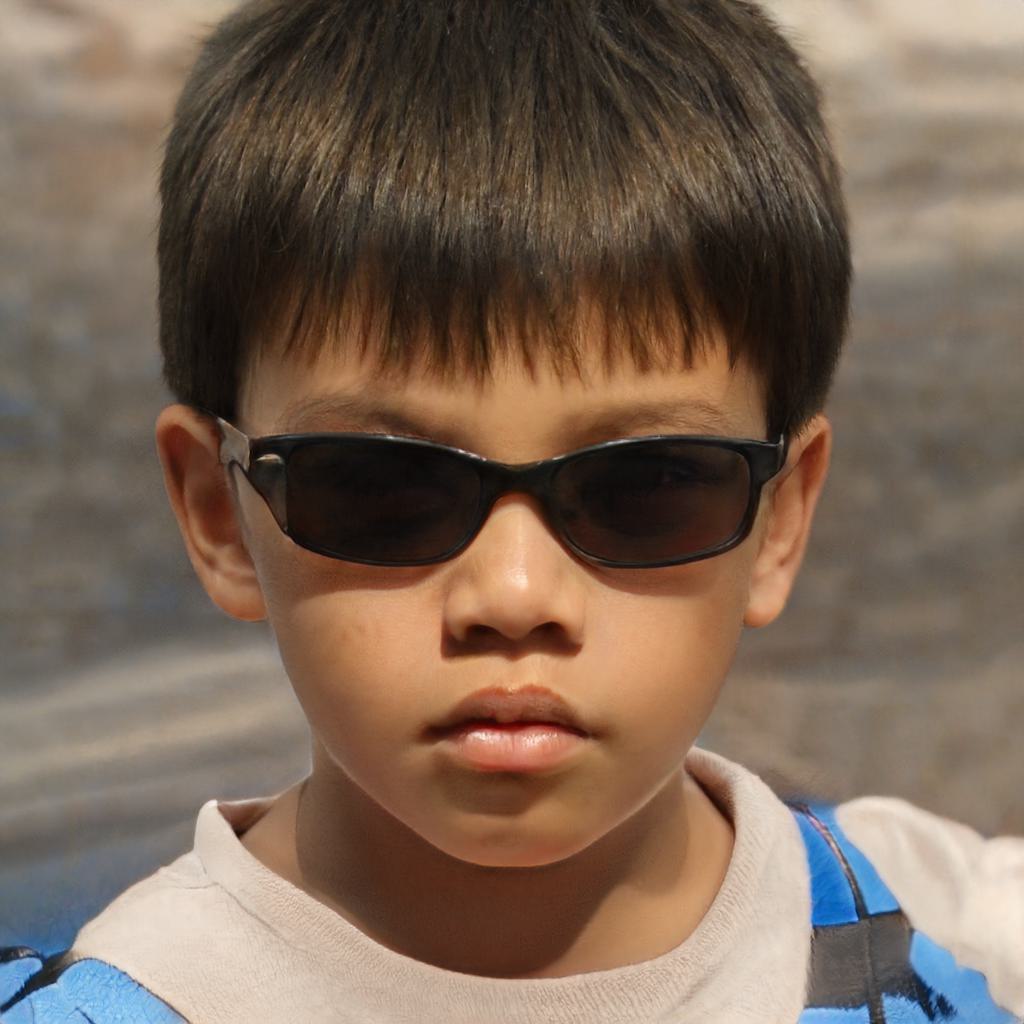} & \includegraphics[width=2.25cm]{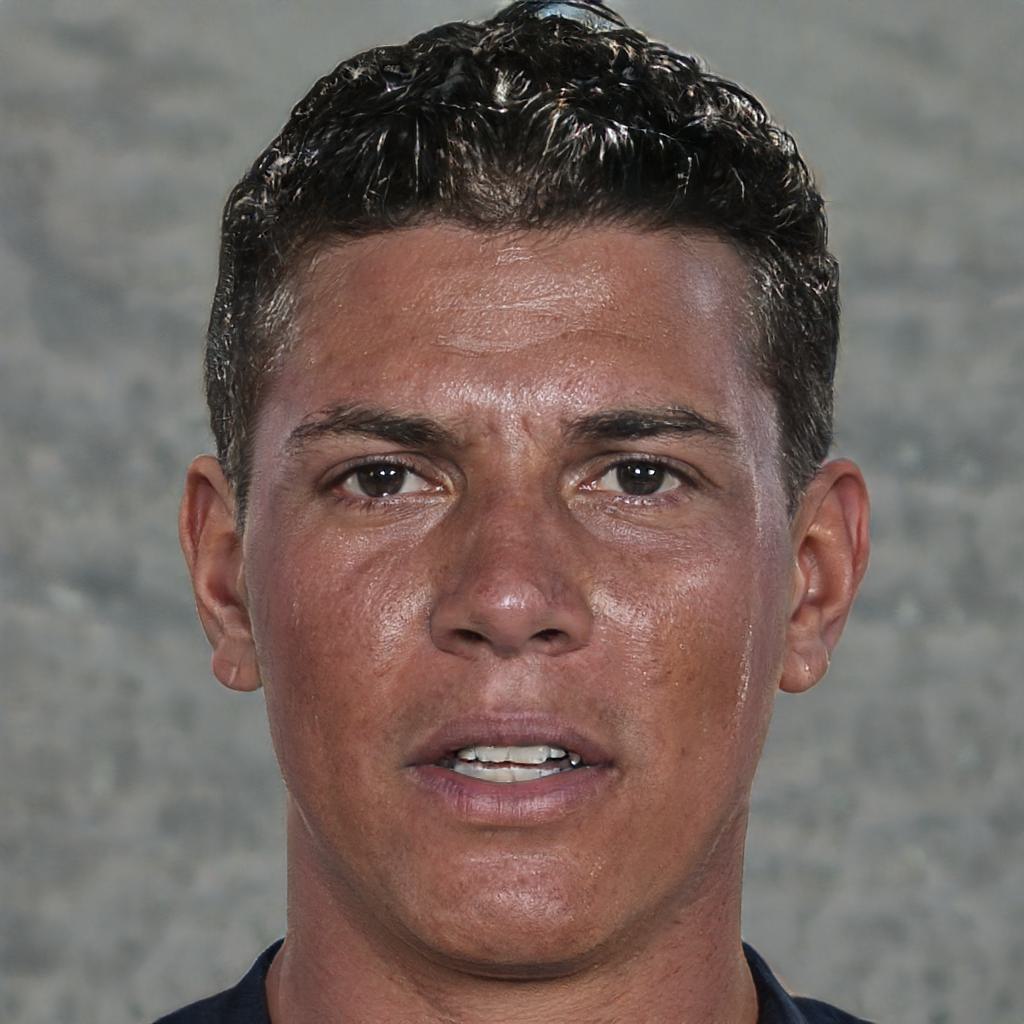} & \includegraphics[width=2.25cm]{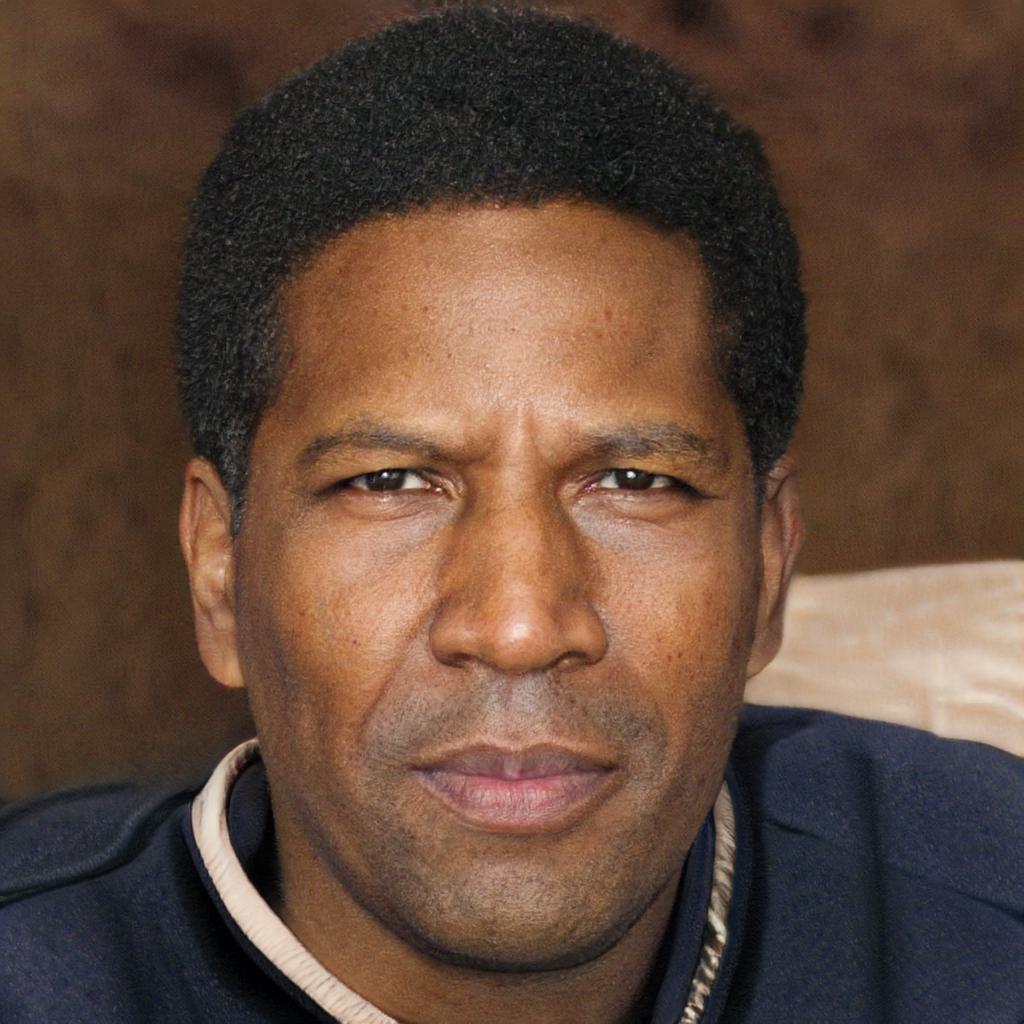} & \includegraphics[width=2.25cm]{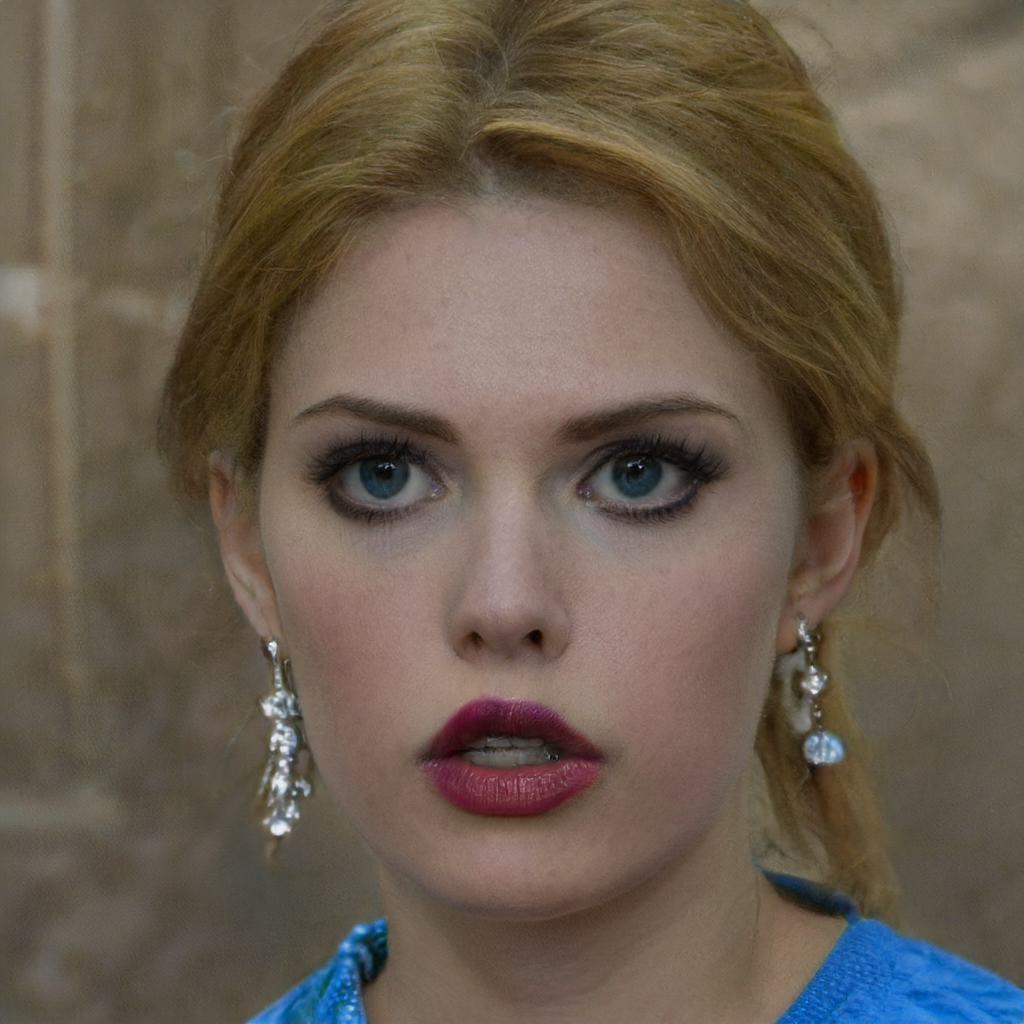} \\ 
    \hspace{2.5ex}0.24 $\pm$ 0.002 & \hspace{2.5ex}0.21 $\pm$ 0.002 & \hspace{2.5ex}0.28 $\pm$ 0.002 & \hspace{2.5ex}0.25 $\pm$ 0.003 & \hspace{2.5ex}0.23 $\pm$ 0.003 & \hspace{2.5ex}0.21 $\pm$ 0.003 \\
    \includegraphics[width=2.25cm]{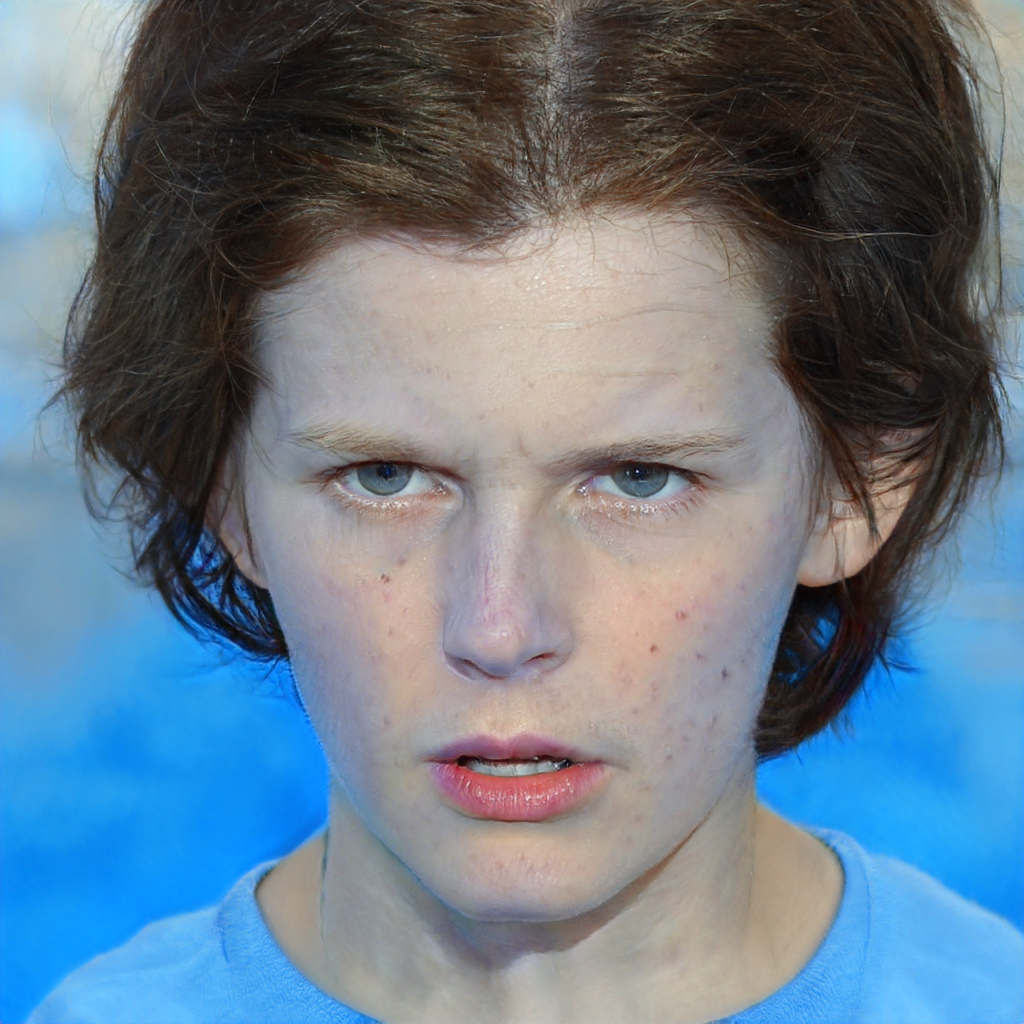} & \includegraphics[width=2.25cm]{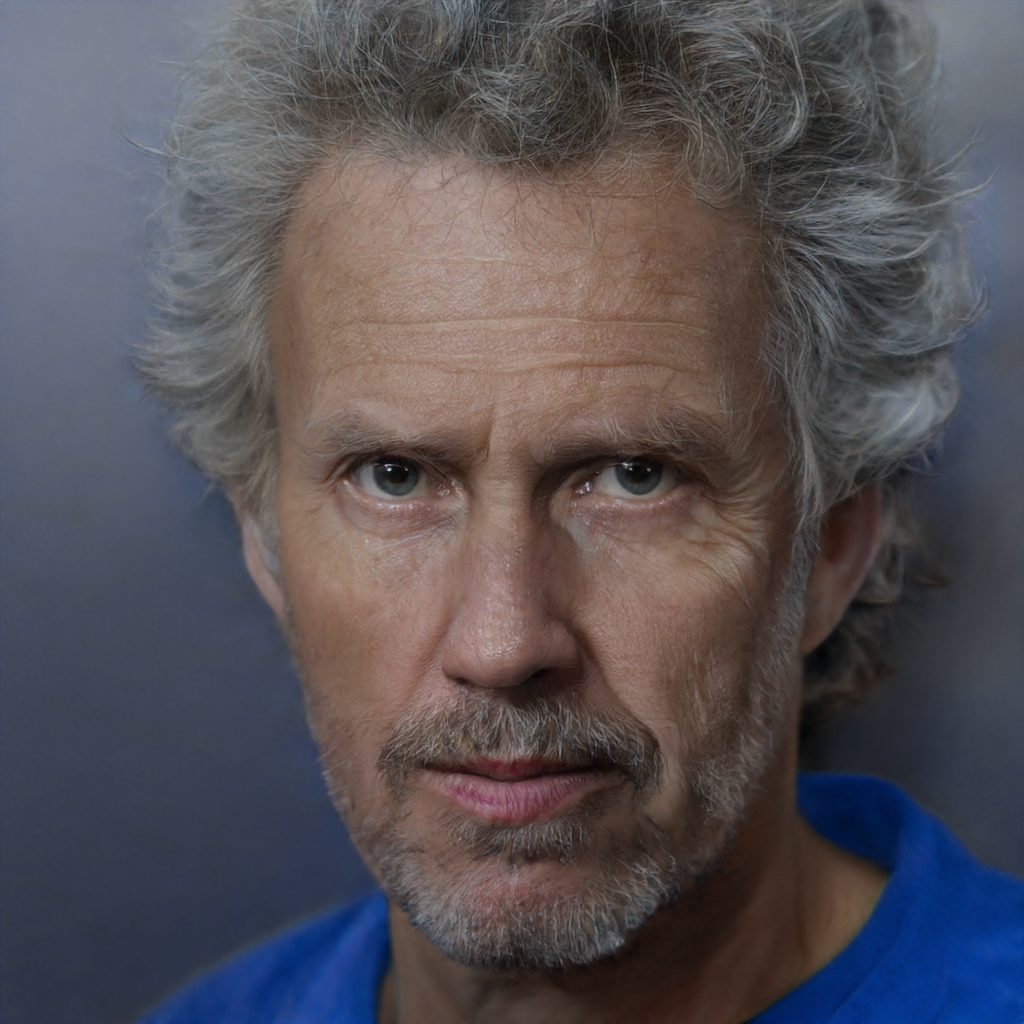} & \includegraphics[width=2.25cm]{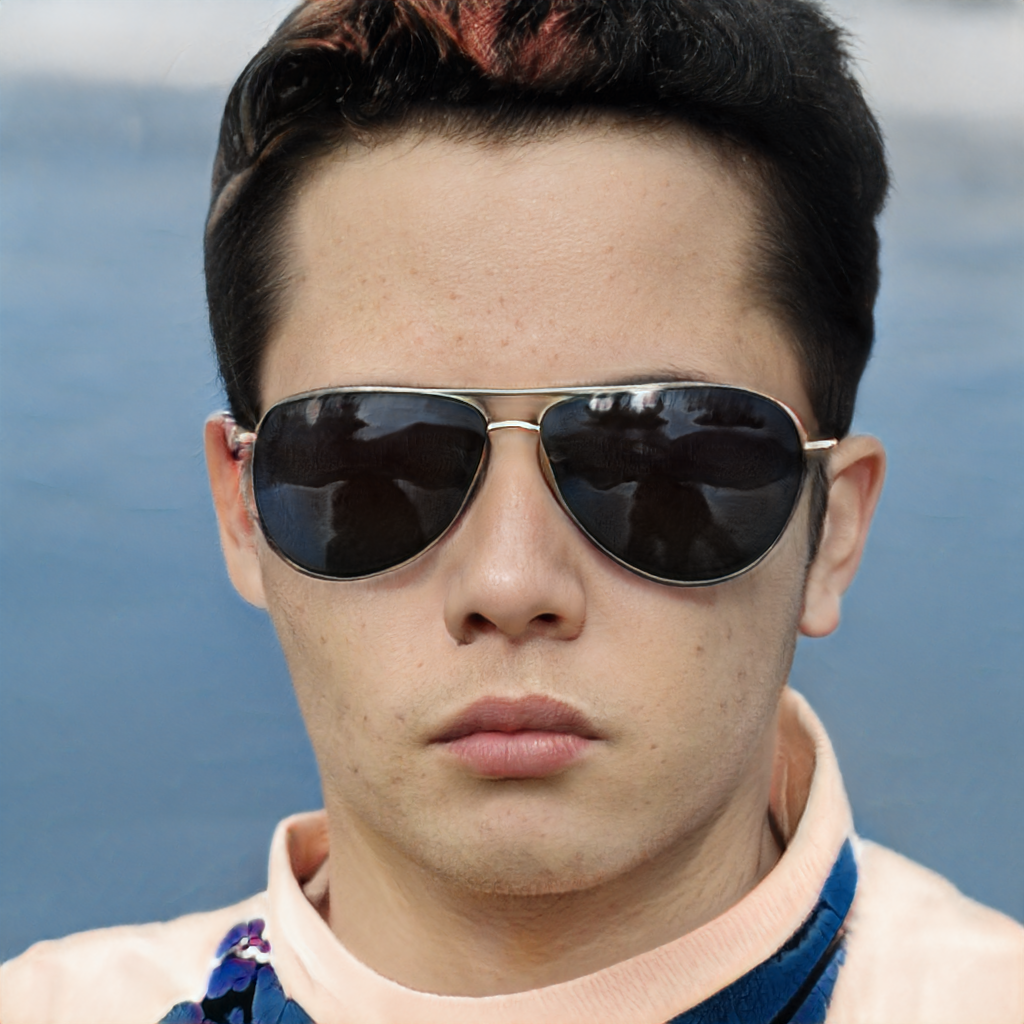} & \includegraphics[width=2.25cm]{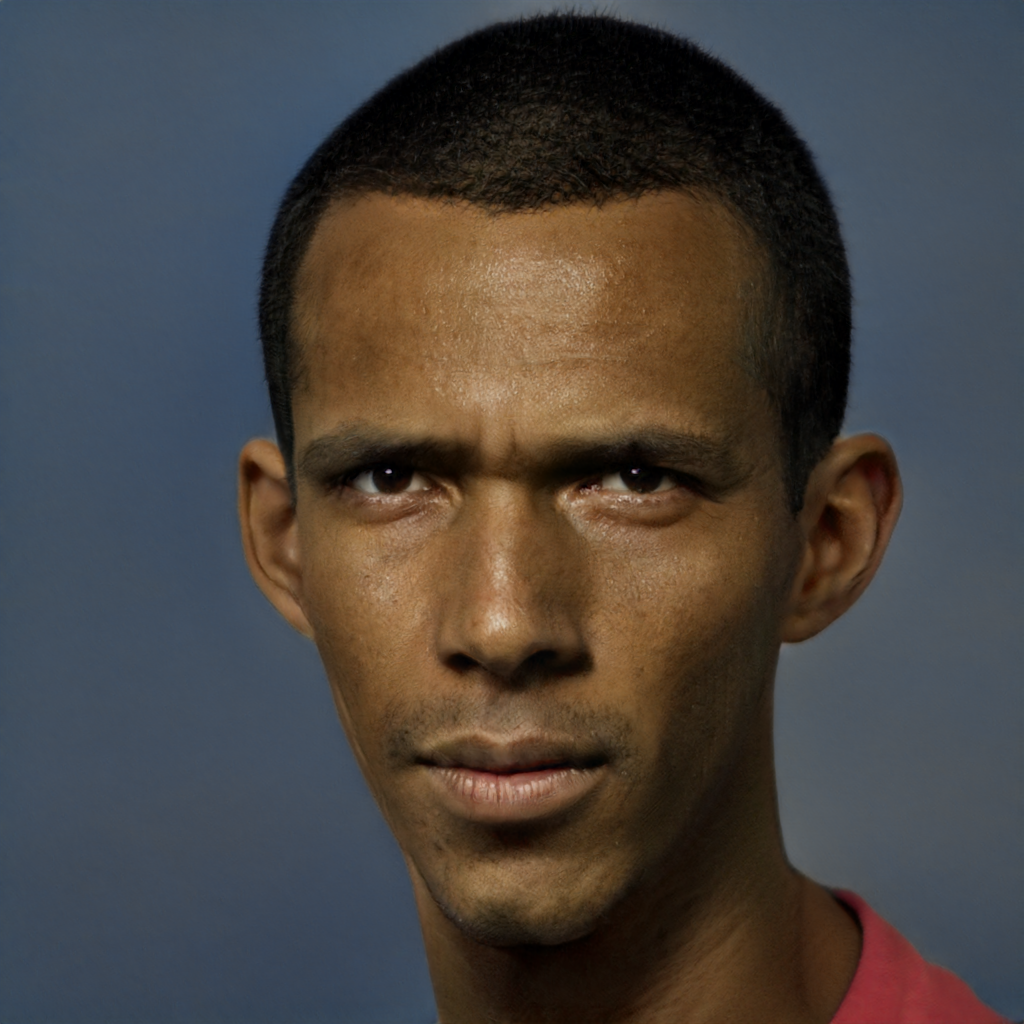} & \includegraphics[width=2.25cm]{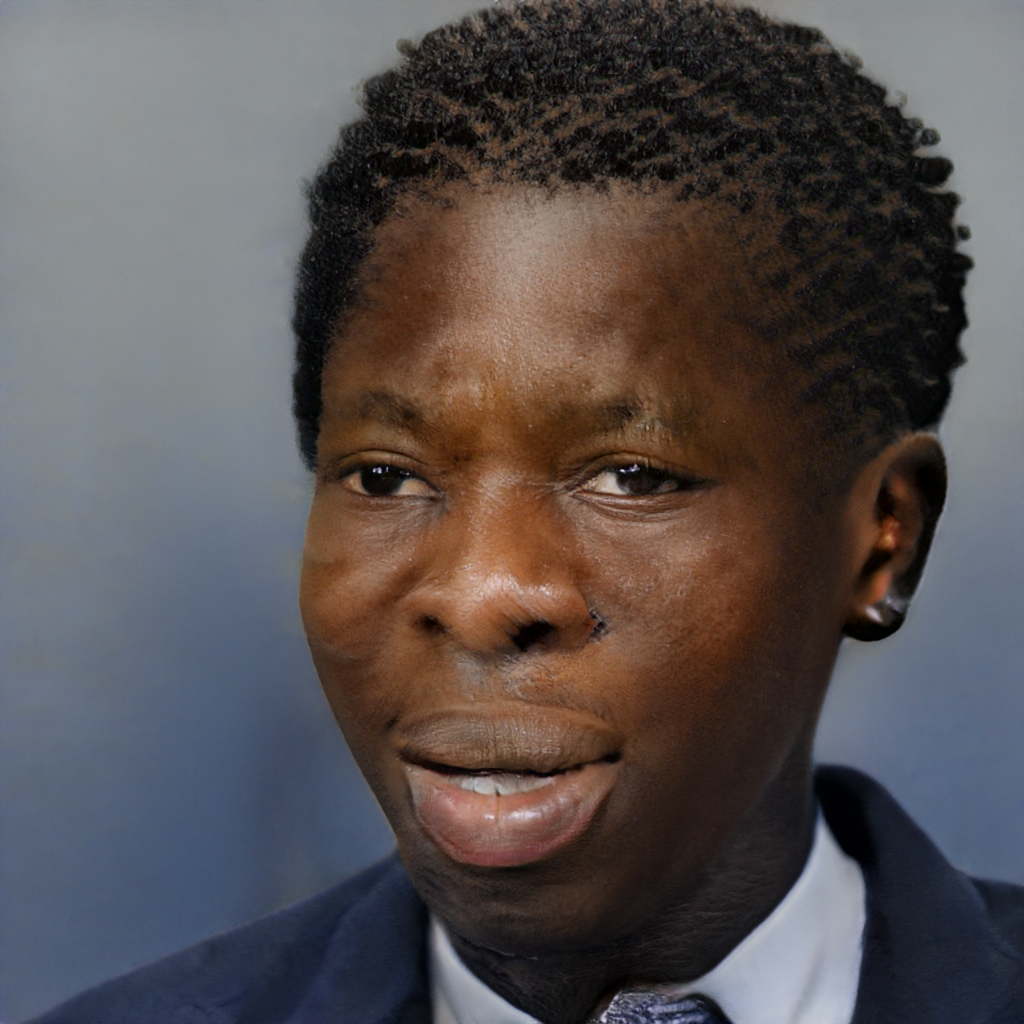} & \includegraphics[width=2.25cm]{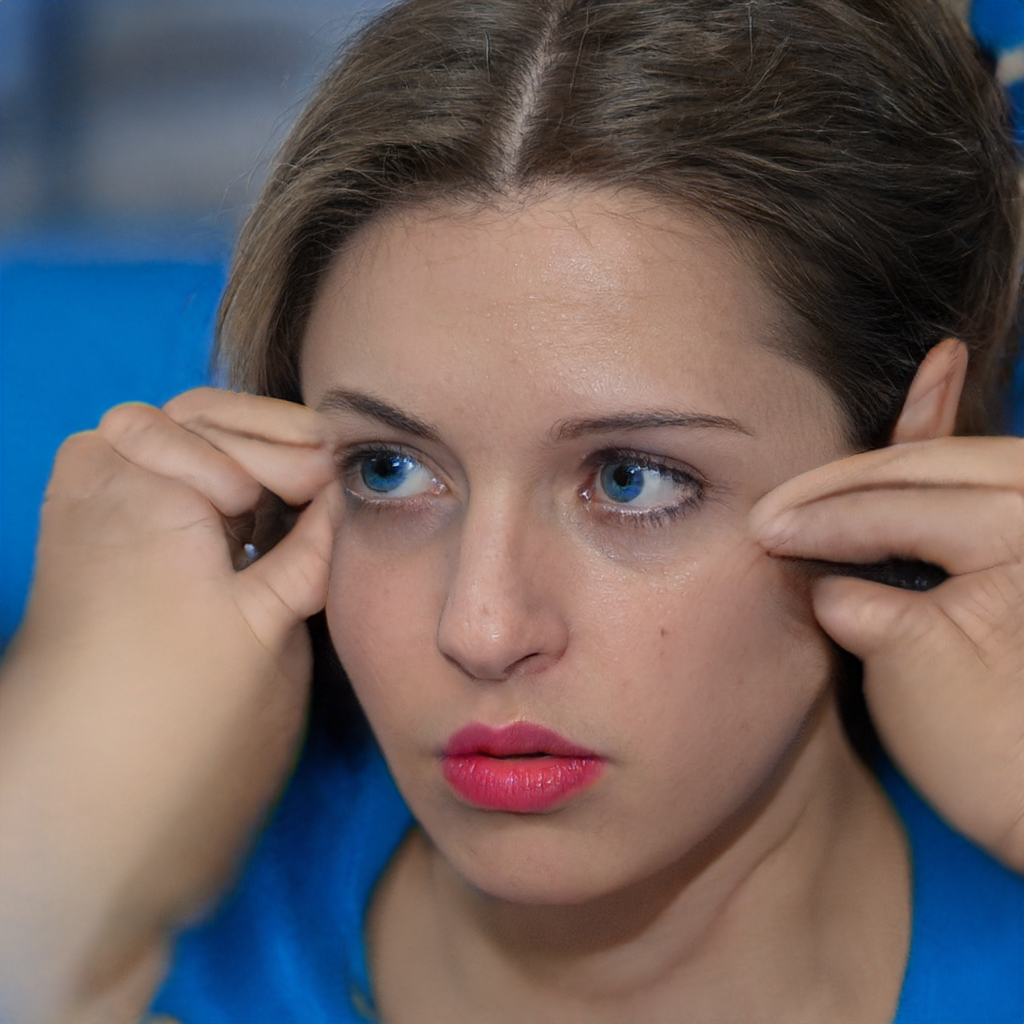} \\
    A child with blue eyes and straight brown hair in the sunshine & \hspace{2.5ex}A haidresser & A young boy with glasses and an angry face & Cristiano Ronaldo & Denzel Washington & \hspace{3.7ex}Cinderella \\[-0.25cm] 
\end{tabular}
\caption{Comparison between TR0N:StyleGAN2+CLIP (\textbf{top row}) and clip2latent (\textbf{bottom row}). Numbers above each image correspond to average augmented CLIP score (higher is better) plus/minus standard error over $10$ samples from the given caption. Thanks to the error correction step, TR0N better semantically matches the input text in its generated images than clip2latent.}
\label{fig:stylegan_faces}
\end{figure*}

We compare TR0N against FuseDream on the MS-COCO dataset, which contains text/image pairs. For each text, we generate a corresponding image with both methods, and then compute both the FID and augmented CLIP score. Results are displayed in \autoref{fig:fid_vs_iters} for various computational budgets (the higher the budget, the bigger $T$, i.e.\ the longer Langevin dynamics is iterated for). 
As a consequence of FuseDream's expensive initialization scheme, TR0N can achieve similar performance much faster. This is true for our first choice of $f$, where TR0N uses the same components as FuseDream (red vs orange lines), emphasizing once again the relevance of the translator, as also evidenced by the light blue lines in \autoref{fig:fid_vs_iters}, which correspond to TR0N with no translator (or equivalently, FuseDream with na\"ive initialization). 
It is also true for our second choice of $f$ (with a caption model), which allows TR0N to not only be faster than FuseDream (which cannot incorporate this $f$ as it has no translator), but also outperform it (blue vs orange lines). 
% TR0N achieves an FID of $16.0$, while FuseDream achieves $17.6$. 
We once again perform ablations over different design choices of the translator, which we include in Appendix \ref{app:extra_exps}.

\autoref{fig:fig1} and \autoref{fig:TR0N_diversity} show text-to-image samples from TR0N. Although BigGAN was trained on ImageNet and remains fixed throughout, the images that TR0N manages to produce from it using text prompts are highly out-of-distribution for this dataset: TR0N's ability to efficiently leverage CLIP to explore the GAN's latent space $\mathcal{Z}$ is noteworthy. We include additional samples in Appendix \ref{app:extra_exps}, showing both how images evolve throughout Langevin dynamics, and failure cases of TR0N.

% By choosing the components of TR0N to match those of FuseDream whenever possible, the previous experiments show that TR0N outperforms this baseline from a methodological standpoint, rather than due to specific network choices. 
By using the same $G$ and version of CLIP as FuseDream, the previous experiments show that TR0N outperforms it thanks to its methodology, rather than an improved choice of networks.
Yet, these networks can be improved. To further strengthen TR0N,  we upgrade: $G$ to a StyleGAN-XL \citep{sauer2022stylegan} -- also pre-trained on ImageNet, CLIP to its LAION2B \citep{schuhmann2022laionb} version, and the caption model to BLIP-2 \citep{li2023blip} (using BLIP-2 instead of BLIP as in other experiments again highlights the plug-and-play nature of TR0N). \autoref{table:ms-coco-fid} shows quantitative results, where we can see that these updates significantly boost the performance of TR0N, to the point of making it competitive with very large models requiring text/image data and much more compute to train. While this StyleGAN-XL-based version of TR0N achieves particularly strong results on MS-COCO in terms of FID, we find that the images it produces are not consistently better, visually, than those from the BigGAN-based model. Samples and further discussion can be found in Appendix \ref{app:extra_exps}.
% we find that the images it produces for out-of-distribution prompts are not visually as good as those from the BigGAN-based version of TR0N. Samples and further discussion can be found in Appendix \ref{app:extra_exps}.

\paragraph{Facial images} To further highlight the wide applicability of TR0N, we show it can be used for other text-to-image tasks. We now use a StyleGAN2 \citep{Karras2019stylegan2} and an NVAE as $G$, both pre-trained on FFHQ \citep{karras2019style}. We use CLIP's image encoder $f^{\text{img}}$ as $f$ (we do not use a caption model here as the descriptions of faces it outputs are too generic to be useful), and use the negative augmented clip score $E_{\text{CLIP}}$ as $E$. We compare against clip2latent, which uses the same setup with the StyleGAN2, but with a diffusion model instead of a GMM as a translator network, and no error correction procedure.

\begin{table}[t]
\vskip -0.1in
\caption{FID score on MS-COCO. The top part of the table shows models trained directly for text-to-image generation using paired text/image data (these are sometimes called zero-shot as they were not trained on MS-COCO, but are not zero-shot in the same way as TR0N). The bottom part shows zero-shot methods that require only pre-trained models and no provided dataset.}
\begin{center}
\scalebox{0.76}{
\begin{tabular}{lcr}
\toprule
Model & FID $\downarrow$ \\
\midrule
DALL-E \citep{ramesh2021zero} & $\approx 27.5^\dagger$\\
StyleGAN-T \citep{sauer2023stylegan} & $13.9^\dagger$\\
Latent Diffusion \citep{rombach2022high} & $12.6^\dagger$\\
GLIDE \citep{nichol2021glide} & $12.2^\dagger$\\
DALL-E 2 \citep{ramesh2022hierarchical} & $10.4^\dagger$\\
Imagen \citep{saharia2022photorealistic} & $7.3^\dagger$\\
Parti \citep{yu2022scaling} & $\mathbf{7.2}^\dagger$\\
\midrule
FuseDream \citep{liu2021fusedream} & $16.3^*$ \\
TR0N:BigGAN+CLIP (BLIP) & $15.0$ \\
TR0N:StyleGAN-XL+LAION2BCLIP (BLIP-2) & $\mathbf{10.9}$\\
\bottomrule
\multicolumn{2}{l}{\small \hspace{5pt} $^\dagger$ Score as reported by the authors, not computed by us.}\\
\multicolumn{2}{l}{\small \hspace{5pt} $^*$ \citet{liu2021fusedream} report an FID of $21.9$ since they use ADAM instead of}\\
\multicolumn{2}{l}{\small \hspace{12pt} Langevin dynamics.}
% and they use a BigGAN which generates $256\times 256$ images, instead of $512\times 512$ and resizing to $256\times 256$ like we do.
\end{tabular}
}
\end{center}
\label{table:ms-coco-fid}
\vskip -0.2in
\end{table}

\begin{figure*} [h!]
\newcommand{\centered}[1]{\begin{tabular}{c}  \\[-1.7cm] #1 \end{tabular}}
\newcommand{\centeredi}[1]{\begin{tabular}{c}  \\[-1.6cm] #1 \end{tabular}}
\renewcommand\fbox{\fcolorbox{black!15!red}{white}}
\centering
\hspace*{-0.2cm}
\begin{tabular}
{ccccc}
   \hspace{-0.1cm}\centered{{\setlength{\fboxsep}{0pt}\setlength{\fboxrule}{0.075cm}\fbox{\includegraphics[width=1.35cm]{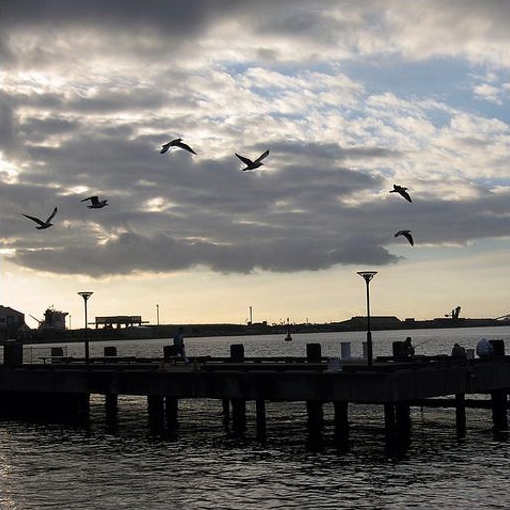}}}} & \hspace{-0.64cm}\includegraphics[width=1.5cm]{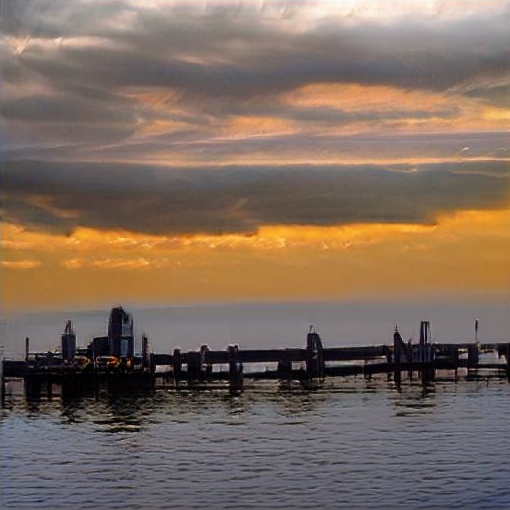} & \centered{{\setlength{\fboxsep}{0pt}\setlength{\fboxrule}{0.075cm}\fbox{\includegraphics[width=1.35cm]{figs/image_var/bear.jpg}}}} & \hspace{-0.64cm}\includegraphics[width=1.5cm]{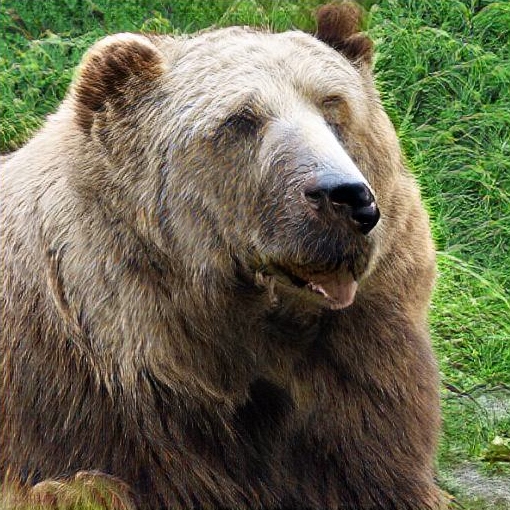} & \centeredi{\includegraphics[width=9.5cm]{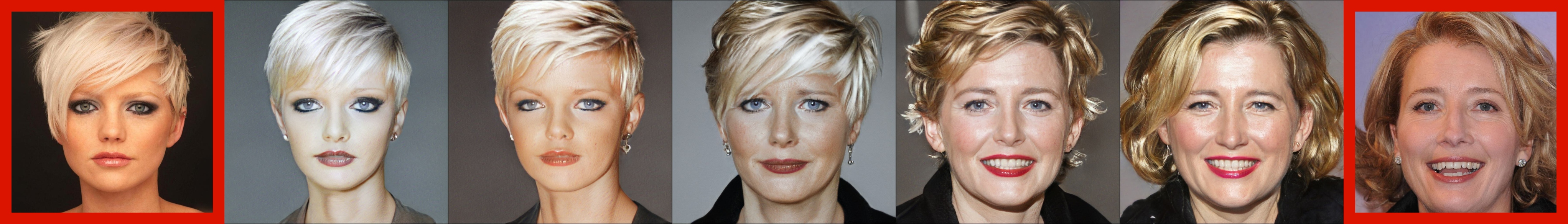}} \\[-0.13cm]
   \hspace{-0.1cm}\includegraphics[width=1.5cm]{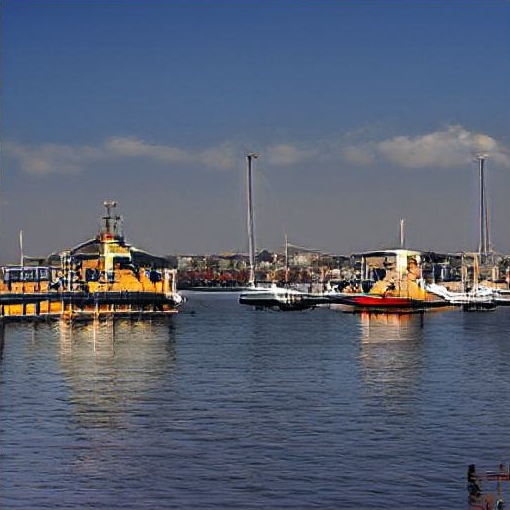} & \hspace{-0.64cm}\includegraphics[width=1.5cm]{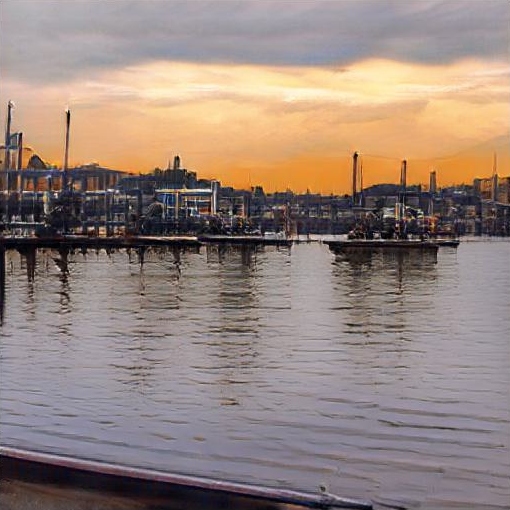} & \includegraphics[width=1.5cm]{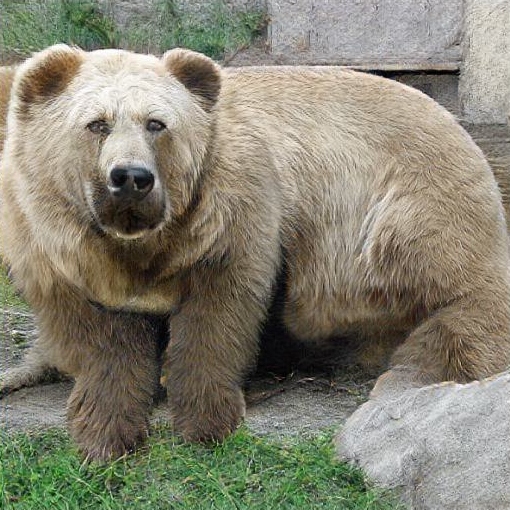} & \hspace{-0.64cm}\includegraphics[width=1.5cm]{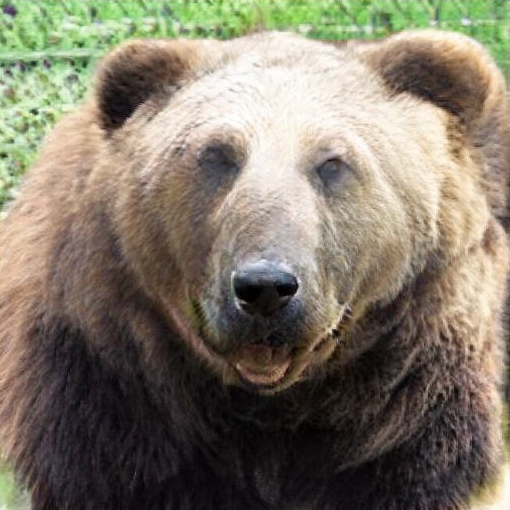} & \centeredi{\includegraphics[width=9.5cm]{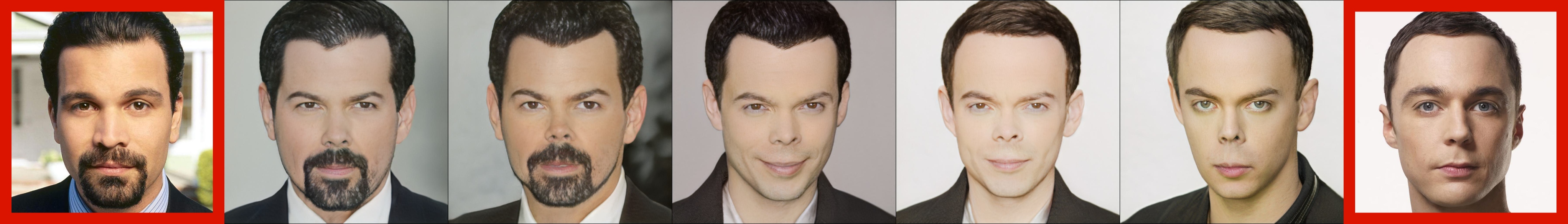}}
\end{tabular}
\caption{TR0N samples conditioning on image semantics with $G$ as a BigGAN (\textbf{first and second panels}, $x'$ is highlighted in red), and interpolations with $G$ as a StyleGAN2 (\textbf{third panel}, $x_1'$ and $x_2'$ are highlighted in red).}
\label{fig:image_semantics}
\end{figure*}

\autoref{fig:fig1} and \autoref{fig:stylegan_faces} show qualitative results. We can see that TR0N produces images that are much more semantically aligned with the input text, which further corroborates that using a GMM as the translator is enough, while also emphasizing the relevance of error-correcting through Langevin dynamics. We highlight that the pushforward models were pre-trained on FFHQ -- not CelebA \citep{liu2015faceattributes} -- and thus likely have not seen celebrities such as Cristiano Ronaldo, Denzel Washington, and Muhammad Ali: we believe TR0N's performance is once again noteworthy. 
We omit large scale quantitative comparisons here because of several reasons: First, text descriptions of FFHQ images are highly generic, which makes it challenging to compute FID against FFHQ. Second, the FID score has recently been shown to be particularly poor at evaluating facial images \citep{kynkaanniemi2022role}. We thus only include the average augmented CLIP score for the used text prompts in \autoref{fig:stylegan_faces}. We include additional samples for the NVAE-based TR0N model in Appendix \ref{app:extra_exps}.

\subsection{Conditioning on image semantics}\label{sec:semantics}

We follow \citet{ramesh2022hierarchical} and consider two tasks which involve conditioning on image semantics: 
For the first, given an image $x'$, the goal is to generate diverse images $x$ which share semantics with $x'$. 
Here, $\mathcal{C}$ is still the latent space of CLIP, $\mathcal{C}_{\text{CLIP}}$, and $f$ is CLIP's image encoder, $f^{\text{img}}$. Instead of obtaining conditions $c$ from a text prompt, we take $c = f^{\text{img}}(x')$. We use both BigGAN and StyleGAN2 as $G$, and still use the negative augmented CLIP score, $E_{\text{CLIP}}$, as $E$. 
For the second task, instead of computing $c$ from a single image $x'$, we compute it by interpolating between the encodings $f^{\text{img}}(x'_1)$ and $f^{\text{img}}(x'_2)$ of two given images, $x'_1$ and $x'_2$. 
Results are shown in \autoref{fig:image_semantics}, where we can see that TR0N produces meaningful samples and interpolations: this highlights that TR0N allows for arbitrary conditioning -- not just class labels or text prompts. We show additional samples in Appendix \ref{app:extra_exps}.

% We follow \citet{ramesh2022hierarchical} and consider the task of conditioning on image semantics: 
% given an image $x'$, the goal is to generate diverse images $x$ which share semantics with $x'$. 
% Our goal here is to show that TR0N allows for arbitrary conditioning (not just class labels or text prompts), but we highlight that this could enable using TR0N as a data-augmentation scheme -- although verifying its effectiveness in this setting falls outside the scope of our work.
% Here, $\mathcal{C}$ is still the latent space of CLIP, $\mathcal{C}_{\text{CLIP}}$, and $f$ is CLIP's image encoder, $f^{\text{img}}$. Instead of obtaining conditions $c$ from a text prompt, we take $c = f^{\text{img}}(x')$. We use both BigGAN and StyleGAN2 as $G$, and still use the negative augmented CLIP score, $E_{\text{CLIP}}$, as $E$. 
% Results are shown in \autoref{fig:image_semantics}, where we can see that TR0N produces meaningful samples. 
% We can also use this type of conditioning to semantically interpolate between images: instead of computing $c$ from a single image $x'$, we can compute it by interpolating between the encodings $f^{\text{img}}(x'_1)$ and $f^{\text{img}}(x'_2)$ of two given images, $x'_1$ and $x'_2$. We show results in Appendix \ref{app:extra_exps}.

\section{Conclusions, limitations, and future work}
In this paper we introduced TR0N, a highly general and simple-to-train framework to turn pre-trained unconditional generative models into conditional ones by learning a stochastic map from conditions to latents, whose output is used to initialize Langevin dynamics. TR0N is quick to sample from, outperforms competing methods, and has a remarkable ability to generate images outside of the distribution used to train $G$. 
Despite the empirical performance of TR0N being good, 
it is inevitably limited by that of the pre-trained model $(p(z), G)$. Diffusion models have been shown to outperform GANs, but have no low-dimensional latent space $\mathcal{Z}$ that the translator can map to, and thus applying TR0N in this setting is not straightforward.

We thus believe extending TR0N to diffusion models to be an interesting direction for future work. We also hope that our ideas can be extended to initialize Langevin dynamics in other EBM settings. Given our results on CIFAR-10 where TR0N improved upon its pre-trained unconditional model, we also believe that further exploring how large pre-trained models can be used to improve upon existing generative models -- rather than endowing them with conditional capabilities -- to be a promising research avenue. 
Finally, here we focused exclusively on generating images, but combining large pre-trained models is of interest outside of this task. For example, zero-shot conditional text generation \citep{su2022language} is a relevant problem, and we hope that the ideas behind TR0N can be extended to this task.
% Finally, here we focused exclusively on generative modeling, but combining large pre-trained models is of interest outside of this task \citep{su2022language}, and we believe that extending translator networks to translate between the latent spaces of various types of models to have potential as a future research direction.

\paragraph{Broader impact} Generative models have many applications, including among others: 
% Generative models have many applications besides image and text-to-image generation, such as 
audio generation \citep{oord2016wavenet, engel2017neural}, chemistry \citep{gomez2018automatic}, neuroscience \citep{sussillo2016lfads, gao2016linear, loaiza2019deep}, and text generation \citep{bowman2016generating, devlin2019bert, brown2020language}. Each of these applications can have meaningful and positive effects on society, but can also be potentially misused for unethical purposes. Text-to-image generation is no exception, and thus the possibility exists that TR0N could be misemployed to generate inappropriate or deceitful content. We do highlight however that other powerful text-to-image models exist and are publicly available, and as such we do not foresee TR0N enabling nefarious actors to abuse text-to-image models in previously unavailable ways.

\section*{Acknowledgements}
We thank Harry Braviner for early discussions, Anthony Caterini for comments on a preliminary draft, and the anonymous reviewers, whose feedback helped improve our paper.

% \textbf{Do not} include acknowledgements in the initial version of
% the paper submitted for blind review.

% If a paper is accepted, the final camera-ready version can (and
% probably should) include acknowledgements. In this case, please
% place such acknowledgements in an unnumbered section at the
% end of the paper. Typically, this will include thanks to reviewers
% who gave useful comments, to colleagues who contributed to the ideas,
% and to funding agencies and corporate sponsors that provided financial


\bibliography{main}
\bibliographystyle{icml2023}

%%%%%%%%%%%%%%%%%%%%%%%%%%%%%%%%%%%%%%%%%%%%%%%%%%%%%%%%%%%%%%%%%%%%%%%%%%%%%%%
%%%%%%%%%%%%%%%%%%%%%%%%%%%%%%%%%%%%%%%%%%%%%%%%%%%%%%%%%%%%%%%%%%%%%%%%%%%%%%%
% APPENDIX
%%%%%%%%%%%%%%%%%%%%%%%%%%%%%%%%%%%%%%%%%%%%%%%%%%%%%%%%%%%%%%%%%%%%%%%%%%%%%%%
%%%%%%%%%%%%%%%%%%%%%%%%%%%%%%%%%%%%%%%%%%%%%%%%%%%%%%%%%%%%%%%%%%%%%%%%%%%%%%%
\newpage
\appendix
\onecolumn
% \section{Appendix}
\section{TR0N for Bayesian inference}\label{app:bayes}

As mentioned in the main manuscript, in the setting where we have access to a probabilistic model $p(c|x)$, a joint distribution $p(z,x,c)=p(z)\delta_{G(z)}(x)p(c|x)$ is implied. The goal of Bayesian inference is to sample from the corresponding conditional (posterior) distribution $p(x|c)$. Since $p(z,x|c) = p(z|c) \delta_{G(z)}(x)$, sampling from $p(z|c)$ and transforming the result through $G$ allows to obtain $(z,x)$ pairs from $p(z,x|c)$. Discarding $z$, or equivalently marginalizing out $z$ from $p(z,x|c)$, results in samples from $p(x|c)$. In other words, we only need to sample from $p(z|c)$ and transform the result through $G$ in order to perform Bayesian inference. Furthermore, we know that $p(z|c) \propto p(z,c) = \int p(z)\delta_{G(z)}(x)p(c|x) dx= p(z)p(c|x=G(z))$, which means that TR0N can be used to sample from $p(z|c)$ by setting $\beta=1$ and taking $E$ as $E_{\text{Bayes}}$, where
\begin{equation}\label{eq:bayes}
    E_{\text{Bayes}}(z,c) = -\log p(z) - \log p(c|x=G(z)).
\end{equation}
Note that when $p(z) = \mathcal{N}(z;0, \nu^2 I)$, the first term is just an $L_2$ penalty, i.e.\ $-\log p(z) = \frac{1}{2\nu^2}\Vert z \Vert_2^2$ (up to an additive constant that does not depend on $z$ and is thus irrelevant for TR0N). 
As mentioned in the main manuscript, we find that non-Bayesian choices obtain stronger empirical results. This observation is consistent with previous works, e.g.\ \citet{dhariwal2021diffusion} find -- in a different context -- that equally weighting the density $p(z)$ and classifier $p(c|x=G(z))$ terms as in $E_{\text{Bayes}}$ is suboptimal, and that more heavily weighting the classifier term leads to improved empirical performance. As an attempt to improve performance in our experiments in Appendix \ref{app:extra_exps}, we thus also consider sampling from a distribution proportional to $e^{-E_{\text{Bayes}}'(z,c)}$, where
\begin{equation}\label{eq:bayes_prime}
    E_{\text{Bayes}}'(z,c) = -\beta_1 \log p(z) - \beta_2 \log p(c|x=G(z)),
\end{equation}
and $\beta_1$ and $\beta_2$ are used as hyperparameters instead of $\beta$. Note that $\beta_1=\beta_2=1$ corresponds to Bayesian inference.

\section{Experimental details}\label{app:exp_details}

\paragraph{FuseDream's initialization} FuseDream \citep{liu2021fusedream} uses a similar procedure to TR0N  to initialize $z^{(0)}$ as described in Algorithm \ref{alg:sampling} except, since \citet{liu2021fusedream} have no translator, they have to use the prior $p(z)$ to sample candidates $z_m$. As a result, they need orders of magnitude more samples $M$ in order to obtain a decent initializer (i.e. a sample roughly matching the given condition $c$). This is costly, as $E_\text{CLIP}(z_m, c)$ has to be evaluated for every $m=1,\dots,M$, which requires a forward pass for $G$ and for $f^{\text{img}}$. When $M$ is very large -- as is required by FuseDream, e.g.\ $M=1000$ -- this initialization takes a non-negligible amount of time. In particular, FuseDream uses the parameterization from \eqref{eq:langevin_params}, but since no GMM parameters are available, FuseDream uses the $K$ latents with lowest $E_\text{CLIP}$ out of the $M$ sampled ones to initialize each $\mu_k^{(0)}$, for $k=1,\dots, K$, and initializes $w^{(0)}$ randomly.

\paragraph{Deterministic translators} For our ablations using a deterministic translator, we specify a neural network $S_\theta: \mathcal{C} \rightarrow \mathcal{Z}$ instead of $q_\theta(z|c)$. We train this translator as a regressor with an $L_2$ loss:
\begin{equation}\label{app_eq:l2}
    \theta^* = \argmin_\theta \mathbb{E}_{p(z)}\left[\Vert S_\theta(f(G(z))) - z\Vert_2^2\right].
\end{equation}

\paragraph{NVAE-based models} The NVAE model is presented as having a hierarchical latent space $\mathcal{Z}' = \mathcal{Z}_0 \times \dots \times \mathcal{Z}_L$, whose prior is given by
\begin{equation}\label{eq:nvae_prior}
    p(z') = p(z_0) \displaystyle \prod_{\ell=1}^L p(z_\ell|z_{\ell - 1}),
\end{equation}
where $z'=(z_0,\dots,z_L)$, $p(z_0)$ is a standard Gaussian, and $p(z_\ell|z_{\ell - 1}) = \mathcal{N}(z_\ell; g_\ell(z_{\ell - 1}), \Sigma_\ell(z_{\ell - 1}))$, where $g_\ell$ and $\Sigma_\ell$ are neural networks. The NVAE model also has a decoder $p(x|z_L)$. In order to produce a data sample, once $z_L$ is sampled from \eqref{eq:nvae_prior}, $G'(z_L)$ is computed to obtain a sample on $\mathcal{X}$, where $G'$ is the mean of the decoder $p(x|z_L)$. Thus, the NVAE fits exactly into the framework of a pushforward model with $\mathcal{Z}=\mathcal{Z}_L$, $p(z)$ as the $z_L$-marginal of \eqref{eq:nvae_prior}, and $G$ as $G'$. Na\"ively, we could thus have the translator be a distribution over $z_L$, $q_\theta(z_L|c)$, and apply TR0N. However, $z_L$ is actually high-dimensional, and we found this approach did not work particularly well. We thus take a slightly different view of the same NVAE model, where $\mathcal{Z}=\mathcal{Z}_0$, $p(z)$ is just $p(z_0)$, and $G$ is now a stochastic map: in order to compute $G(z_0)$, one samples $z_{\ell}$ from $p(z_\ell|z_{\ell-1})$ for $\ell=1,\dots,L$, until $z_L$ is obtained, and then computes $G'(z_L)$. In this view, the NVAE is a pushforward model with a low-dimensional latent space and a stochastic $G$, to which we can also apply TR0N. While this approach worked better, we found that $\mathcal{Z}_{0}$ generally did not provide sufficient semantic control over generated images due to the added noise from $(p(z_\ell|z_{\ell - 1}))_{\ell=1}^{L}$, essentially rendering conditioning very hard. We thus slightly modify $G$ as follows so as to effectively remove sources of randomness: we fix an index $\ell^*$, and deterministically transform $z_\ell$ until $\ell^*$ using the mean neural networks, i.e.\
\begin{equation}\label{eq:nvae_z_det}
    z_\ell = g_\ell(z_{\ell-1}) \text{ for }\ell=1,\dots,\ell^*,
\end{equation}
and only sample the remaining entries of the hierarchy:
\begin{equation}\label{eq:nvae_z_rand}
    z_\ell|z_{\ell-1} \sim p(z_\ell|z_{\ell-1}) \text{ for }\ell=\ell^*+1,\dots,L.
\end{equation}
Now, $G(z_0)$ is given by $G'(z_L)$, where $z_L$ is now obtained from $z_0$ from \eqref{eq:nvae_z_det} and \eqref{eq:nvae_z_rand}. While the formalism of Langevin dynamics is broken here since $E(z,c)$ cannot be evaluated exactly due to the randomness of $G$, the intuition behind TR0N remains, and we find that even if for a fixed $z_{0}$, different calls to $G(z_{0})$ return different random images (i.e. $G$ is stochastic), TR0N provides strong empirical performance in this setting. We also point out that using a single scalar $\lambda$ in Langevin dynamics -- which as mentioned in the main manuscript is a common practice -- also formally breaks down the interpretation of EBM sampling. We thus do not see randomness in $G$ as a fundamental limitation and argue that we are performing approximate Langevin dynamics in this setting. 
% Similarly, while having a stochastic $G$ biases our stochastic gradient estimates of \eqref{eq:ml} (when an expectation is taken over $G$), we find that the translator is still properly trained since TR0N works well with NVAE-based models.

\paragraph{On using $\mathbf{E_\text{CLIP}}$ and the caption model} We note that much like the randomness in $G$ as described above for the NVAE-based models, when we use $E_\text{CLIP}$, the formalism of Langevin dynamics breaks since $E$ cannot be exactly evaluated, and only approximated by sampling data augmentations. Similarly to the NVAE case though, we find that TR0N works well regardless, and we see this as approximate Langevin dynamics. Also, the caption model $h$ that we use to perform text conditioning using BigGAN is stochastic, which means that $f$ is stochastic when training the translator.
% although unlike the case of stochastic $G$, stochastic gradient estimates of \eqref{eq:ml} (when an expectation is taken over $f$) remain unbiased.

\paragraph{BigGAN-based models} As mentioned in the main manuscript, the BigGAN model is a class-conditional model and its latent space $\mathcal{Z}=\mathcal{Z}_1 \times \mathcal{Z}_2$ has a continuous component $\mathcal{Z}_1$, and a discrete component $\mathcal{Z}_2$ composed of a discrete set of tokens, one per ImageNet class. We found that, while the stochastic GMM translator was fundamental to model $\mathcal{Z}_1$, it was not so for $\mathcal{Z}_2$. We thus used a GMM only for the continuous part of the latent space, $\mathcal{Z}_1$, and a deterministic translator (trained with an $L_2$ loss as described above) for the discrete part, $\mathcal{Z}_2$. In practice we use a single shared translator, with GMM heads for $\mathcal{Z}_1$ and a regression head for $\mathcal{Z}_2$.

\paragraph{StyleGAN2-based models} To ensure that we are directly comparable with clip2latent \citep{Pinkney2022clip2latent}, for StyleGAN2-based models, we define $\mathcal{Z}$ as the intermediate latent space of StyleGAN2, $\mathcal{W}$. During training, we first sample Gaussian noise and pass it through the mapping network of StyleGAN2 to obtain a latent $w \in \mathcal{W}$. Note that this procedure implicitly defines the prior $p(w)$ as a pushforward distribution whose density cannot be evaluated, although this is not a problem since training the translator \eqref{eq:ml} requires only sampling from the prior, not evaluating its density.
% , so that we define $p(z)$ as $p(w)$.

\paragraph{StyleGAN-XL-based models} 
% To further show that TR0N can be improved upon by upgrading its individual pre-trained components, we included the StyleGAN-XL model as the generator to achieve better performance.
StyleGAN-XL is a class-conditional model like BigGAN. However, similar to StyleGAN2-based models, we define $\mathcal{Z}$ as the intermediate latent space of StyleGAN-XL, $\mathcal{W}$, which has already encoded the class conditioning. During training of the translator, to obtain a latent $w \in \mathcal{W}$, we first sample a random noise latent $z_1  \sim \mathcal{N}(z_1; 0, I)$ and a class label $z_2 \in \mathcal{Z}_2$ uniformly at random, then pass them both through StyleGAN-XL's mapping network to recover a desired sample $w \in \mathcal{W}$.
% , so that we define $p(z)$ as $p(w)$.

\subsection{Details and hyperparameters that are shared across all experiments}

\paragraph{Translator architecture} We use an MLP as the translator consisting of two hidden layers with a residual connection in between, and ReLU activations after each hidden layer. 
Both hidden layers have the same number of hidden units $H$, which we set for each experiment. 
A set of heads following the second hidden layer then produce the desired outputs. For the GMM heads, we have one head to predict the $K$ Gaussian means $(\mu_{\eta, k}(c))_{k=1}^K \in \mathcal{Z}^K$ (the head outputs the concatenated means which are then reshaped) and another head to predict the mixture weights $w_\eta(c) \in \mathbb{R}^K$. In all experiements, $K=10$. For the regression heads, we directly predict the latent $z$. For the GMM translator, we also learn a separate standard deviation $\sigma$ of the same size as $z$. To ensure its training remains stable, we instead learn a parameter $\nu$ and then recover $\sigma$ as $\sigma = e^{\nu}+10^{-6}$, guaranteeing that $\sigma > 10^{-6}$. We initialize the standard deviation to $\sigma=0.01$. All other translator weights are randomly initialized using the default PyTorch \citep{NEURIPS2019_9015} linear layer initializer.

\paragraph{Translator training} In practice, instead of sampling from $p(z)$ at each step during translator training and then feeding the latent sample through $G$ and $f$ as in Algorithm \ref{alg:training}, we generate a synthetic dataset of $(z, c)$ pairs ahead of time by sampling $N=1,000,000$ latents $z\sim p(z)$ and recovering the corresponding conditions $c=f(G(z))$. While not necessary, we perform this step simply to speed up training the translator since, once the synthetic dataset is generated, we no longer need to perform expensive forward passes through $G$ and $f$. We train the translator for 10 epochs on said synthetic dataset with a batch size $B=16$. 
We find in our experiments that the maximum likelihood loss outlined in Algorithm \ref{alg:training} takes values with high orders of magnitude, so we scale the loss with a scalar -- we use $10^{-4}$ -- so that we can use standard learning rates schemes. We thus use ADAM to optimize the translator network with a learning rate of $10^{-4}$ and a cosine scheduler to anneal the learning rate to $0$ throughout training. Unless otherwise stated, $p(z)$ is a standard Gaussian in our experiments.

\paragraph{TR0N sampling} While Langevin dynamics can be understood as gradient descent with added noise, in practice we use gradient descent with momentum and added noise since it helps empirical performance. We set the momentum to 0.99 and add noise with $\lambda=10^{-4}$ in all experiments. Unless otherwise stated, we use $T=100$ steps of Langevin dynamics. We also note that while in \eqref{eq:langevin_params} we update both $w$ and $\mu$ through Langevin dynamics for didactic purposes, we find that applying Langevin dynamics only to $\mu$ and that deterministically updating $w$ with ADAM to work better in practice (we use a learning rate for $w$ of $5 \times 10^{-3}$ for all experiments). Also, we find that expanding each $(w_k)_{k=1}^K$ to be of the same size as $\mu$ such that each component of $\mu$ has its own weights to be a helpful heuristic.

\subsection{Conditioning on class labels}

To train the translator, we set $H=100$. For Langevin dynamics, we use $\beta=20$ and iterate for $T=25$ steps for AutoGAN and $T=50$ steps for NVAE. In addition, we use the initialization described in Algorithm \ref{alg:sampling} with $M=5$ for AutoGAN and $M=10$ for NVAE. Also, for NVAE, we use $p(z)$ as a standard Gaussian with standard deviation of 0.7 and we use $\ell^{*}=0$.

\subsection{Conditioning on text}

As mentioned in the main manuscript, we use two choices of $f$. The first uses the CLIP image encoder $f=f^\text{img}$. For this choice of $f$, taking inspiration from \citet{zhou2021lafite} and \citet{nukrai2022text}, we add noise during TR0N training to simulate ``pseudo-text embeddings'' from image embeddings in the joint latent space of CLIP, $\mathcal{C}_{\text{CLIP}}$. This helps account for the fact that the fixed condition $c=f^{\text{txt}}(y)$ uses the the text encoder, while $c_i$ in Algorithm \ref{alg:training} uses the image encoder, $f^{\text{img}}$. More specifically, we modify \eqref{eq:ml} to: 
\begin{equation}
    \theta^* = \argmin_\theta \mathbb{E}_{p(z)}\left[-\log q_\theta \left(z|c=f(G(z))+\gamma \epsilon \right)\right],
\end{equation}
where $\epsilon \sim \mathcal{N}(\epsilon; 0, I)$. We find $\gamma=0.2$ to work well in these experiments. 

For the second choice of $f$ -- which uses a caption model $h$ to obtain $f=f^\text{txt} \circ h$ -- we do not add Gaussian noise. This is because the caption model produces embeddings that are closer to the text conditions $c$ that the translator will observe during TR0N sampling. As mentioned, we use BLIP (or BLIP-2) as the caption model and leverage nucleus sampling \citep{holtzman2019curious} to generate more diverse and semantically meaningful captions. 

In all these experiments, we set $H=2048$ in the translator's architecture. In terms of TR0N sampling, we perform Langevin dynamics as described in \eqref{eq:langevin_params}. Also, we approximate $E_{\text{CLIP}}$ by sampling 50 times from $p(\phi)$ and then taking the average of $U_\text{sim}$ across these sampled data augmentations. For BigGAN, we use $\beta=10^5$ and for StyleGAN2, StyleGAN-XL and NVAE, we use $\beta=2000$.   

For BigGAN, we bound the continuous component $\mathcal{Z}_1$ of $\mathcal{Z}$ (see note above) to be in range $[-2,2]$, similar to FuseDream. When training the translator, we enforce this in a differentiable way using a $\tanh$ activation at the output of the GMM mean and regression heads. Moreover, we can sample from the discrete component $\mathcal{Z}_2$ of $\mathcal{Z}$ by uniformly sampling an ImageNet class. In practice, when generating our synthetic dataset, we generate the same number of samples from each ImageNet class to ensure mode coverage when training the translator. As mentioned before, we use a deterministic translator for $\mathcal{Z}_2$, whose output we deterministically update during TR0N sampling with ADAM using a learning rate of $5 \times 10^{-3}$. We therefore only apply \eqref{eq:langevin_params} to the continuous component $\mathcal{Z}_1$.

For StyleGAN-XL, to sample a class label, we again in practice sample the same number of samples from each ImageNet class when generating the synthetic dataset and use $p(z)$ as outlined above.

Also, for StyleGAN2, we use $p(z)$ as outlined above. For NVAE, we use $p(z)$ as a standard Gaussian with standard deviation scaled by $0.6$ and we use $\ell^{*}=7$.

We note that to compute all FID scores in \autoref{fig:fid_vs_iters}, we use the entire validation set of MS-COCO, which contains $40k$ text/image pairs. To compute the FID scores in \autoref{table:ms-coco-fid}, we randomly subsample $30k$ points from the validation set. We do this for valid comparisons, since the papers whose FID we report on the top part of \autoref{table:ms-coco-fid} use only $30k$ samples. While we find it more principled to use the entire dataset, we find no major difference between these two ways of computing FID. Note also that the FID scores we report on the upper part of \autoref{table:ms-coco-fid} are from models which produce images at a $256 \times 256$ resolution, whereas the BigGAN and StyleGAN-XL models that we use for the lower part of the table produce images at a $512 \times 512$ resolution. For the fairest possible comparison, we thus downscale the $512 \times 512$ images to $256 \times 256$ before computing the FID score.

The experiments that required timing were all run on an NVIDIA TITAN RTX. Note that a single Langevin dynamics step takes the exact same amount of time in TR0N as it does in FuseDream, as we use the same choice of $E$ and $K$, so that the gains we observe in TR0N are truly due to the translator's efficient initialization, since FuseDream requires $M=1000$ forward passes through $G$ just to initialize Langevin dynamics. In contrast, since for TR0N we use \eqref{eq:langevin_params}, we need no forward passes through $G$ to initialize Langevin dynamics: a single pass through the translator is enough. For these experiments, we evaluate metrics at $T \in \{0, 10, 25, 50, 100, 150, 250, 500\}$.

\subsection{Conditioning on image semantics}

To train the translator, we set $H=2048$. For Langevin dynamics, we use use the same hyperparameters as in text conditioning and we also approximate $E_{\text{CLIP}}$ the same way. In addition, we use the initialization described in Algorithm \ref{alg:sampling} with $M=1$ for the first task (with a single image), and the mode of the GMM for the second task (interpolation). For BigGAN, we follow the same procedure as described for text conditioning to deal with the continuous and discrete components of $\mathcal{Z}$, and for StyleGAN2 we again use the $\mathcal{W}$ space as before. For the interpolation experiments, we use slerp interpolation.

\section{Additional experiments}\label{app:extra_exps}

\subsection{Conditioning on class labels through Bayesian inference}

\begin{table}[t!]
\vskip -0.15in
\caption{FID, IS, and average probability assigned to the intended class of generated samples by a ResNet50 on CIFAR-10.}
\begin{center}
\begin{tabular}{lcccr}
\toprule
Model & FID $\downarrow$ & IS $\uparrow$ & Avg. prob. $\uparrow$\\
\midrule
NVAE & $41.70$ & $6.95$ & $-$ \\
TR0N:NVAE+ResNet50 ($\beta_1=0$, $\beta_2=20$) & $\mathbf{19.79}$ & $\mathbf{8.64}$ & $\mathbf{0.75}$ \\
TR0N:NVAE+ResNet50 ($\beta_1=1$, $\beta_2=1$) & $20.12$ & $8.40$ & $0.51$ \\
TR0N:NVAE+ResNet50 ($\beta_1=1$, $\beta_2=20$) & $20.34$ & $8.62$ & $0.68$ \\

\midrule
AutoGAN & $12.45$ & $8.53$ & $-$ \\
TR0N:AutoGAN+ResNet50 ($\beta_1=0$, $\beta_2=20$) & $\mathbf{10.69}$ & $\mathbf{8.91}$ & $\mathbf{0.95}$\\
TR0N:AutoGAN+ResNet50 ($\beta_1=1$, $\beta_2=1$) & $10.85$ & $8.86$ & $0.89$ \\
TR0N:AutoGAN+ResNet50 ($\beta_1=1$, $\beta_2=20$) & $10.71$ & $ 8.90 $ & $\mathbf{0.95}$ \\
\bottomrule
\end{tabular}
\end{center}
\label{table:cifar-bayes}
\vskip -0.20in
\end{table}

We compare in \autoref{table:cifar-bayes} the performance of $E_{\text{Bayes}}'$ from \eqref{eq:bayes_prime} against the choice from the main manuscript for conditioning on class labels on CIFAR-10, i.e.\ using \eqref{eq:score} with $U$ as the cross entropy loss, $U_{\text{ent}}(c', c) = -\sum_j c_j \log c_j'$. Note that since $U_{\text{ent}}(f(G(z)), c) = -\log p(c|x=G(z))$, using $E_{\text{Bayes}}'$ amounts simply to adding an $L_2$ penalty on $z$ and re-weighting the classifier term. For all experiments, we use the exact same translator: only the energy function used for Langevin dynamics changes. \autoref{table:cifar-bayes} includes results for: the energy function from used in the main manuscript ($\beta_1= 0, \beta_2=20$); the fully Bayesian choice ($\beta_1=\beta_2=1$), which as previously mentioned hurts performance; and also for $\beta_1=1$ and a tuned value of $\beta_2$, which loses the Bayesian interpretation and does not outperform the first choice.

\subsection{Ablations for text-to-image results with natural images}

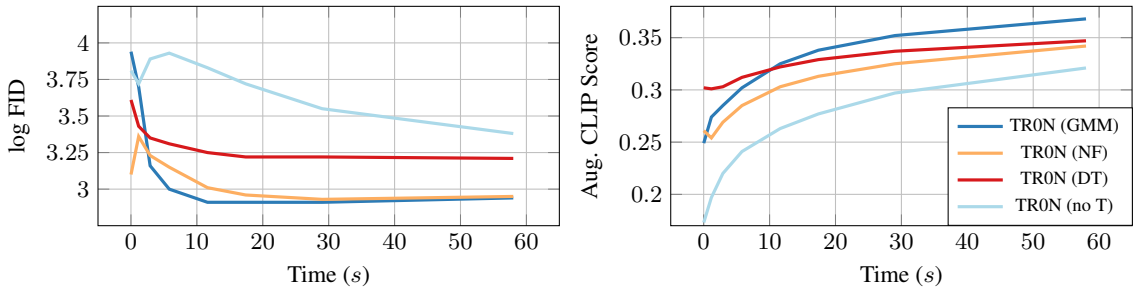
\begin{figure}[t!]
    \centering
    \begin{tikzpicture}
\definecolor{red}{RGB}{215,25,28}
\definecolor{orange}{RGB}{253,174,97}
\definecolor{cyan}{RGB}{171,217,233}
\definecolor{blue}{RGB}{44,123,182}
\footnotesize
\begin{axis}[
    height=4.5cm,
    width = 0.45\linewidth,
    xlabel near ticks,
    xlabel={Time ($s$)},
    ylabel near ticks,
    ylabel={log FID},
    xmin=-5, xmax=65,
    ymin=2.75, ymax=4.25,
    xtick={0, 10, 20, 30, 40, 50, 60},
    ytick={3, 3.25, 3.5, 3.75, 4},
    legend pos=north east,
    ymajorgrids=true,
    xmajorgrids=true,
    every axis plot/.append style={very thick}
]

% %TR0N w caption
% \addplot[
%     color=blue,
%     ]
%     coordinates {
%     (0,3.94)(2,3.65)(5,3.06)(10,2.85)(20,2.77)(30,2.77)(50,2.77)(100,2.82)
%     };
% %TR0N without caption
% \addplot[
%     color=red,
%     ]
%     coordinates {
%     (0,3.80)(2,3.40)(5,3.01)(10,2.87)(20,2.83)(30,2.84)(50,2.86)(100,2.89)
%     };
% %TR0N no translator
% \addplot[
%     color=cyan,
%     ]
%     coordinates {
%     (0,3.77)(2,3.66)(5,3.84)(10,3.88)(20,3.78)(30,3.65)(50,3.47)(100, 3.28)
%     };
% %FuseDream
% \addplot[
%     color=orange,
%     ]
%     coordinates {
%     (40, 4.05)(42,3.51)(45,3.19)(50,3.00)(60,2.90)(70,2.87)(90,2.87)(140,2.90)
%     };

% \legend{TR0N (w/ caption), TR0N (w/o caption), TRON (no T), FuseDream}

%TR0N-GMM
\addplot[
    color=blue,
    ]
    coordinates {
    (0.01,3.94)(1.17,3.70)(2.91,3.16)(5.81,3.00)(11.61,2.91)(17.41,2.91)(29.01,2.91)(58.01,2.94)
    };
%TR0N-NF
\addplot[
    color=orange,
    ]
    coordinates {
    (0.01,3.10)(1.17,3.36)(2.91,3.23)(5.81,3.15)(11.61,3.01)(17.41,2.96)(29.01,2.93)(58.01,2.95)
    };
%TR0N-Det
\addplot[
    color=red,
    ]
    coordinates {
    (0.01,3.61)(1.17,3.43)(2.91,3.35)(5.81,3.31)(11.61,3.25)(17.41,3.22)(29.01,3.22)(58.01,3.21)
    };
%TR0N-noT-1
\addplot[
    color=cyan,
    ]
    coordinates {
    (0.01,3.81)(1.17,3.72)(2.91,3.89)(5.81,3.93)(11.61,3.83)(17.41,3.72)(29.01,3.55)(58.01, 3.38)
    };

\end{axis}
\end{tikzpicture}
    \begin{tikzpicture}
\definecolor{red}{RGB}{215,25,28}
\definecolor{orange}{RGB}{253,174,97}
\definecolor{cyan}{RGB}{171,217,233}
\definecolor{blue}{RGB}{44,123,182}
\footnotesize
\begin{axis}[
    height=4.5cm,
    width=0.45\linewidth,
    xlabel near ticks,
    xlabel={Time ($s$)},
    ylabel near ticks,
    ylabel={Aug. CLIP Score},
    xmin=-5, xmax=65,
    ymin=0.17, ymax=0.38,
    xtick={0, 10, 20, 30, 40, 50, 60},
    ytick={0.20, 0.25, 0.30, 0.35},
    legend pos=south east,
    ymajorgrids=true,
    xmajorgrids=true,
    legend style={font=\scriptsize, xshift=0.195cm, yshift=-0.09cm},
    every axis plot/.append style={very thick}
]

%TR0N-GMM
\addplot[
    color=blue,
    ]
    coordinates {
    (0.01,0.249)(1.17,0.274)(2.91,0.285)(5.81,0.302)(11.61,0.325)(17.41,0.338)(29.01,0.352)(58.01,0.368)
    };
%TR0N-NF
\addplot[
    color=orange,
    ]
    coordinates {
    (0.01,0.261)(1.17,0.254)(2.91,0.269)(5.81,0.285)(11.61,0.303)(17.41,0.313)(29.01,0.325)(58.01,0.342)
    };
%TR0N-Det
\addplot[
    color=red,
    ]
    coordinates {
    (0.01,0.302)(1.17,0.301)(2.91,0.303)(5.81,0.312)(11.61,0.322)(17.41,0.329)(29.01,0.337)(58.01,0.347)
    };
%TR0N-noT-1
\addplot[
    color=cyan,
    ]
    coordinates {
    (0.01,0.173)(1.17,0.197)(2.91,0.220)(5.81,0.241)(11.61,0.263)(17.41,0.277)(29.01,0.297)(58.01, 0.321)
    };
    
\legend{TR0N (GMM), TR0N (NF), TR0N (DT), TR0N (no T)}

%FuseDream
% \addplot[
%     color=orange,
%     ]
%     coordinates {
%     (28.11, 4.05)(29.27,3.51)(31.01,3.19)(33.91,3.00)(39.71,2.90)(45.51,2.87)(57.11,2.87)(85.81,2.90)
% %TR0N w caption
% \addplot[
%     color=blue,
%     ]
%     coordinates {
%     (0.01,0.25)(1.17,0.27)(2.91,0.28)(5.81,0.30)(11.61,0.33)(17.41,0.34)(29.01,0.35)(58.01,0.37)
%     };
% %TR0N without caption
% \addplot[
%     color=red,
%     ]
%     coordinates {
%     (0.01,0.25)(1.17,0.28)(2.91,0.29)(5.81,0.31)(11.61,0.33)(17.41,0.34)(29.01,0.35)(58.01,0.37)
%     };
% %TR0N no translator
% \addplot[
%     color=cyan,
%     ]
%     coordinates {
%     (0.01,0.17)(1.17,0.20)(2.91,0.22)(5.81,0.24)(11.61,0.26)(17.41,0.28)(29.01,0.30)(58.01,0.32)
%     };
% %FuseDream
% \addplot[
% color=orange,
% ]
% coordinates {
% (28.11, 0.23)(29.27, 0.26)(31.01,0.28)(33.91,0.30)(39.71,0.32)(45.51,0.33)(57.11,0.35)(85.81,0.36)
% };
% % \legend{TR0N (w/ caption), TR0N (w/o caption), FuseDream, TRON (no T)}

\end{axis}
\end{tikzpicture}
    % \\[-0.3cm]
    \caption{Ablations of translator choice for TR0N:BigGAN+CLIP (BLIP) on MS-COCO.}
    \label{fig:ablations}
\end{figure}

As mentioned in the main manuscript, we carry out an ablation over different translators so as to empirically justify our choices. \autoref{fig:ablations} shows results of TR0N using BigGAN for text-to-image generation on MS-COCO for the GMM translator that we use on the main text, a normalizing-flow translator (marked as ``NF''), a deterministic translator (``DT''), and no translator (``no T'', as also shown in the main manuscript). We use the choice of $f$ with a caption model for all translators. Due to computational constraints, we only use $10k$ generated samples to compute metrics (this change accounts for any difference with the numbers shown in the main manuscript). We can see that only the NF translator matches the GMM one in terms of FID score, but only for larger time budgets; while no method beats the GMM translator in augmented CLIP score. Once again, these results confirm our choices for the translator.

Finally, we make a note about the preliminary experiments that we mentioned in the main manuscript using Stein discrepancy to train the translator: we did not manage to stabilize training, and the obtained FIDs were an order of magnitude greater than those of the GMM translator.

\subsection{Additional samples}

\begin{figure*} [t!]
\centering
\fontsize{7.5}{9}
\selectfont
\hspace*{-0.3cm}
\begin{tabular}
{c}
   \includegraphics[width=0.98\textwidth]{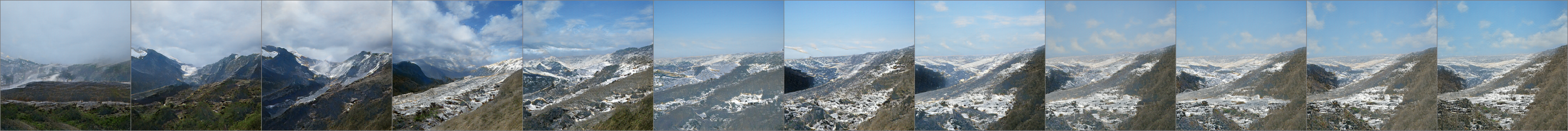} \\[-0.05cm]
   A photo of a snowy valley \\[0.1cm]
   \includegraphics[width=0.98\textwidth]{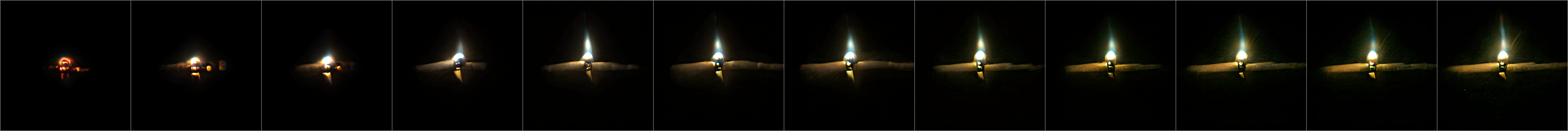} \\[-0.05cm]
   A light shining in the night \\[0.1cm]
   \includegraphics[width=0.98\textwidth]{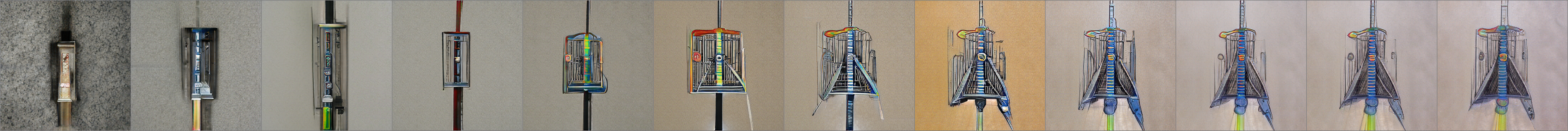} \\[-0.05cm]
   A crayon drawing of a space elevator
\end{tabular}
\caption{Evolution of samples throughout Langevin dynamics for our TR0N:BigGAN+CLIP (BLIP) model. Each row shows $G(z^{(t)})$ for increasing values of $t \in \{0, 15, 30, 45, 60, 75, 90, 140, 210, 280, 350, 500\}$ for the corresponding text prompt.}
\label{fig:langevin_evolution}
\end{figure*}

% \begin{figure}[t!]
%     \centering
%     \input{figs/FailureCase/failurecase.tex}
%     \caption{Remember to say if this is from BigGAN or StyleGAN-XL}
%     \label{fig:tr0n_failure}
% \end{figure}

% \begin{figure}[t!]
%     \centering
%     \includegraphics{figs/layer6_logo.png}
%     \caption{Uncurated samples from our TR0N:StyleGAN-XL+LAION2BCLIP (BLIP-2) model. Text prompts are randomly sampled from MS-COCO.}
%     \label{fig:stylganxl_samples1}
% \end{figure}

% \begin{figure}[t!]
%     \centering
%     \includegraphics{figs/layer6_logo.png}
%     \caption{Samples from our TR0N:StyleGAN-XL+LAION2BCLIP (BLIP-2) model.}
%     \label{fig:stylganxl_samples2}
% \end{figure}

\autoref{fig:langevin_evolution} shows how samples evolve throughout Langevin dynamics. We can see that the output of the translator provides a sensible initialization, and that image quality improves throughout Langevin dynamics. Not surprisingly, the output of the translator more closely matches the given text prompt for in-distribution prompts (e.g.\ ``A photo of a snowy valley'') than for out-of-distribution prompts (e.g.\ ``A crayon drawing of a space elevator''), although the initialization remains useful for all situations.

\begin{figure*} [t!]
\centering
\fontsize{7.5}{9}
\selectfont
\hspace*{-0.3cm}
\begin{tabular}
{p{0.11\linewidth}p{0.11\linewidth}p{0.11\linewidth}p{0.11\linewidth}p{0.11\linewidth}p{0.11\linewidth}}
   \includegraphics[width=2.25cm]{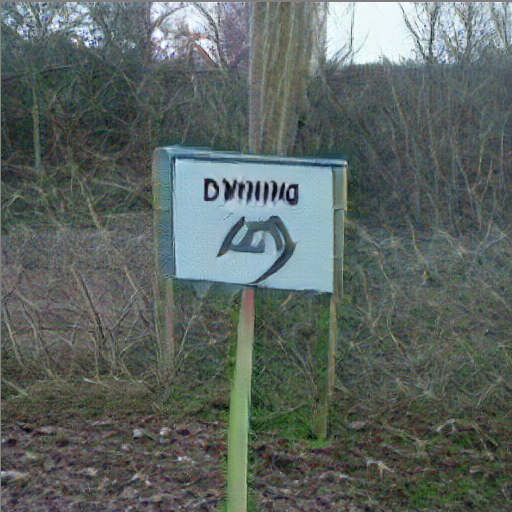} & \includegraphics[width=2.25cm]{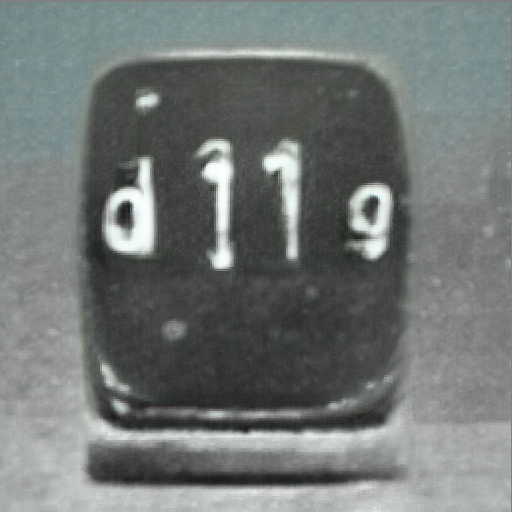} & \includegraphics[width=2.25cm]{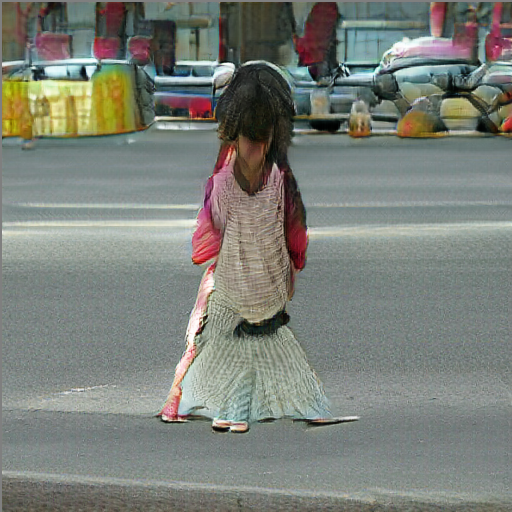} & \includegraphics[width=2.25cm]{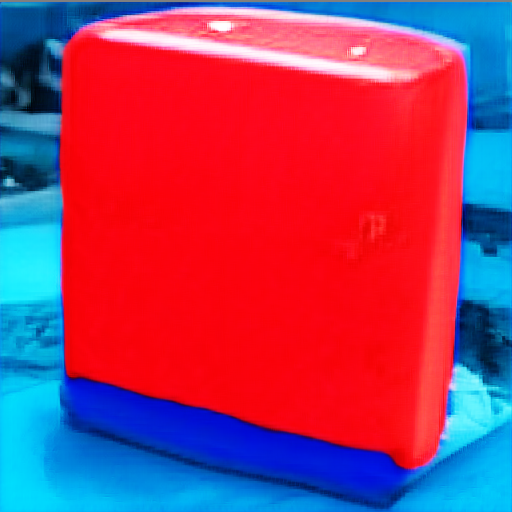} & \includegraphics[width=2.25cm]{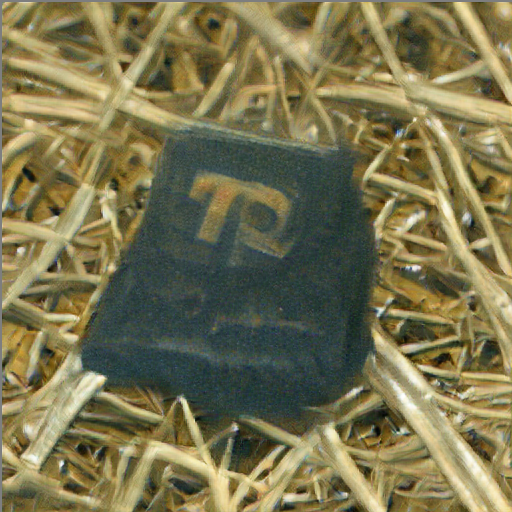}& \includegraphics[width=2.25cm]{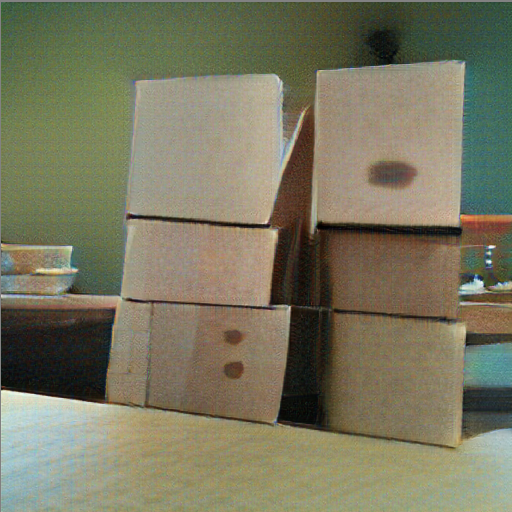}\\[-0.05cm]
   A sign that says `Deep Learning' & A photo of digit number 5 & A photo of a girl walking on the street & A red cube on top of a blue cube & A photo of a Tcennis rpacket & Four boxes on the table
\end{tabular}
\caption{Failure cases from our TR0N:BigGAN+CLIP (BLIP) model.}
\label{fig:tr0n_failure}
\end{figure*}

\autoref{fig:tr0n_failure} shows text prompts for which TR0N fails. In particular, we find that TR0N struggles to produce images which involve: written text (e.g.\ ``A sign that says `Deep Learning' '' or ``A photo of the digit number 5''), humans (e.g.\ ``A photo of a girl walking on the street''), understanding the relationship between various objects (e.g.\ ``A red cube on top of a blue cube''), misspelled text prompts (e.g.\ ``A photo of a Tcennis rpacket''), or counting (e.g.\ ``Four boxes on the table'').

\begin{figure*} [t!]
\centering
\fontsize{7.5}{9}
\selectfont
% \hspace*{-0.8cm}
\begin{tabular}
{p{0.095\linewidth}p{0.095\linewidth}p{0.095\linewidth}p{0.095\linewidth}p{0.095\linewidth}p{0.095\linewidth}p{0.095\linewidth}p{0.095\linewidth}}
   \includegraphics[width=2.cm]{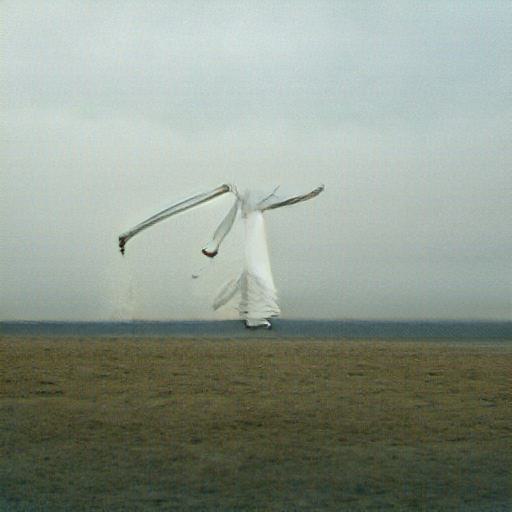} & \includegraphics[width=2.cm]{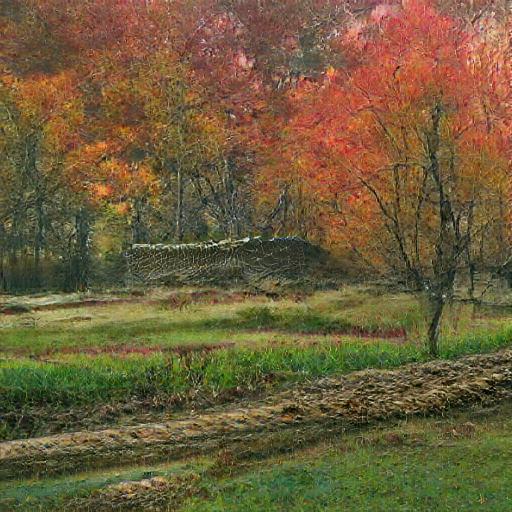} & \includegraphics[width=2.cm]{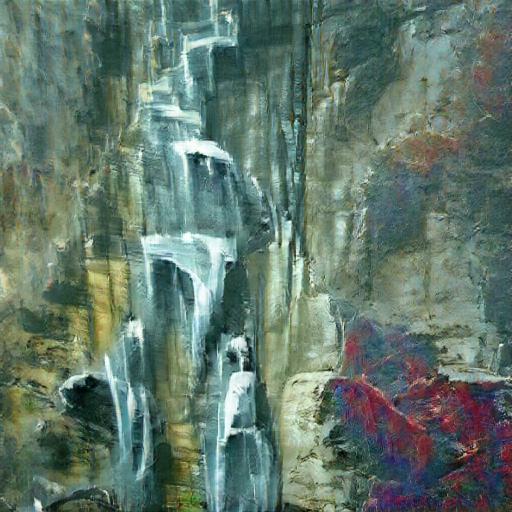} & \includegraphics[width=2.cm]{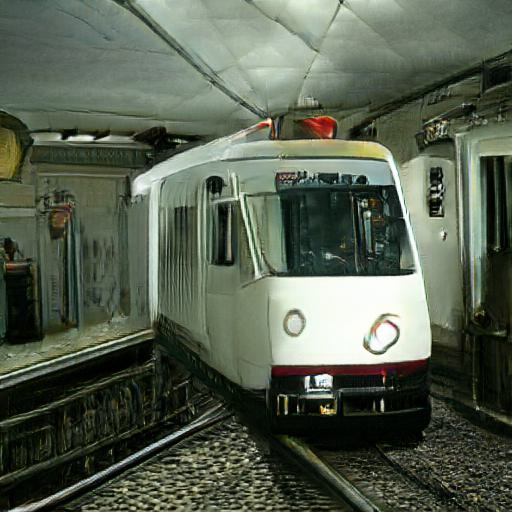} & \includegraphics[width=2.cm]{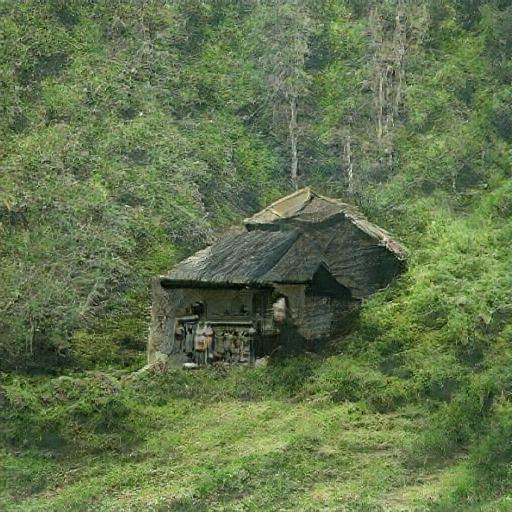} & \includegraphics[width=2.cm]{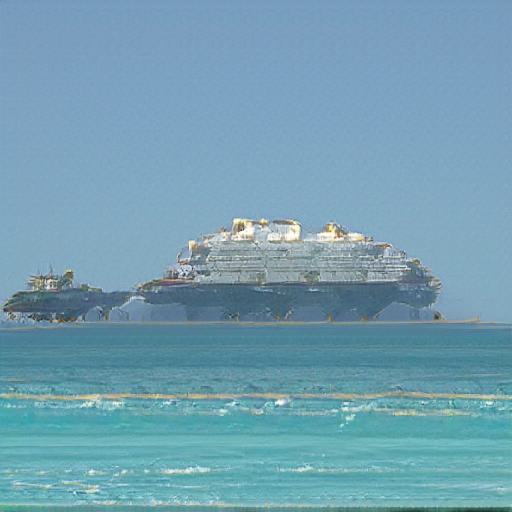} & \includegraphics[width=2.cm]{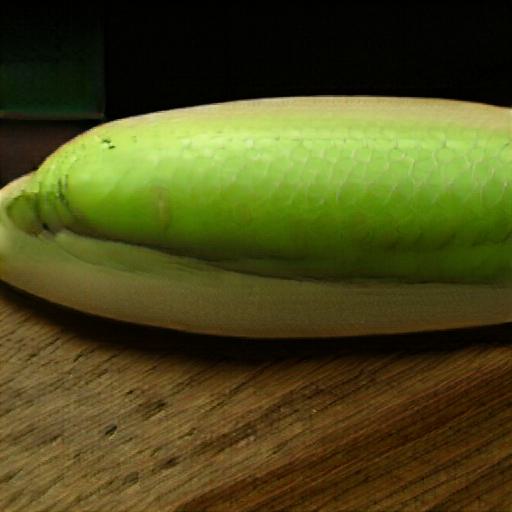}& \includegraphics[width=2.cm]{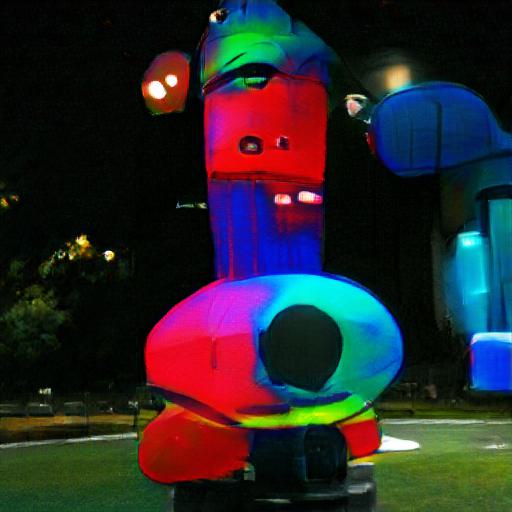}\\
   \includegraphics[width=2.cm]{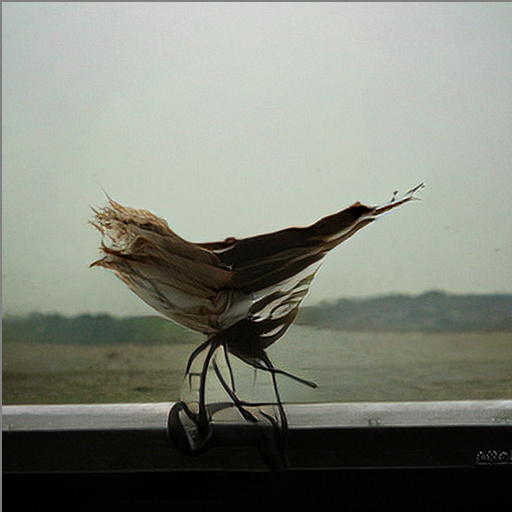} & \includegraphics[width=2.cm]{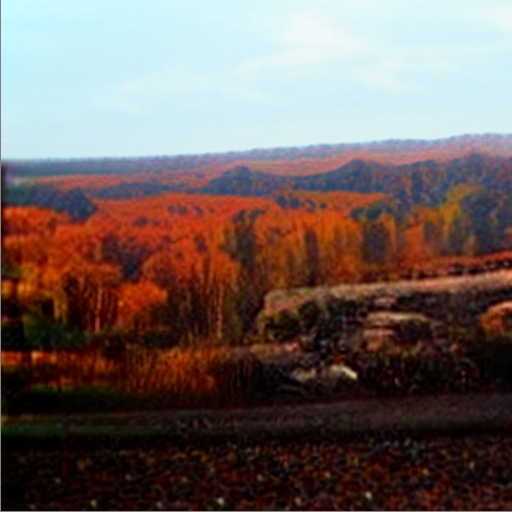} & \includegraphics[width=2.cm]{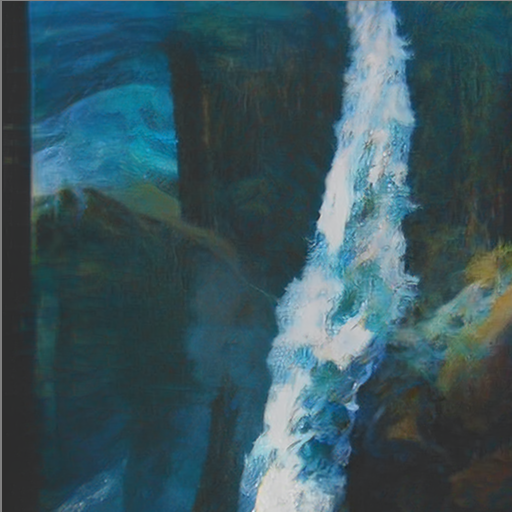} & \includegraphics[width=2.cm]{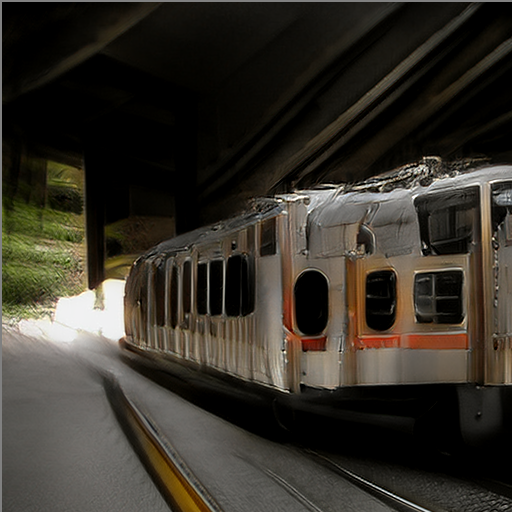} & \includegraphics[width=2.cm]{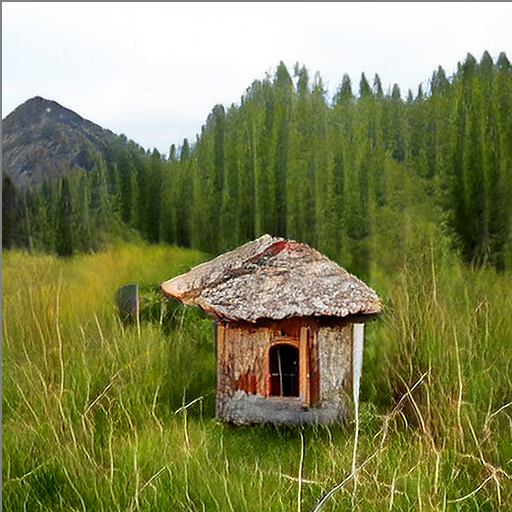} & \includegraphics[width=2.cm]{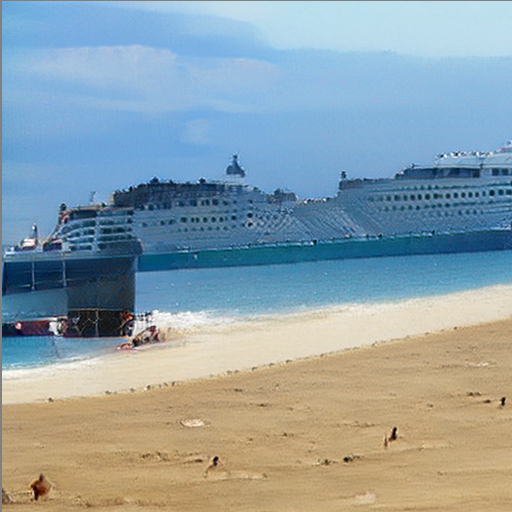} & \includegraphics[width=2.cm]{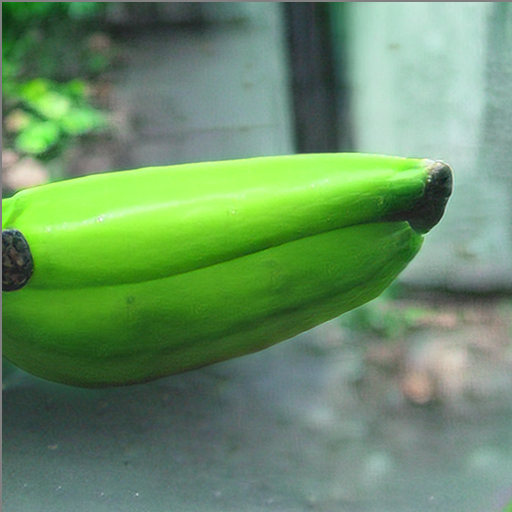}& \includegraphics[width=2.cm]{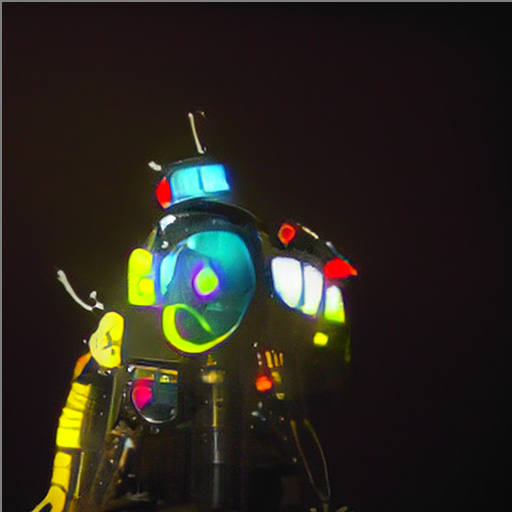}\\[-0.05cm]
   When the wind blows & A photo of a landscape in fall & An abstract painting of a waterfall & \hspace{0.035cm}A subway train coming out of a tunnel & \hspace{0.035cm}A small house in the wilderness & A beach with a cruise ship passing by & A green colored banana & A colorful robot under moonlight \\[0.25cm] 
   \includegraphics[width=2.cm]{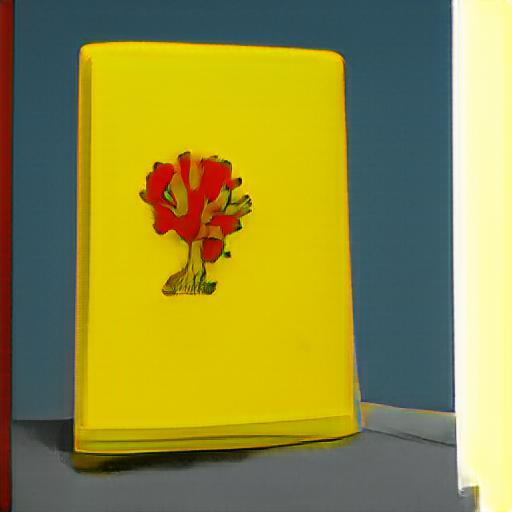} & \includegraphics[width=2.cm]{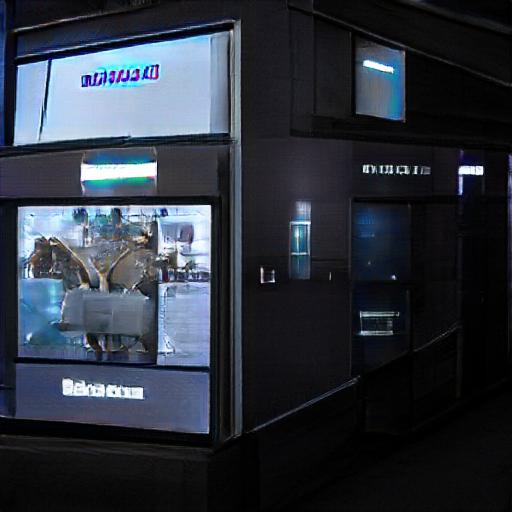}& \includegraphics[width=2.cm]{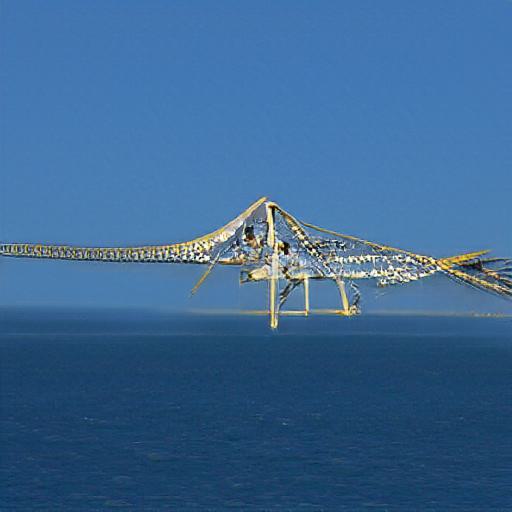}  & \includegraphics[width=2.cm]{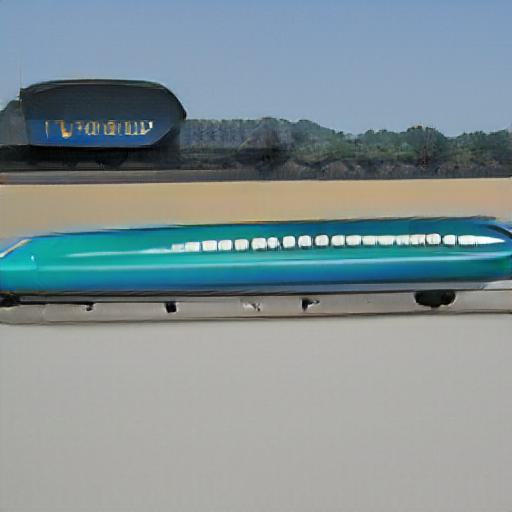} & \includegraphics[width=2.cm]{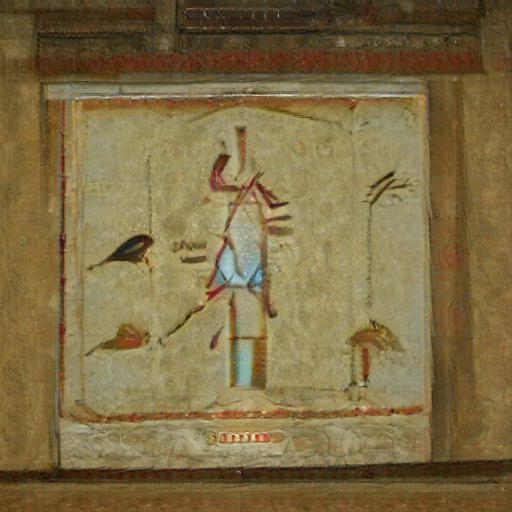} & \includegraphics[width=2.cm]{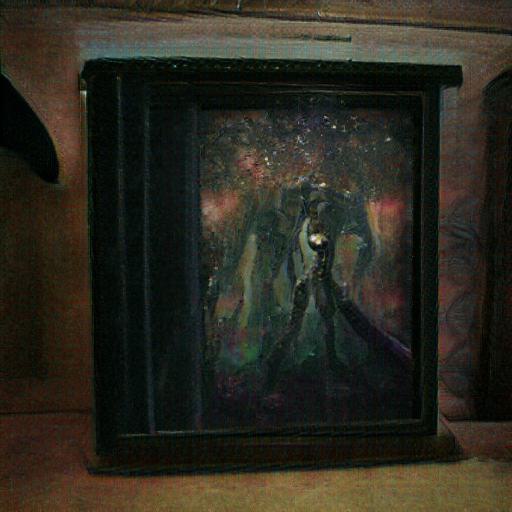} & \includegraphics[width=2.cm]{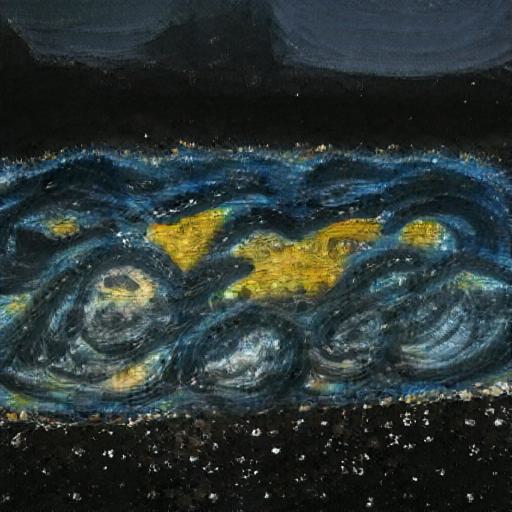}& \includegraphics[width=2.cm]{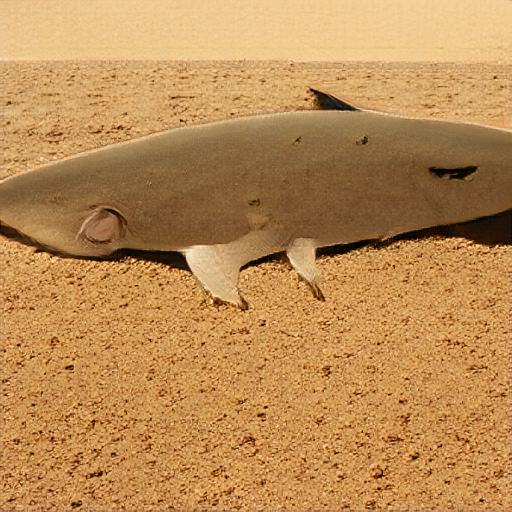}\\
   \includegraphics[width=2.cm]{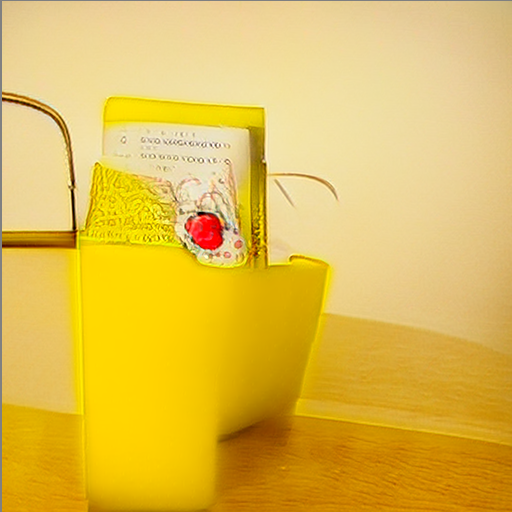} & \includegraphics[width=2.cm]{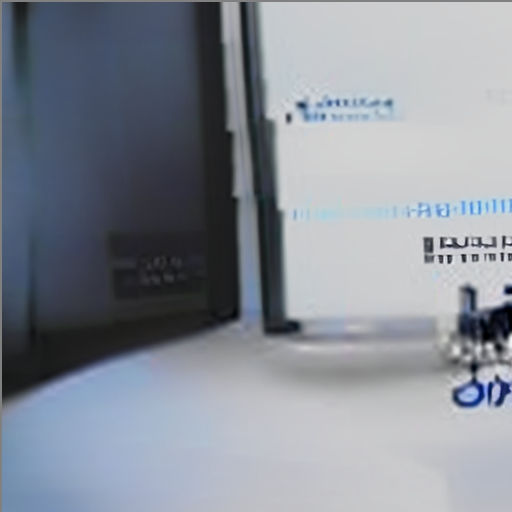}& \includegraphics[width=2.cm]{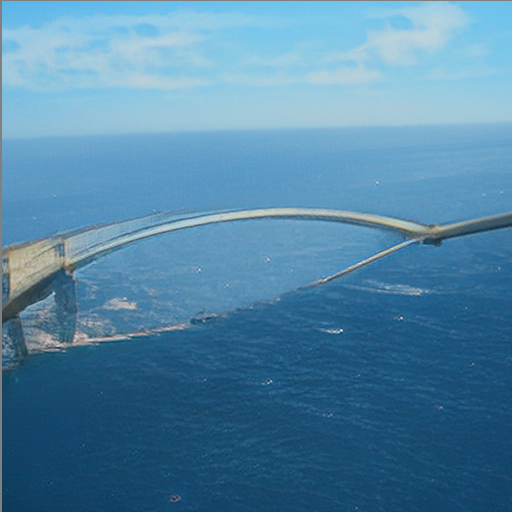}  & \includegraphics[width=2.cm]{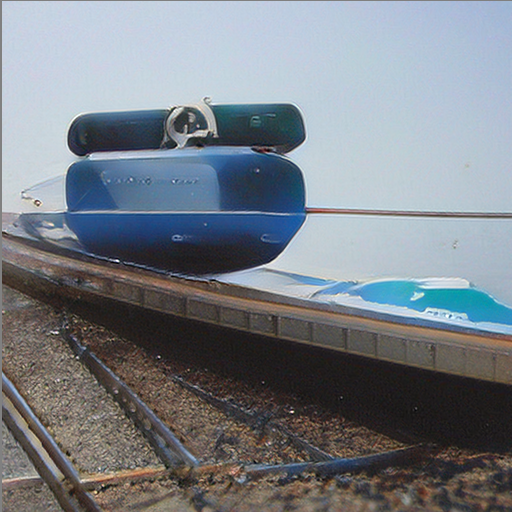} & \includegraphics[width=2.cm]{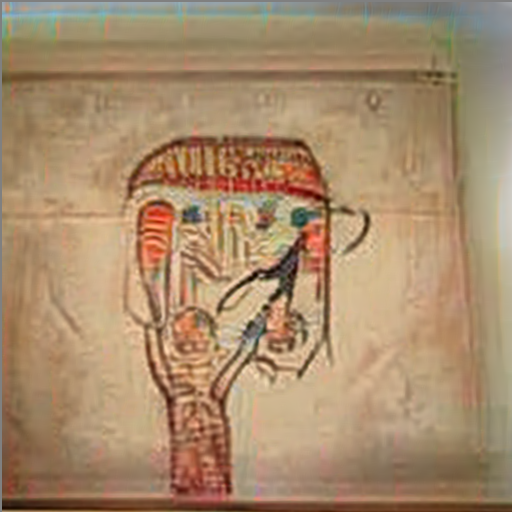} & \includegraphics[width=2.cm]{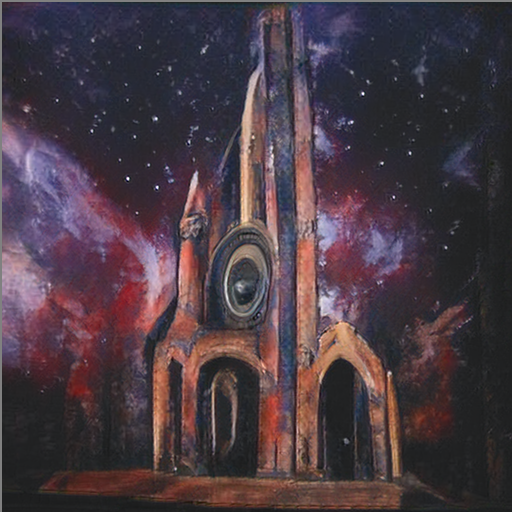} & \includegraphics[width=2.cm]{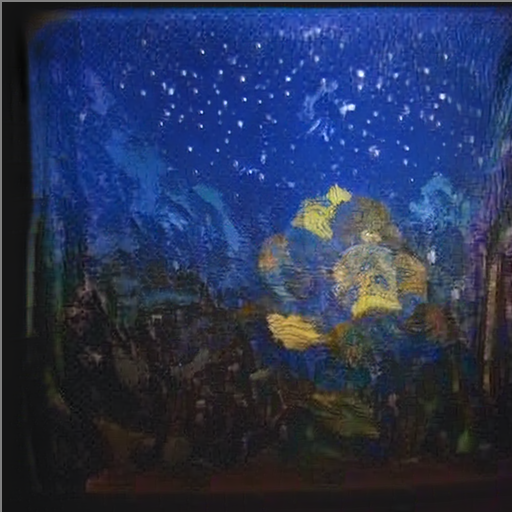}& \includegraphics[width=2.cm]{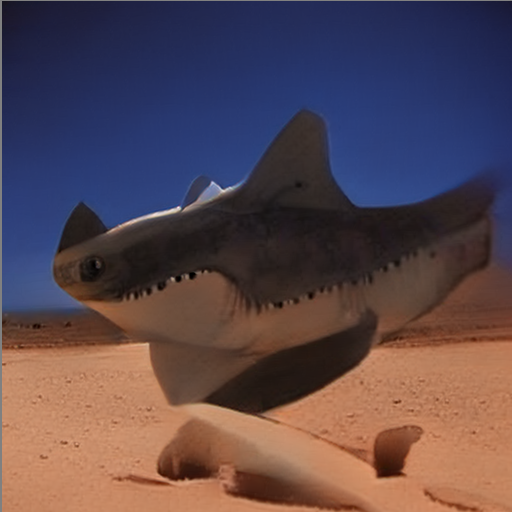}\\[-0.05cm]
  A yellow book and a red vase & An American multinational technology company that focuses on artificial intelligence & A bridge connecting Europe and North America on the Atlantic Ocean, bird's eye view & \hspace{0.035cm}A train on top of a surfboard & An ancient Egyptian painting & \hspace{0.035cm}A painting an outer space, gothic style & A painting of a Starry Night & A shark in the desert \\[-0.25cm]

\end{tabular}
\caption{Uncurated samples from our TR0N:BigGAN+CLIP (BLIP) model (\textbf{first and third image rows}) and from our TR0N:StyleGAN-XL+LAION2BCLIP (BLIP2) model (\textbf{second and fourth image rows}).}
\label{fig:biggan_uncurated_appendix}
\end{figure*}

\autoref{fig:biggan_uncurated_appendix} shows uncurated samples from our TR0N:BigGAN+CLIP (BLIP) and TR0N:StyleGAN-XL+LAION2BCLIP (BLIP2) models. We use out-of-distribution text-prompts we collected from various sources (the entire list of text prompts -- which we randomly subsampled to produce the figure -- along with where we obtained each prompt from, is included with our code). Note that we did not curate the particular images shown in the main manuscript, although we did select text prompts for which TR0N produced good results. Although the images from the StyleGAN-XL-based model look a bit sharper (likely explaining the improved FID of this model), we find it hard to conclude than one model is consistenly better than the other. In particular, the StyleGAN-XL-based model seems to be slightly worse at conditioning (e.g.\ the images it produces with the same prompts as in \autoref{fig:fig1} and \autoref{fig:TR0N_diversity} are worse than those from the BigGAN-based model). We hypothesize that using the style space $\mathcal{W}$ as the latent space for our StyleGAN-XL-based model might bias TR0N towards focusing more on style than matching the semantic content of the given condition.

\begin{figure*} [t!]
\centering
\fontsize{7.5}{9}
\selectfont
% \hspace*{-0.65cm}
\begin{tabular}
{p{0.095\linewidth}p{0.095\linewidth}p{0.095\linewidth}p{0.095\linewidth}p{0.095\linewidth}p{0.095\linewidth}p{0.095\linewidth}p{0.095\linewidth}}
   \includegraphics[width=2.cm]{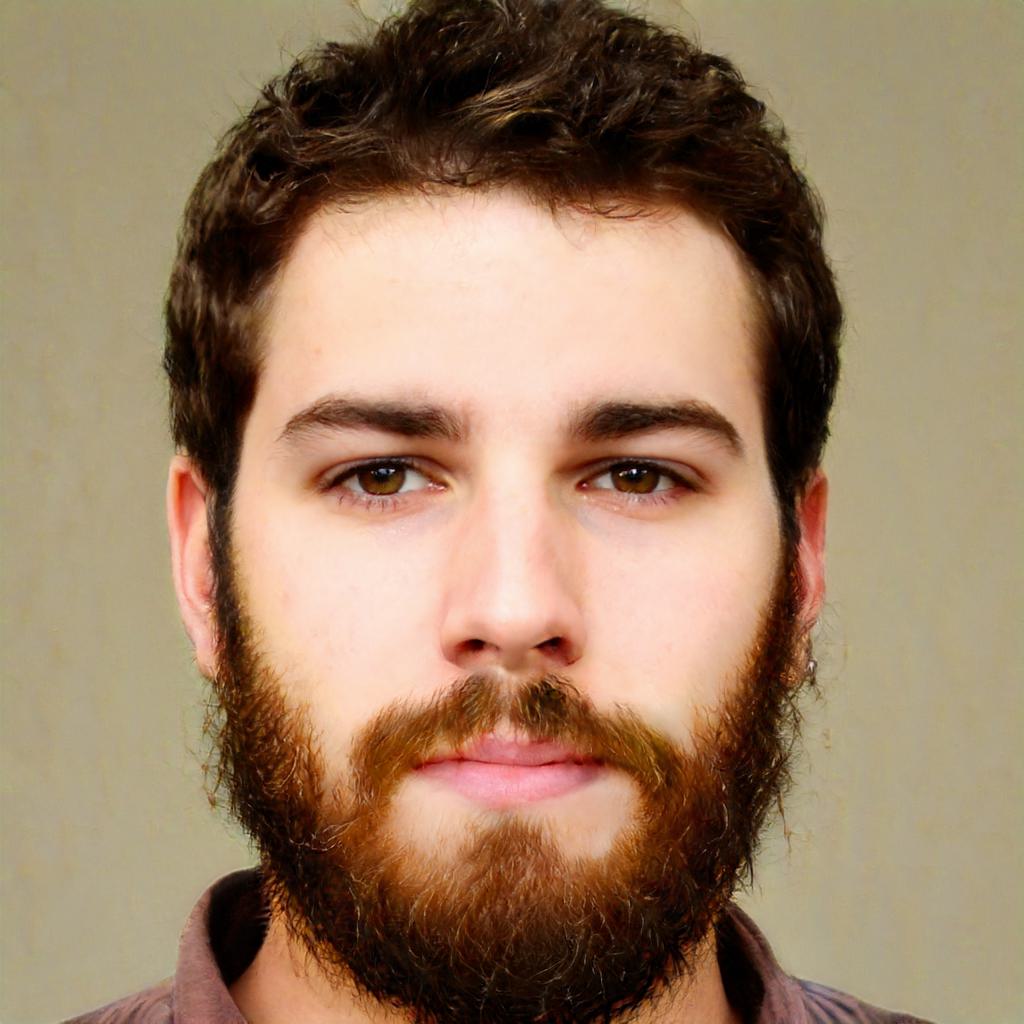} & \includegraphics[width=2.cm]{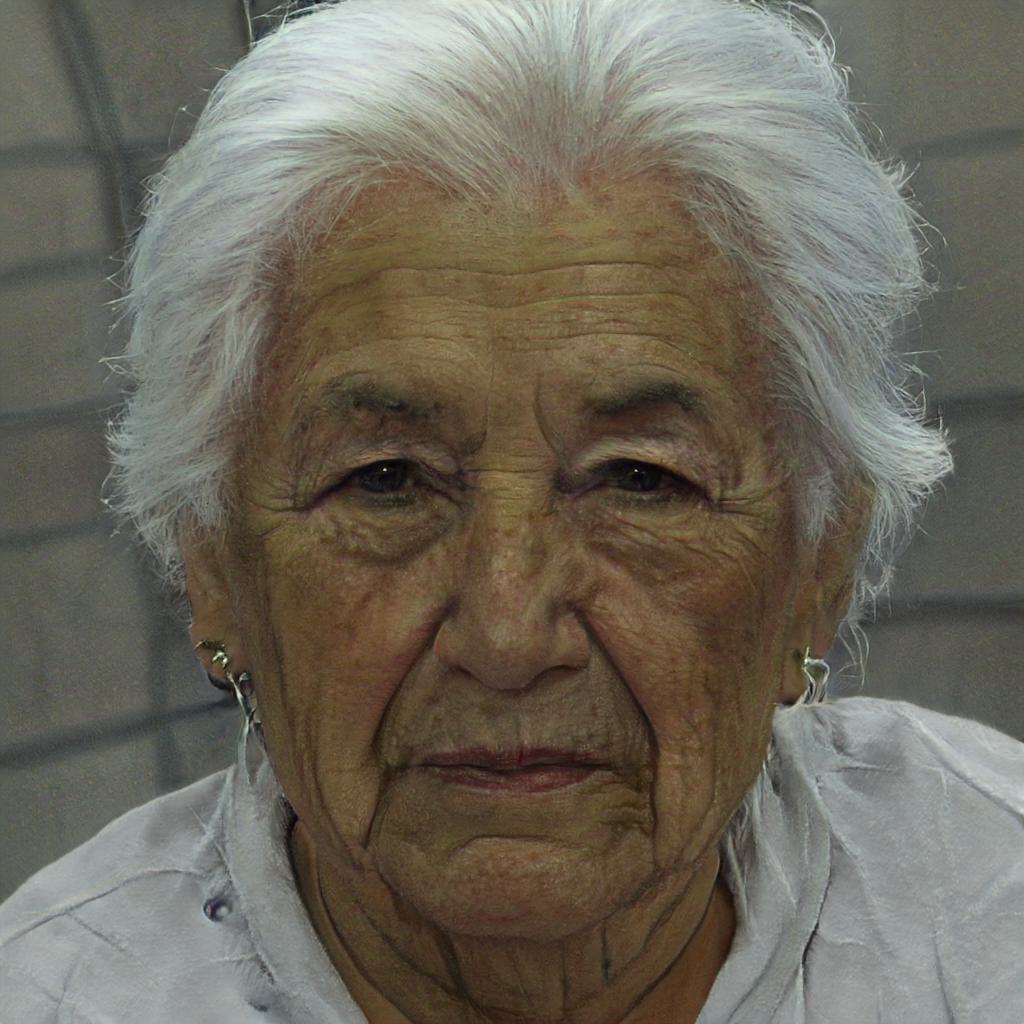} & \includegraphics[width=2.cm]{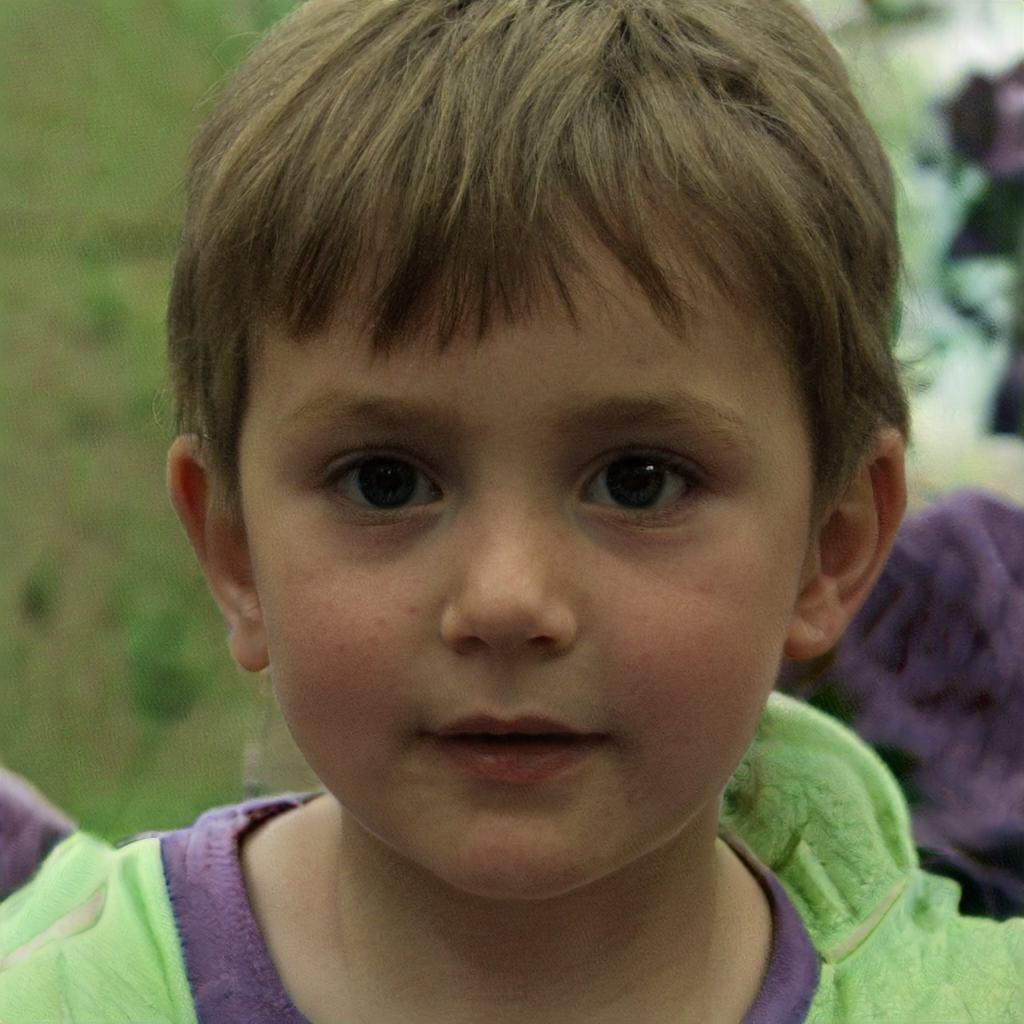} & \includegraphics[width=2.cm]{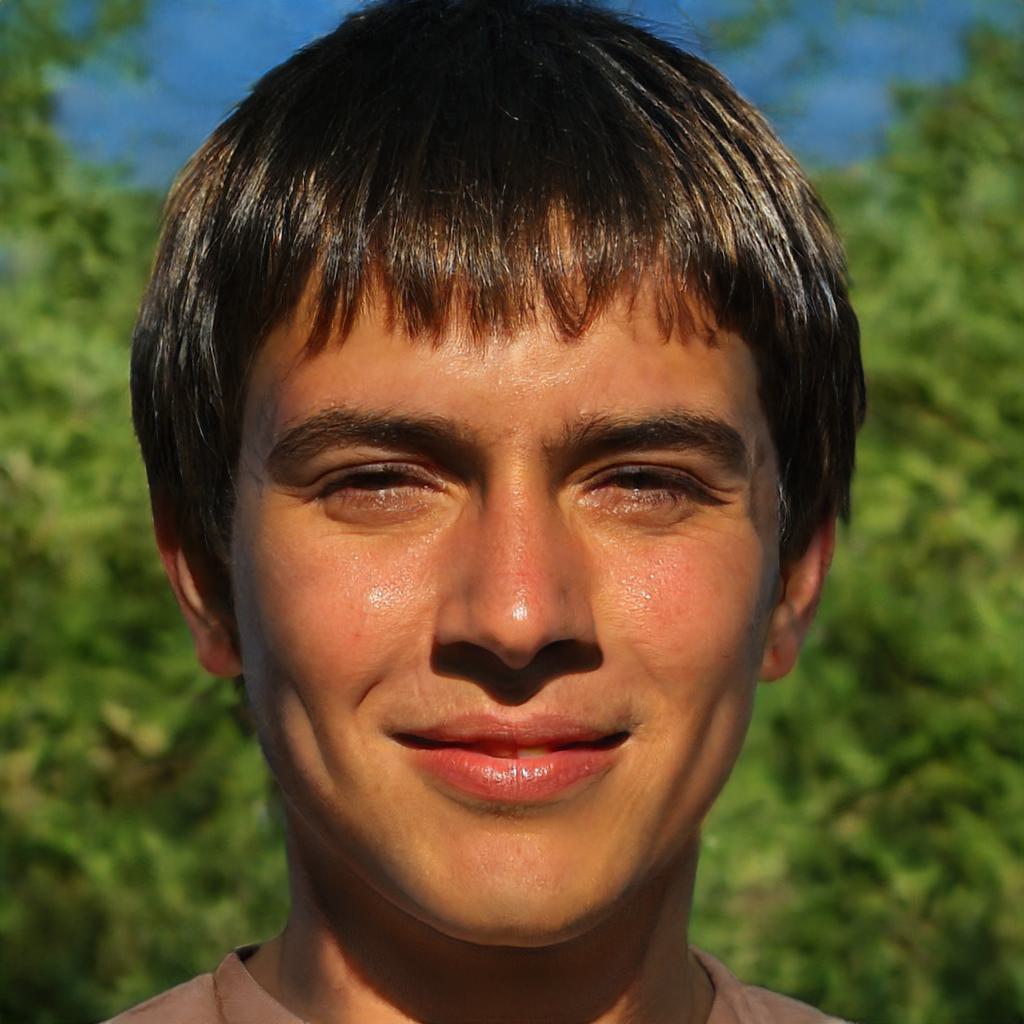} & \includegraphics[width=2.cm]{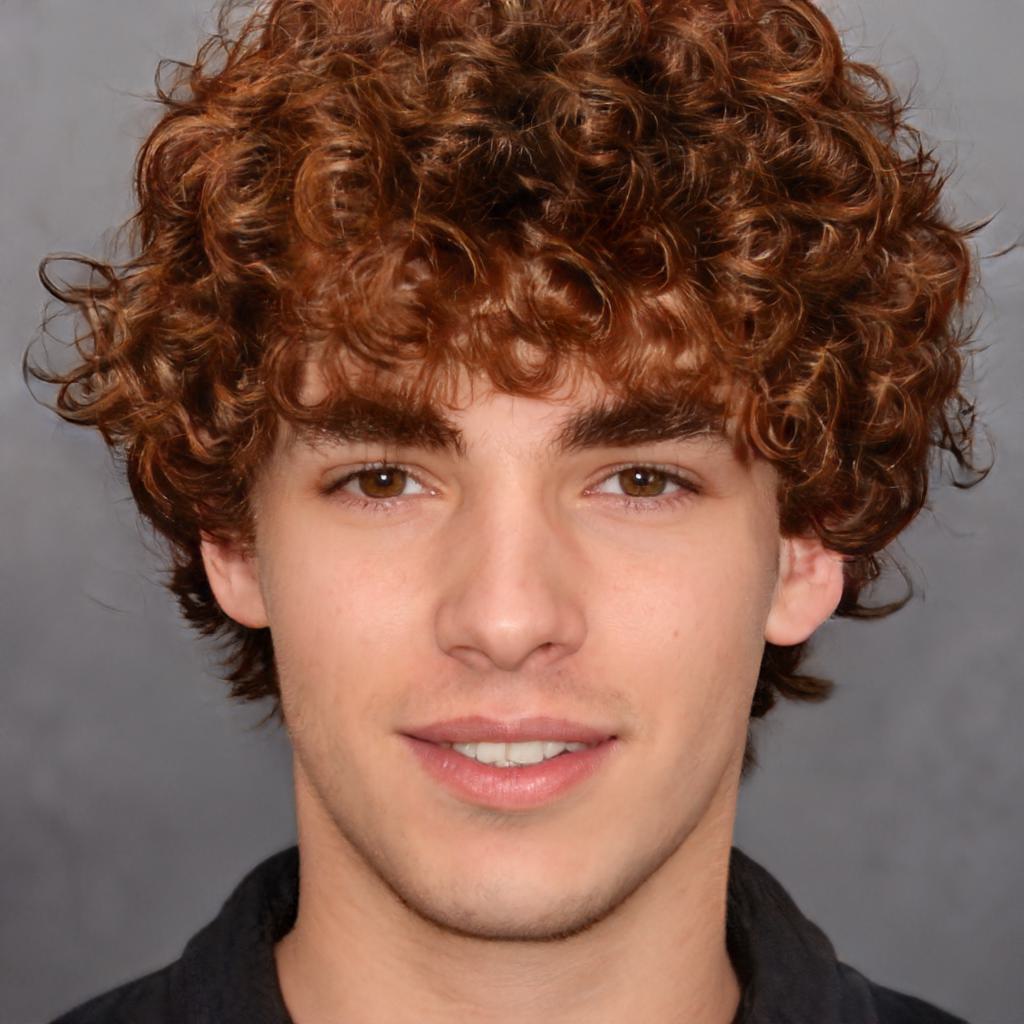} & \includegraphics[width=2.cm]{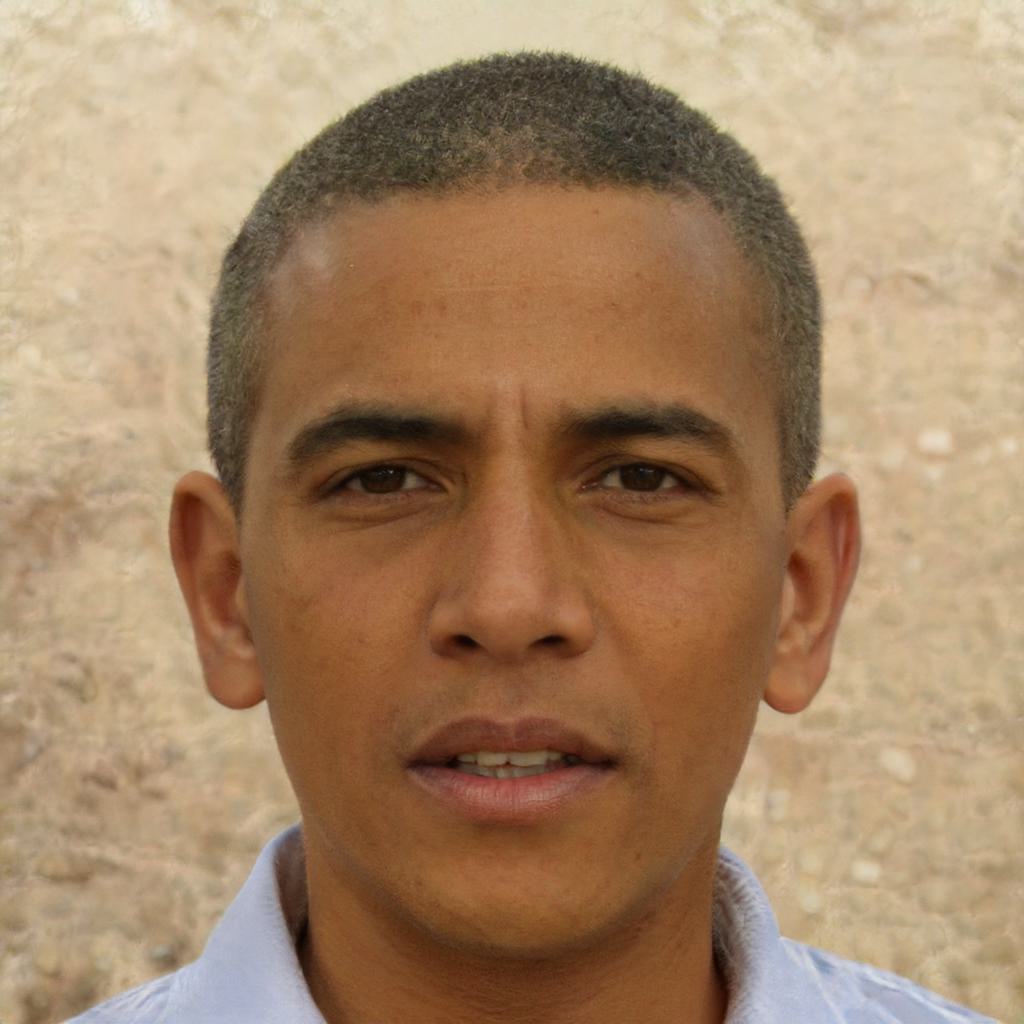} & \includegraphics[width=2.cm]{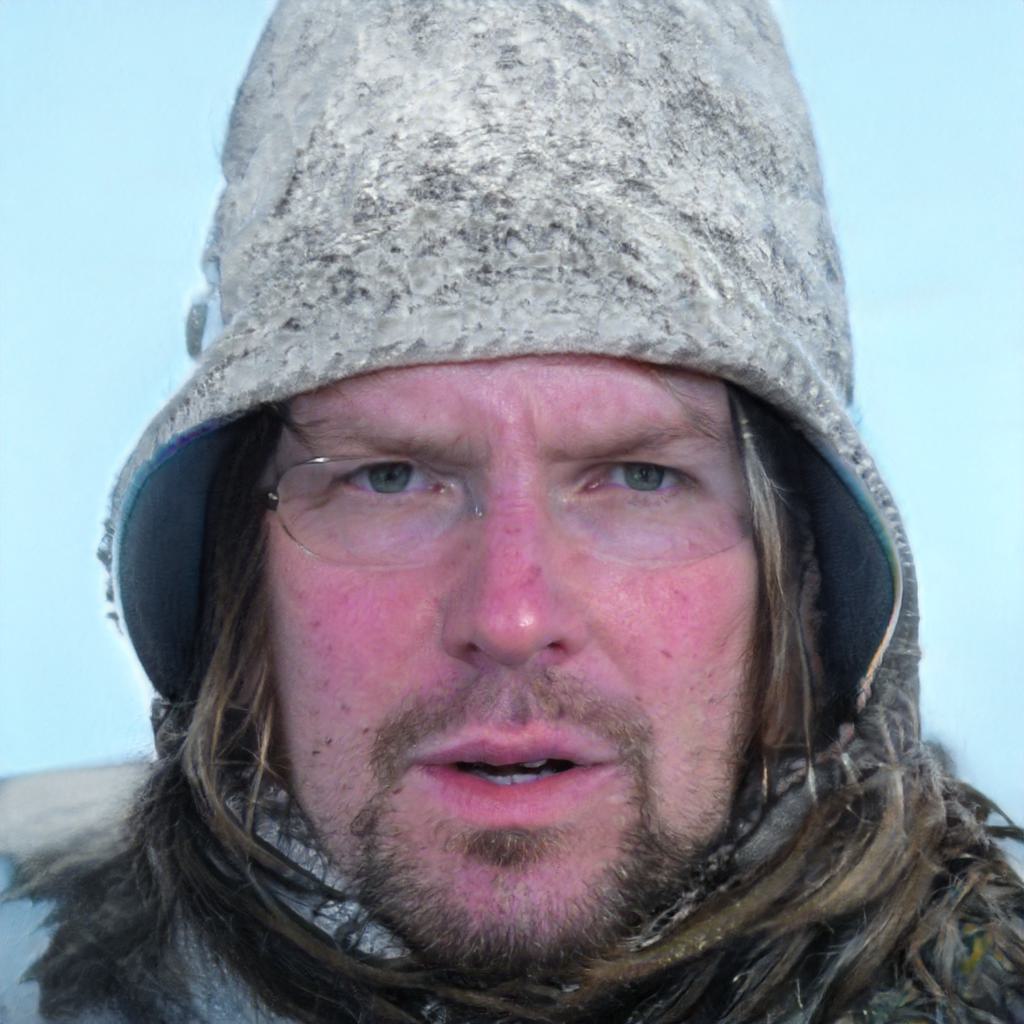}& \includegraphics[width=2.cm]{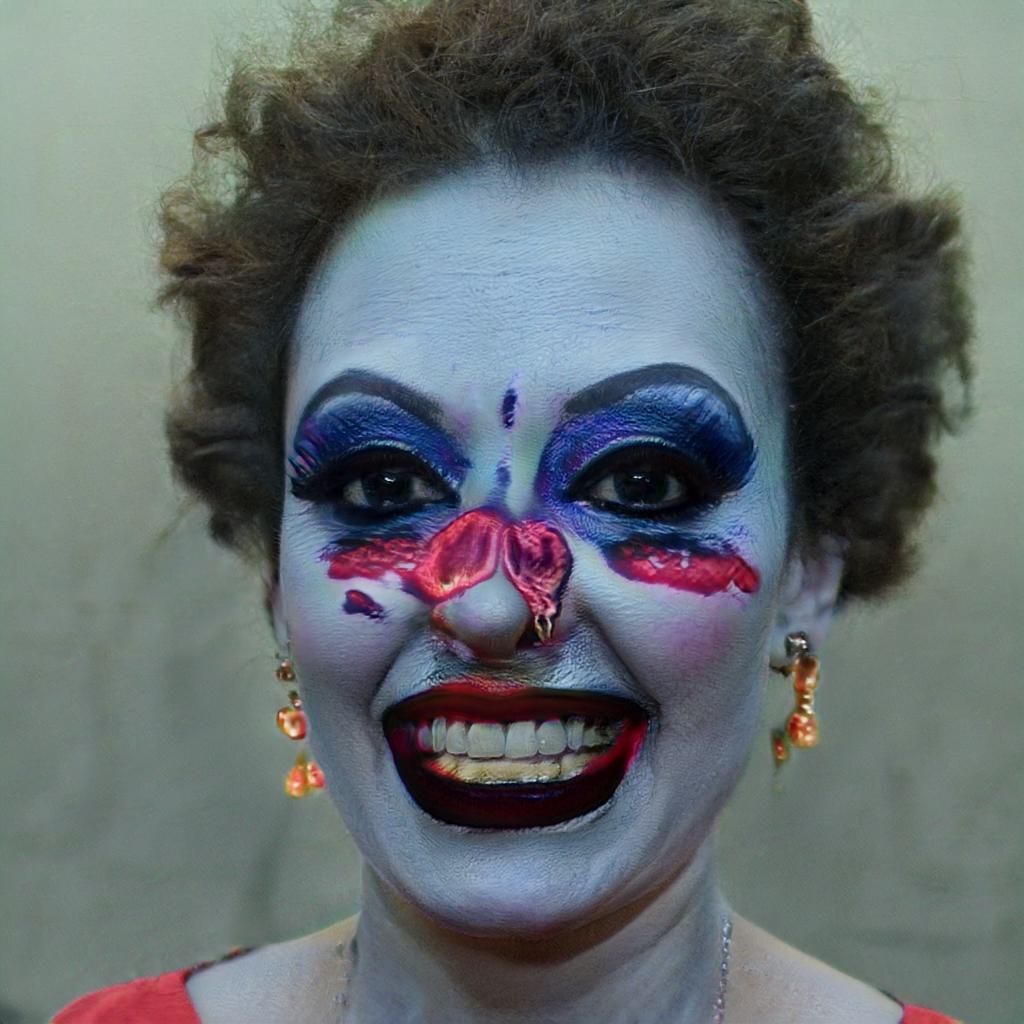}\\[-0.05cm]
   A photograph of a young man with a beard & A photograph of a old woman's face with grey hair & A photograph of a child at a birthday party & \hspace{0.035cm}A picture of a face outside in bright sun in front of green grass & \hspace{0.035cm}This man has bangs arched eyebrows curly hair and a small nose & President Barack Obama & An arctic explorer & A clown's face covered in make up \\[0.25cm] 
   \includegraphics[width=2.cm]{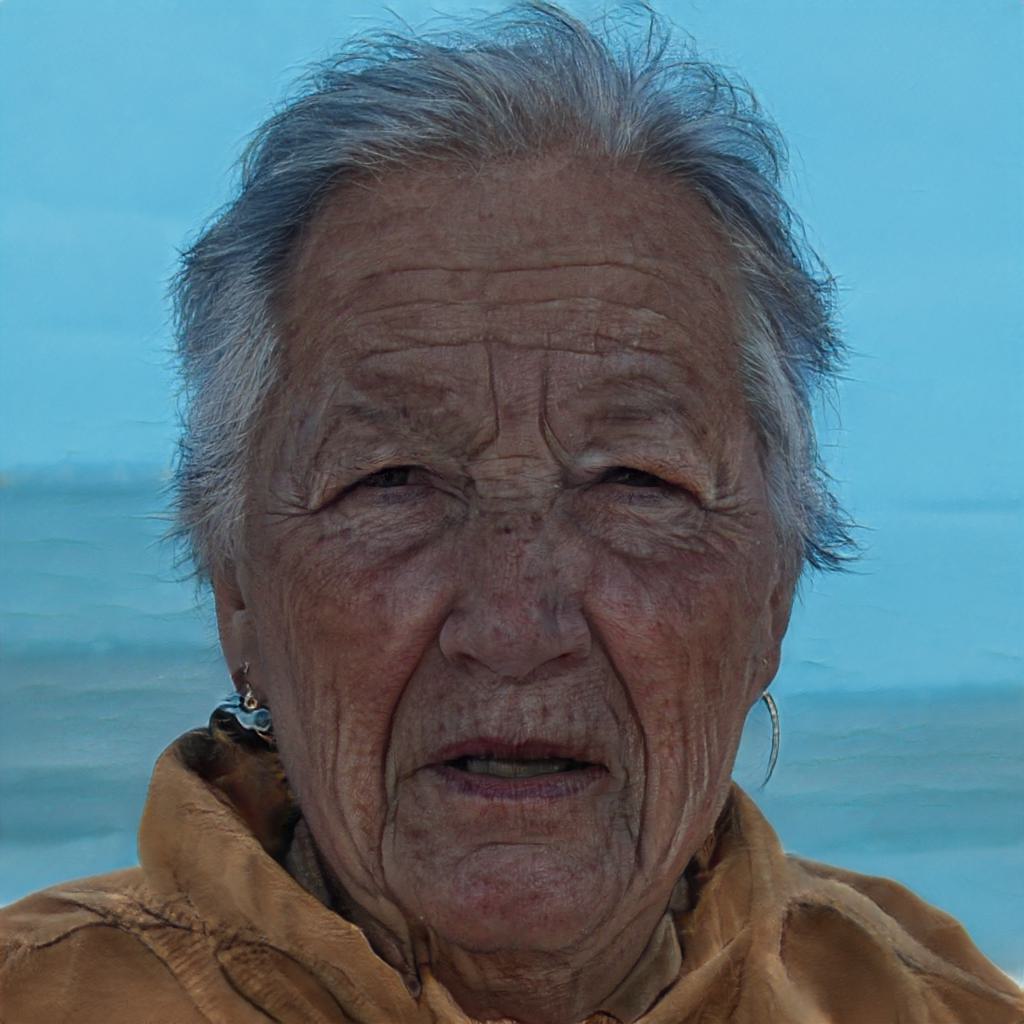} & \includegraphics[width=2.cm]{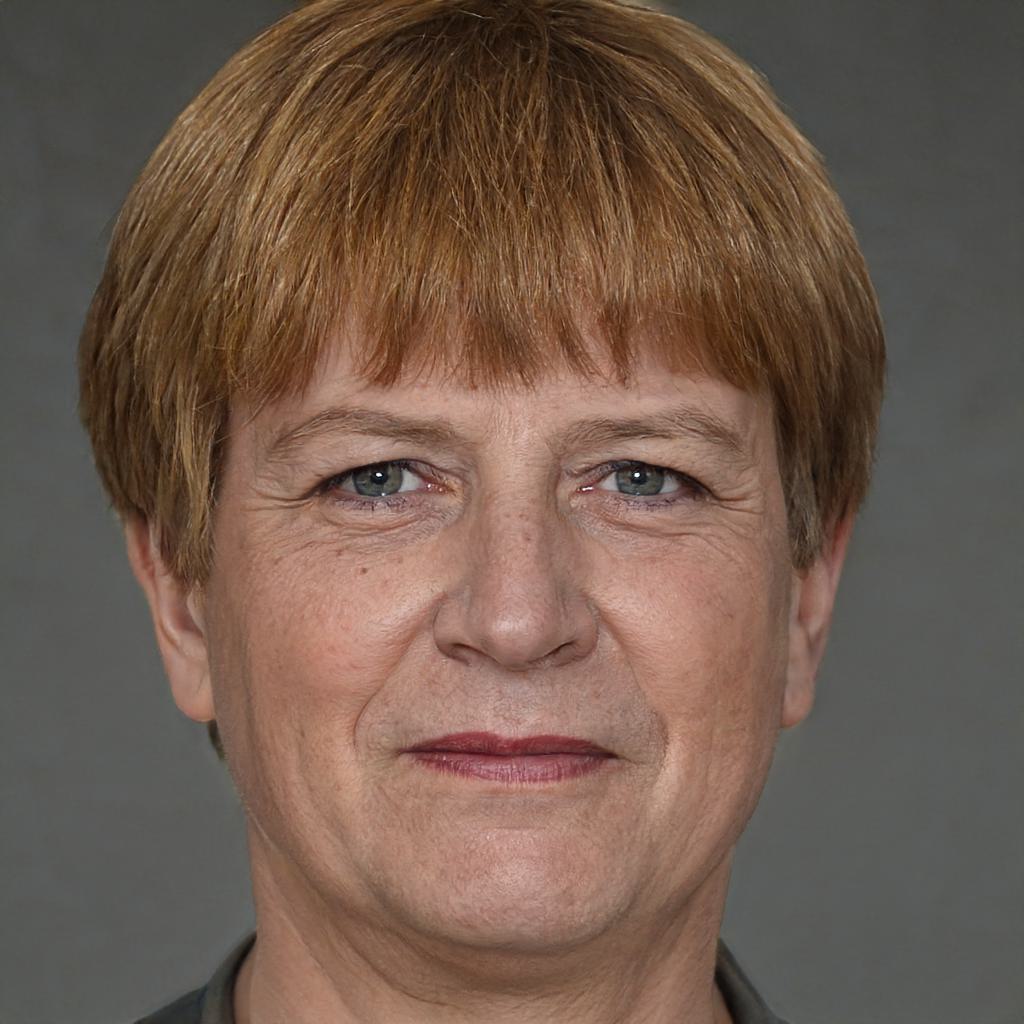} & \includegraphics[width=2.cm]{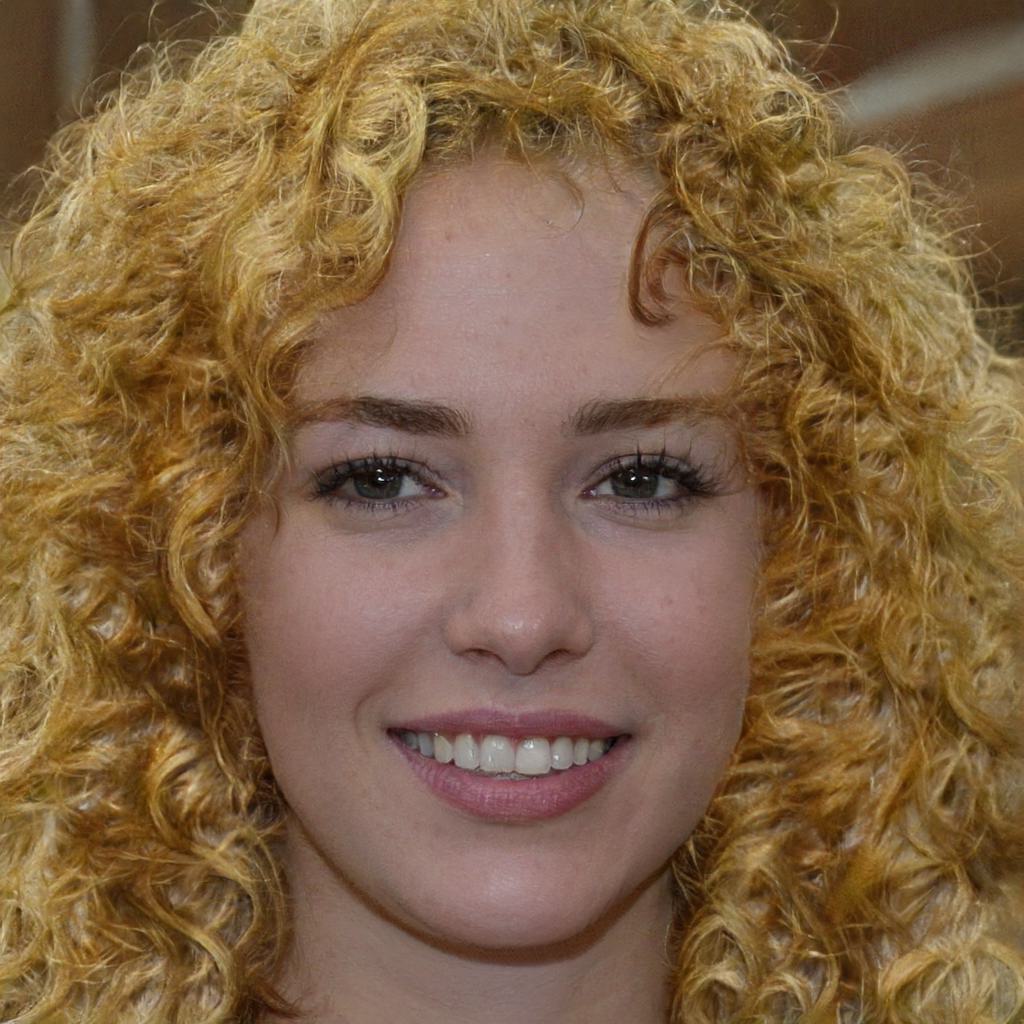} & \includegraphics[width=2.cm]{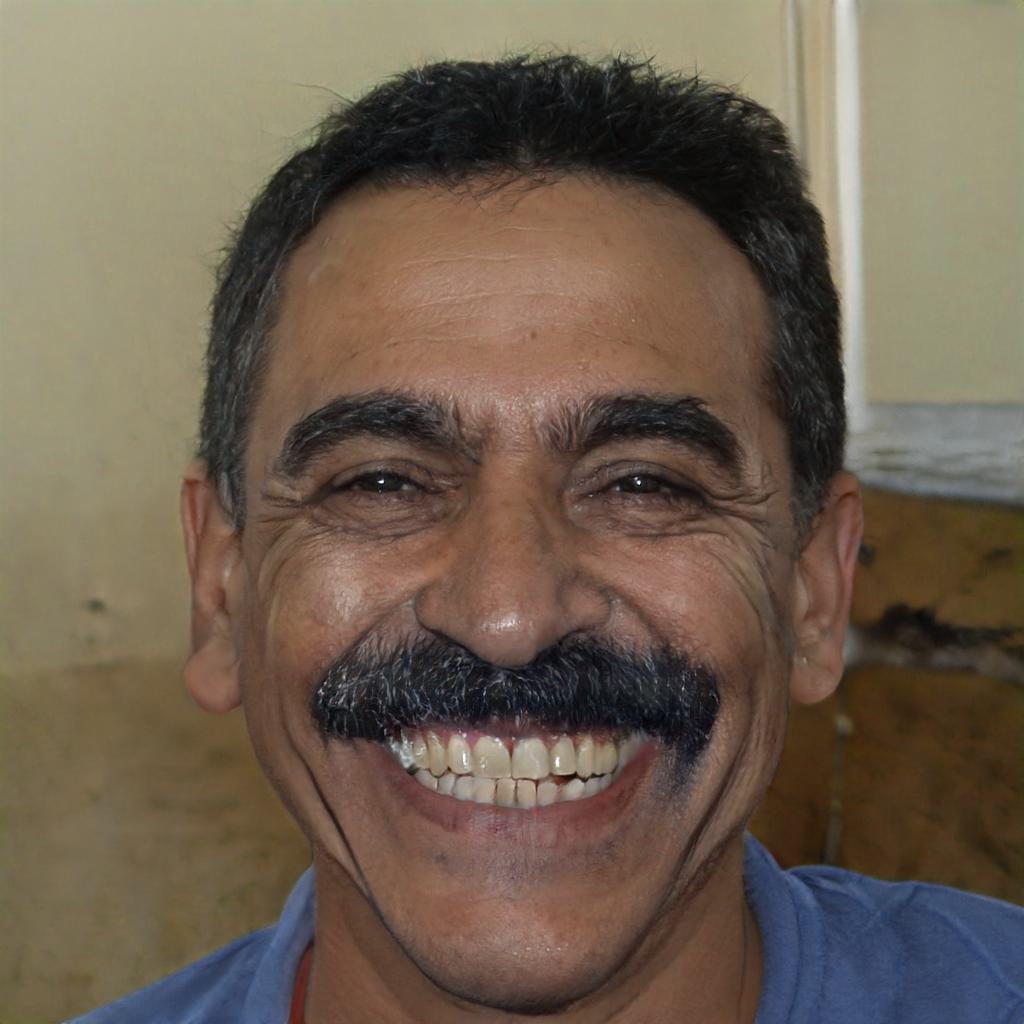} & \includegraphics[width=2.cm]{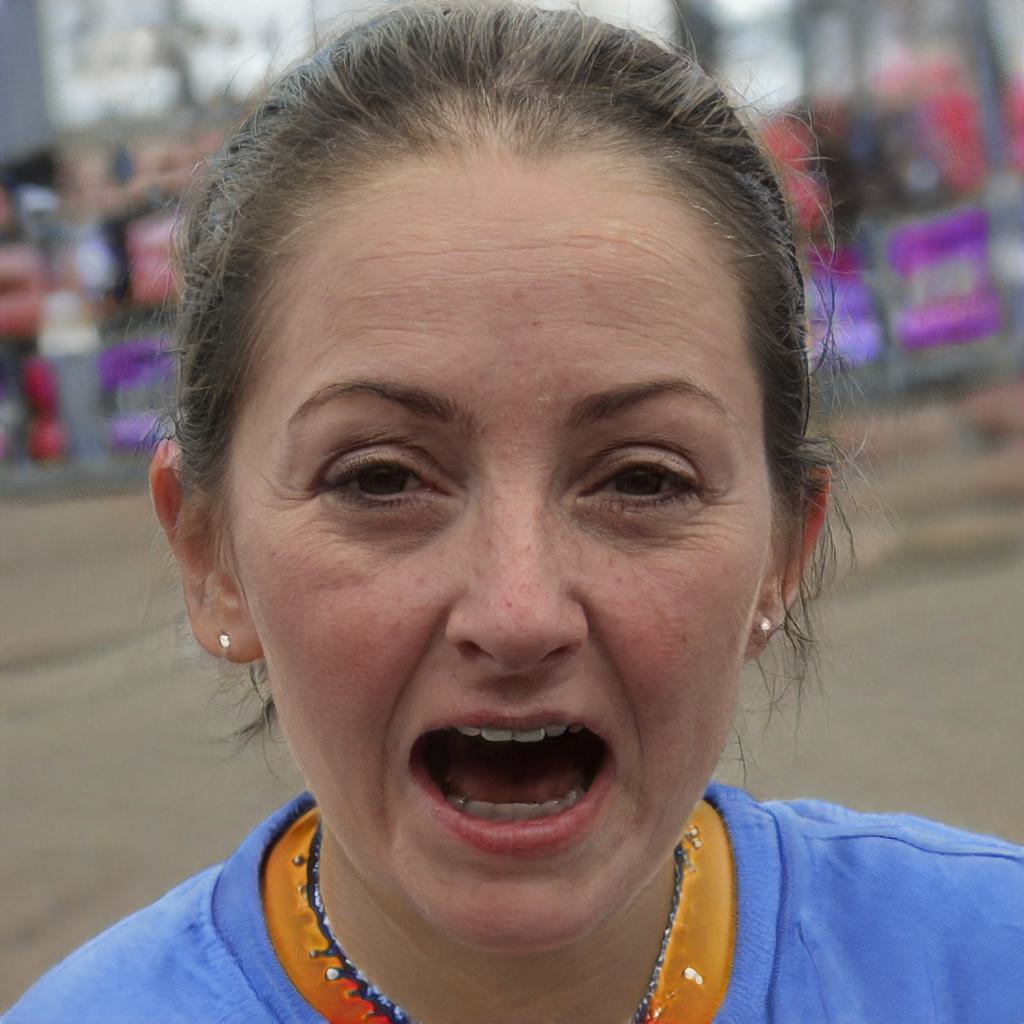} & \includegraphics[width=2.cm]{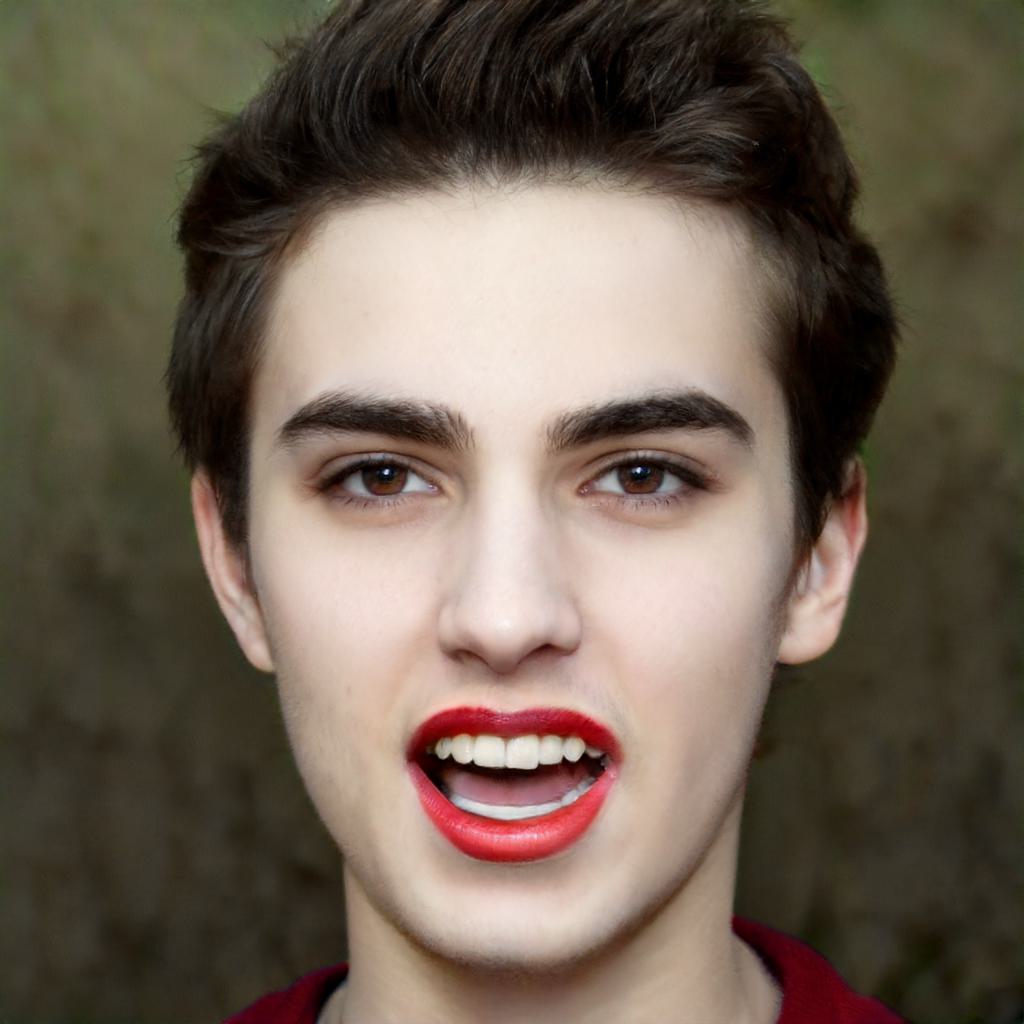} & \includegraphics[width=2.cm]{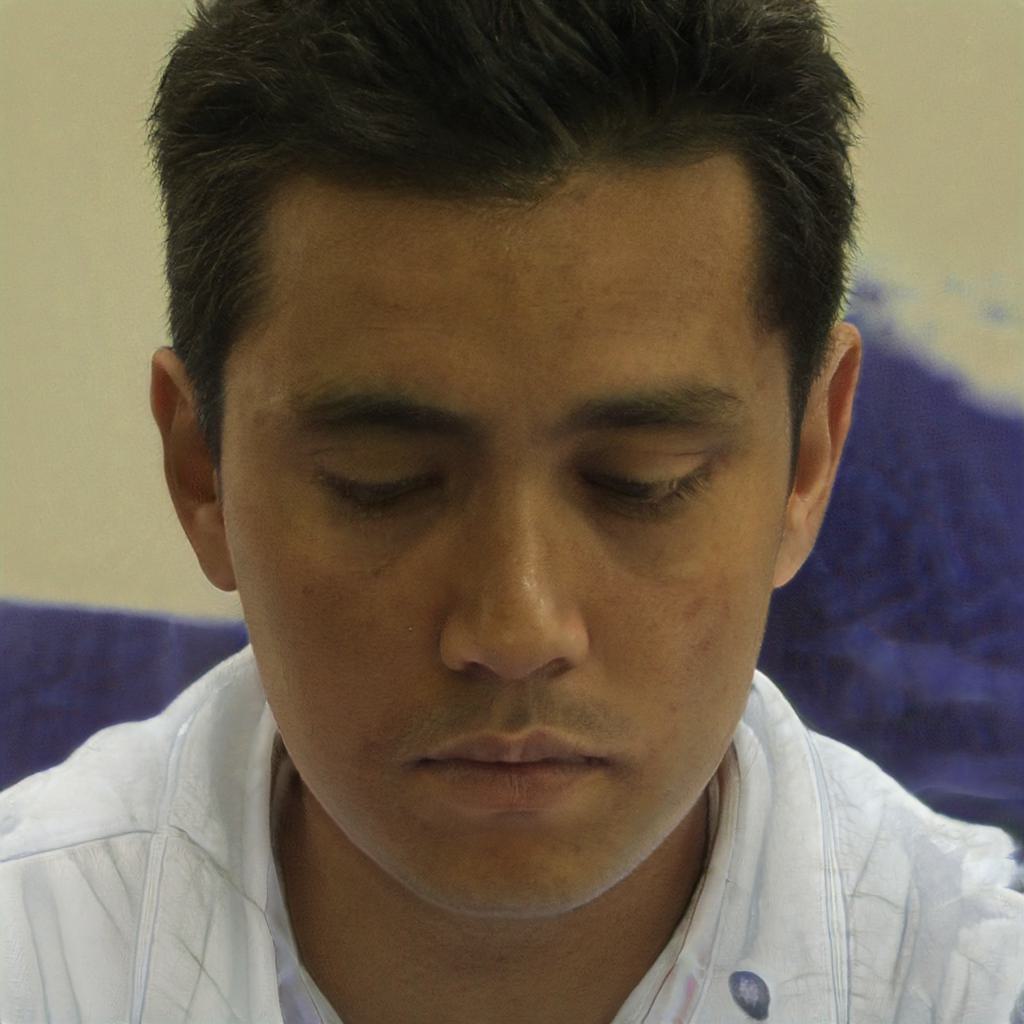}& \includegraphics[width=2.cm]{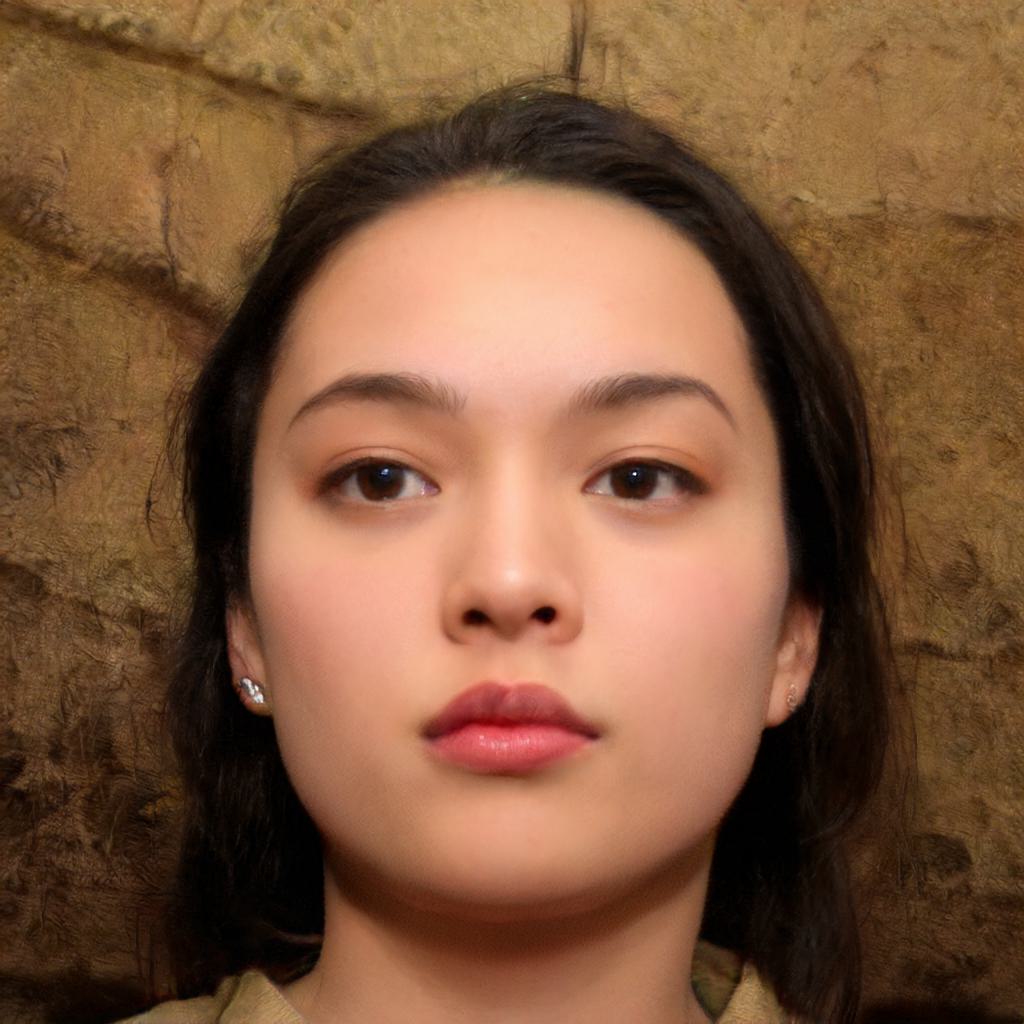}\\[-0.05cm]
  A photo of an old person by the beach & A portrait of Angela Merkel & A person with curly blonde hair & Middle aged man with a moustache and a happy expression & \hspace{0.035cm}The face of a person who just finished running a marathon & \hspace{0.035cm}A vampire's face & An office worker wearing a shirt looking stressed & \hspace{0.08cm}The Mona Lisa \\[-0.25cm]

\end{tabular}
\caption{Uncurated Samples from our TR0N:StyleGAN2+CLIP model.}
\label{fig:stylegan_faces_appendix}
\end{figure*}

\autoref{fig:stylegan_faces_appendix} shows uncurated samples from our TR0N:StyleGAN2+CLIP model. We use (randomly subsampled) out-of-distribution prompts from \citet{Pinkney2022clip2latent}.

% \subsection{TR0N:NVAE+CLIP samples}

\begin{figure*} [t!]
\centering
\fontsize{7.5}{9}
\selectfont
\begin{tabular}
{p{0.11\linewidth}p{0.11\linewidth}p{0.11\linewidth}p{0.11\linewidth}p{0.11\linewidth}p{0.11\linewidth}p{0.11\linewidth}}
   \includegraphics[width=2.25cm]{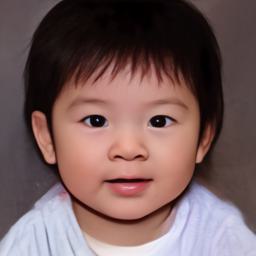} & \includegraphics[width=2.25cm]{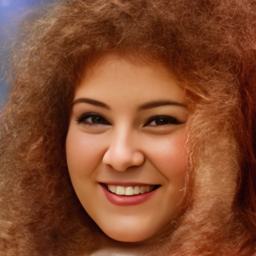} & \includegraphics[width=2.25cm]{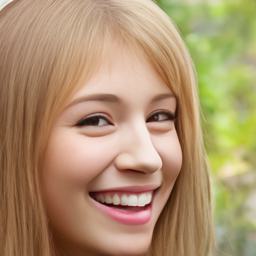} & \includegraphics[width=2.25cm]{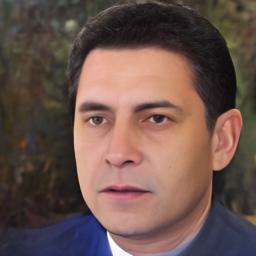} & \includegraphics[width=2.25cm]{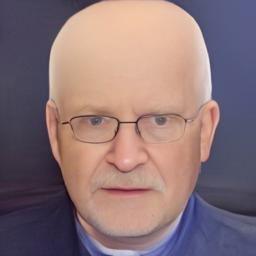} & \includegraphics[width=2.25cm]{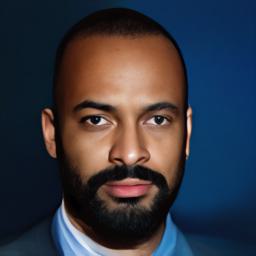} & \includegraphics[width=2.25cm]{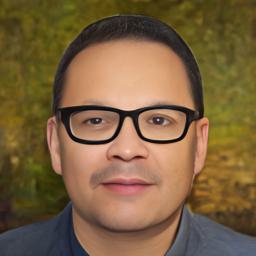}\\[-0.05cm]
   A young boy with black hair & A woman with curly hair & A young woman with a smile & \hspace{0.19cm}An angry person & \hspace{0.2cm}An old bald man & A man with a beard & A man with glasses \\[-0.25cm] 
\end{tabular}
\caption{Samples from our TR0N:NVAE+CLIP model.}
\label{fig:nvae_faces_appendix}
\end{figure*}

\autoref{fig:nvae_faces_appendix} shows samples from the TR0N:NVAE+CLIP. We do not compare this model against clip2latent since clip2latent only uses StyleGAN2, and the comparison would thus not be fair. \autoref{fig:nvae_faces_appendix} is thus meant to show that TR0N can also enable an NVAE model to be conditioned on text, once again highlighting the generality of TR0N.

\begin{figure*} [t!]
\newcommand{\centered}[1]{\begin{tabular}{c}  \\[-2cm] #1 \end{tabular}}
\newcommand{\centeredone}[1]{\begin{tabular}{c}  \\[-1.95cm] #1 \end{tabular}}
\newcommand{\centeredtwo}[1]{\begin{tabular}{c}  \\[-1.79cm] #1 \end{tabular}}
\newcommand{\centeredthree}[1]{\begin{tabular}{c}  \\[-2.1cm] #1 \end{tabular}}
\newcommand{\centeredfour}[1]{\begin{tabular}{c}  \\[-1.95cm] #1 \end{tabular}}
\renewcommand\fbox{\fcolorbox{black!15!red}{white}}
\centering
% \hspace*{-0.41cm}
\begin{tabular}
{ccc}
   \hspace{-0.1cm}\centered{{\setlength{\fboxsep}{0pt}\setlength{\fboxrule}{0.09cm}\fbox{\includegraphics[width=1.67cm]{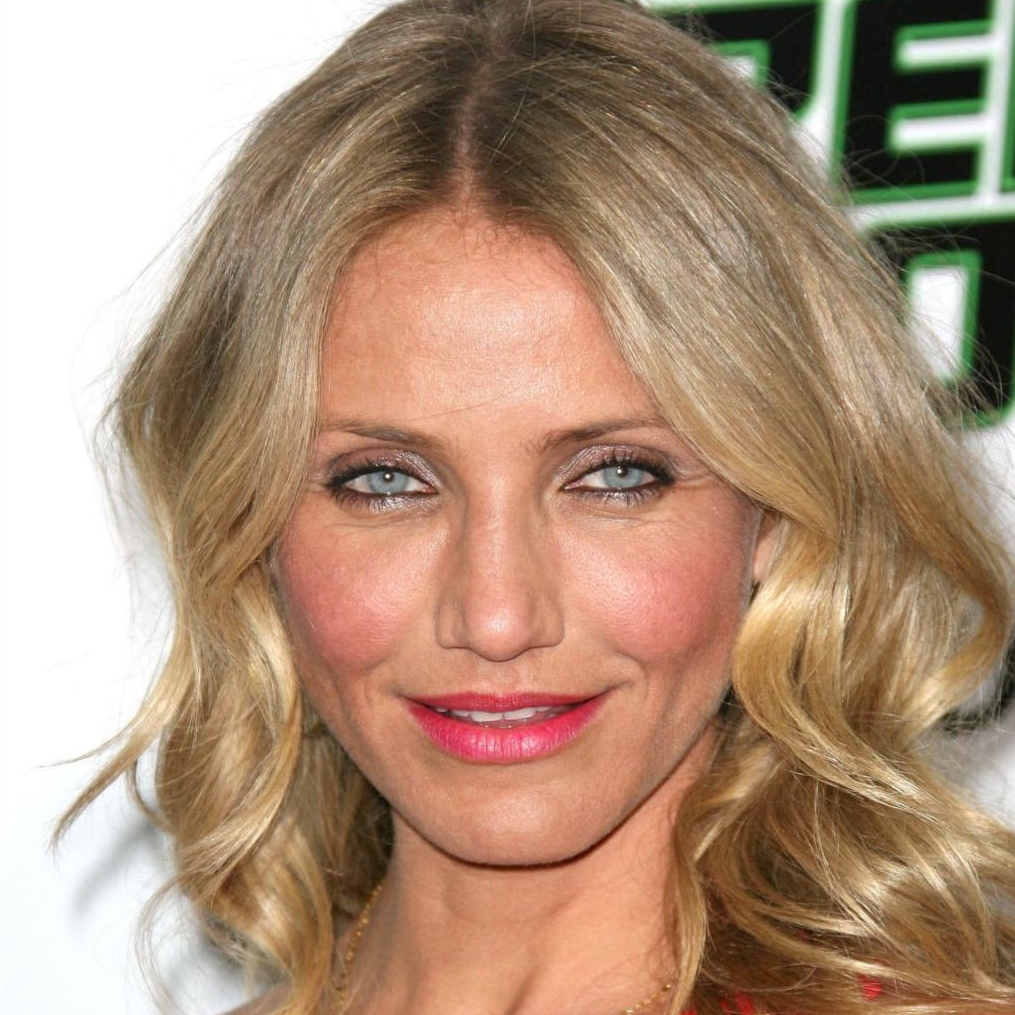}}}} & \hspace{-0.64cm}\includegraphics[width=1.85cm]{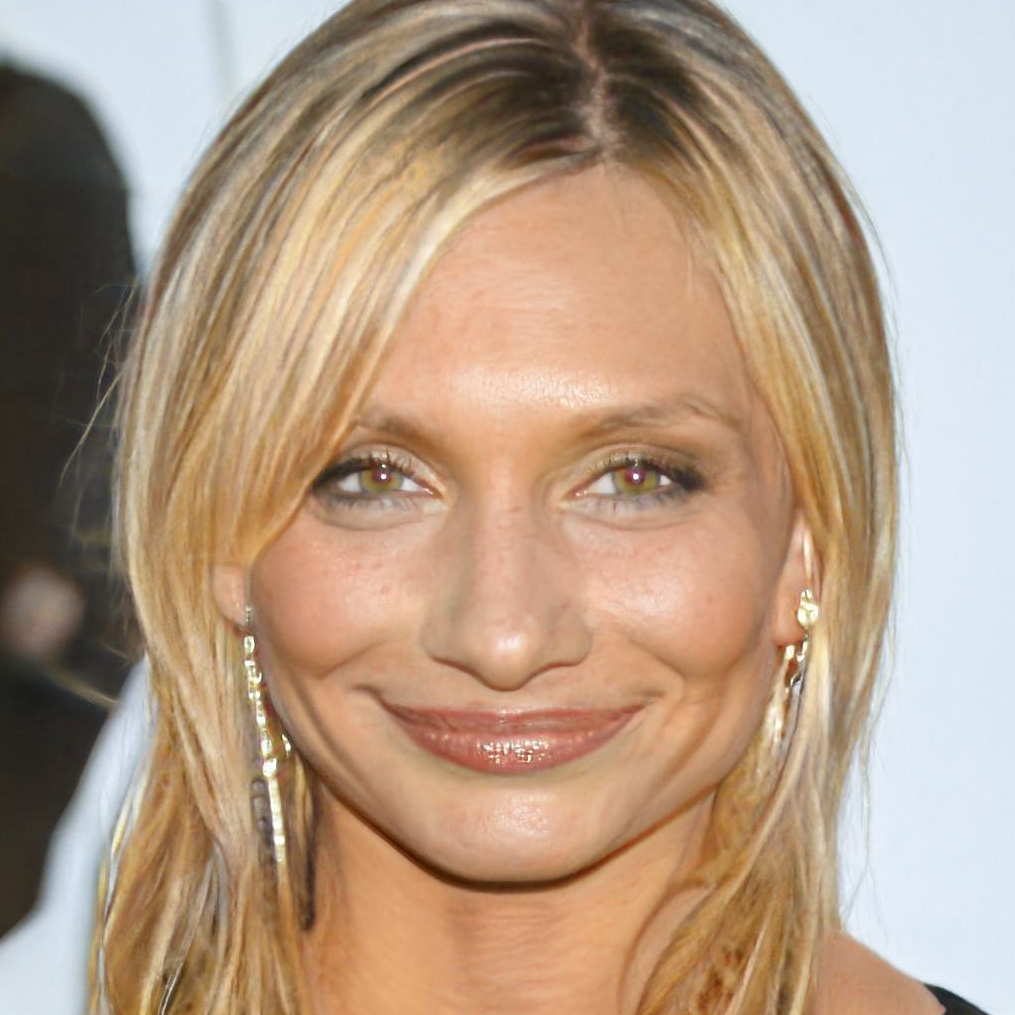} & \centeredone{\includegraphics[width=12cm]{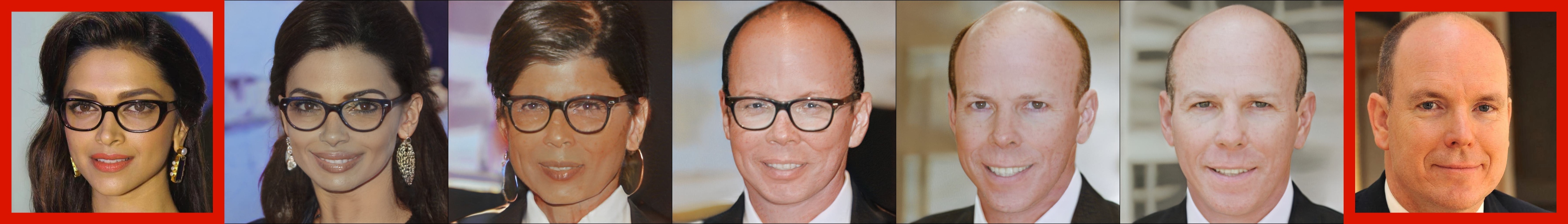}} \\[-0.12cm]
   \hspace{-0.1cm}\includegraphics[width=1.85cm]{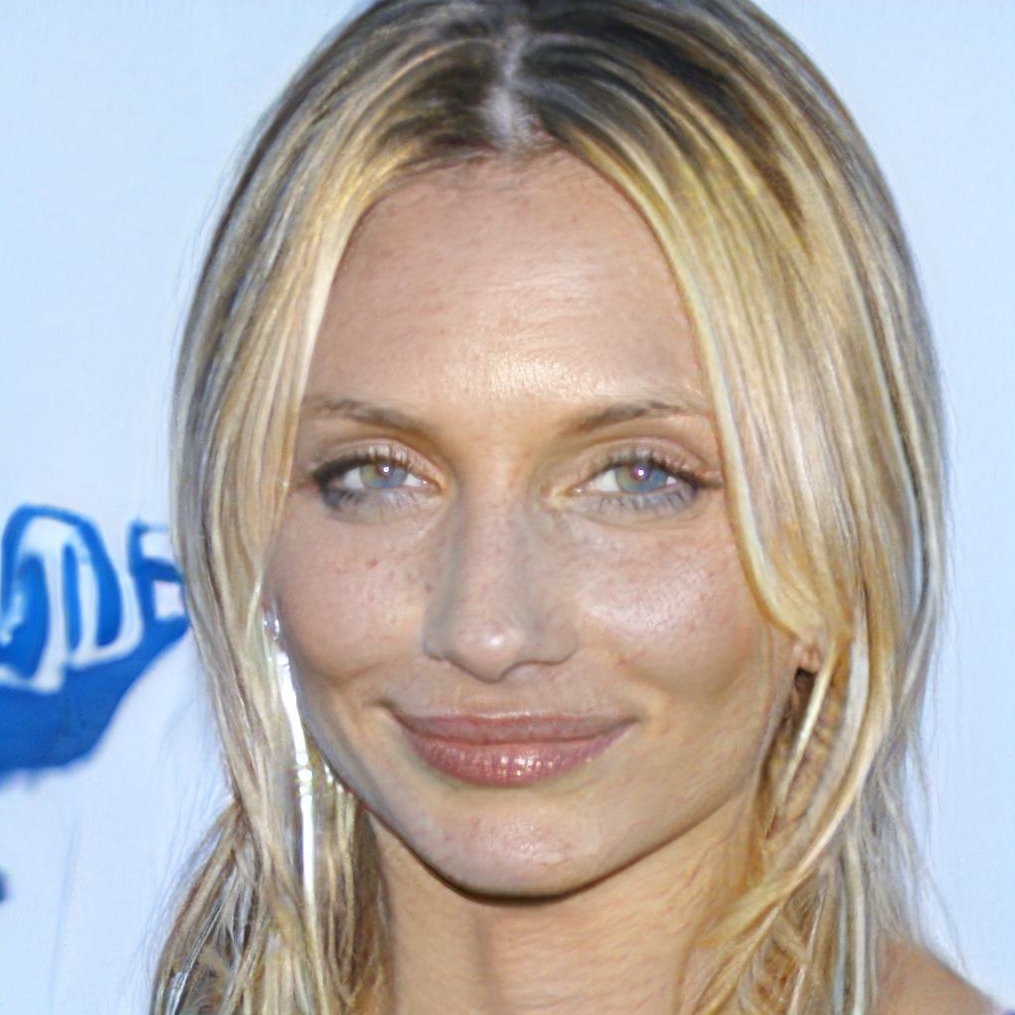} & \hspace{-0.64cm}\includegraphics[width=1.85cm]{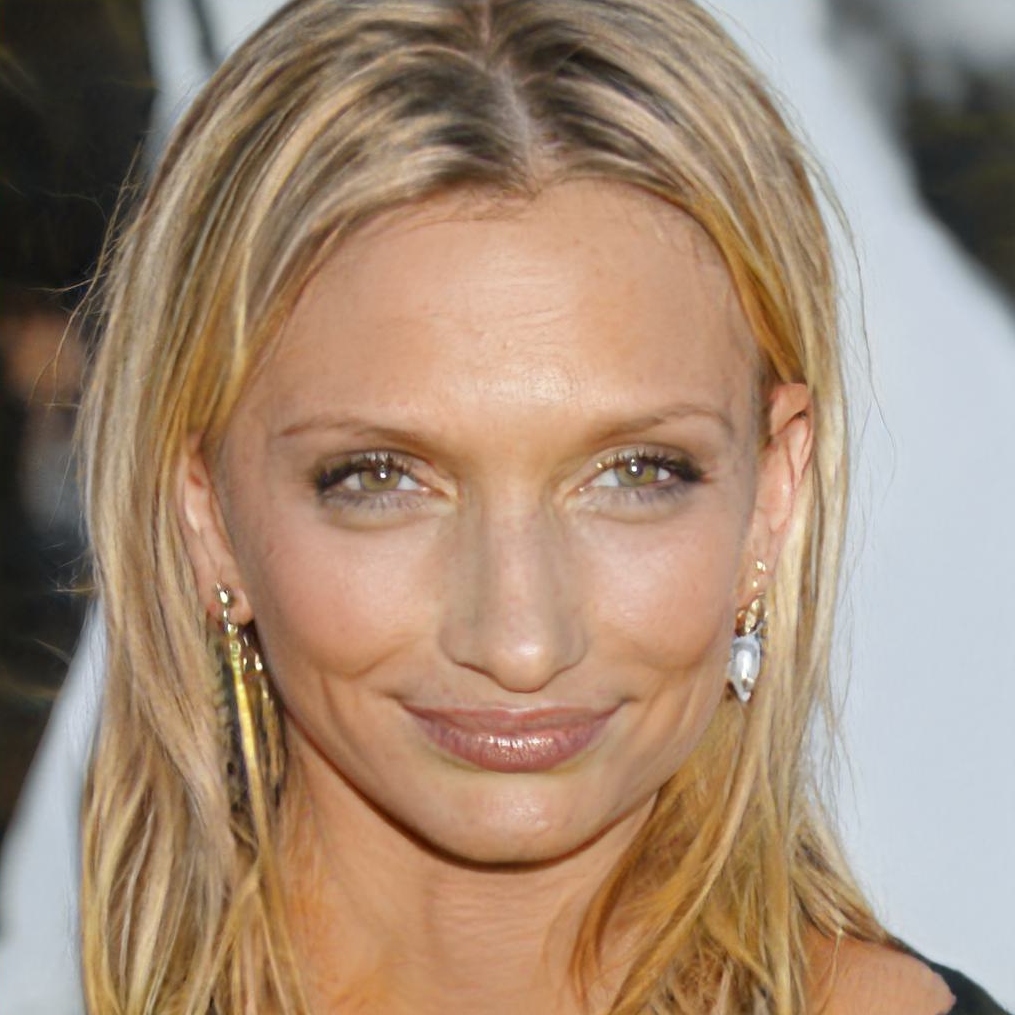} & \centeredtwo{\includegraphics[width=12cm]{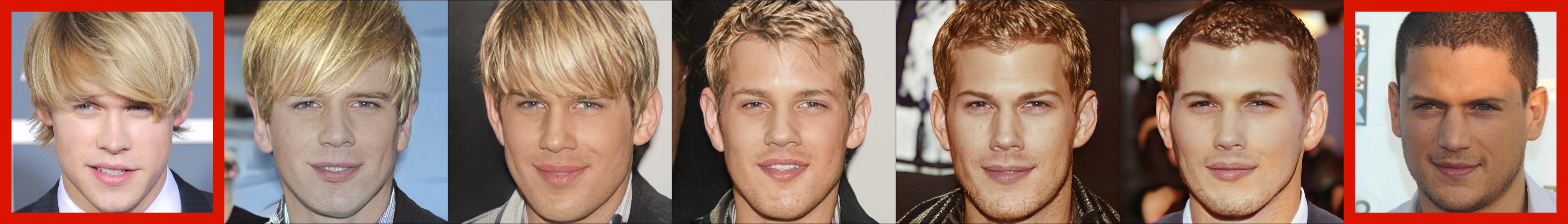}} \\[0.12cm]
   \hspace{-0.1cm}\centered{{\setlength{\fboxsep}{0pt}\setlength{\fboxrule}{0.09cm}\fbox{\includegraphics[width=1.67cm]{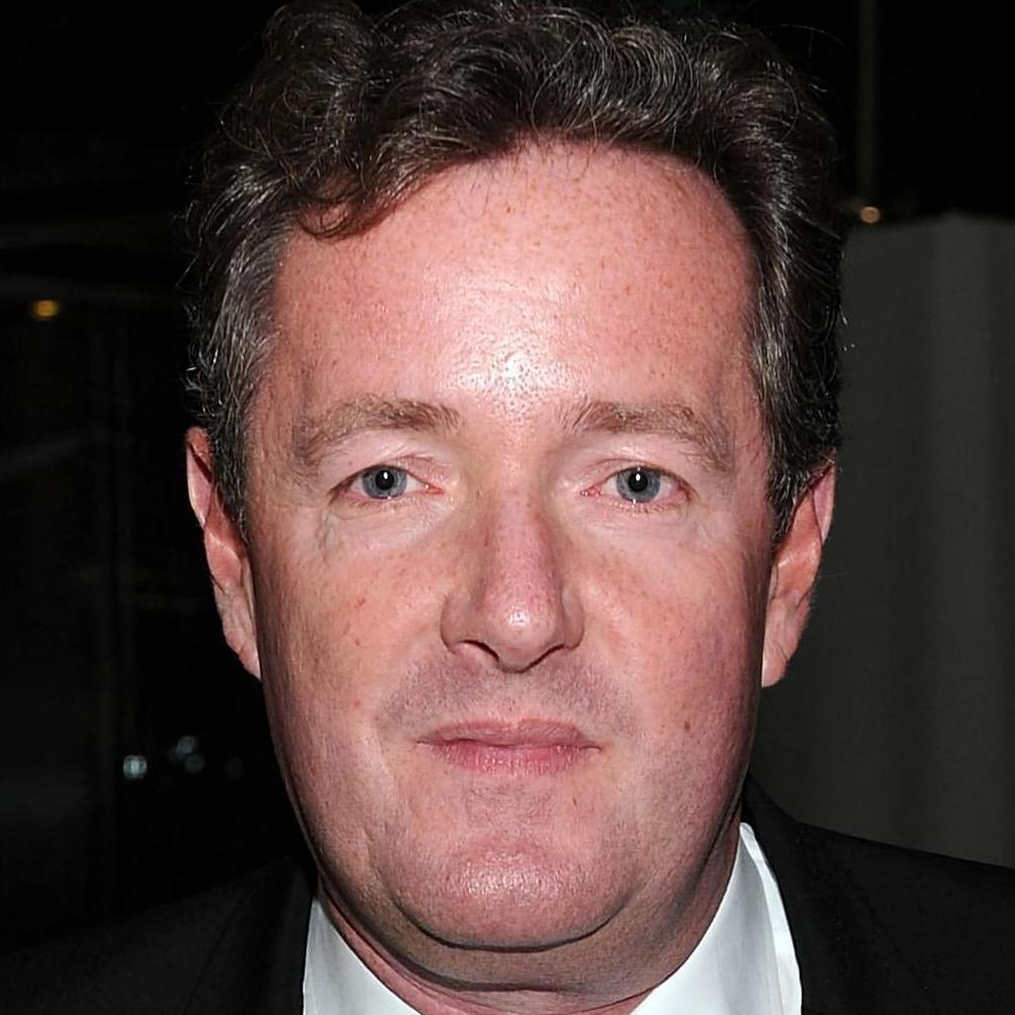}}}} & \hspace{-0.64cm}\includegraphics[width=1.85cm]{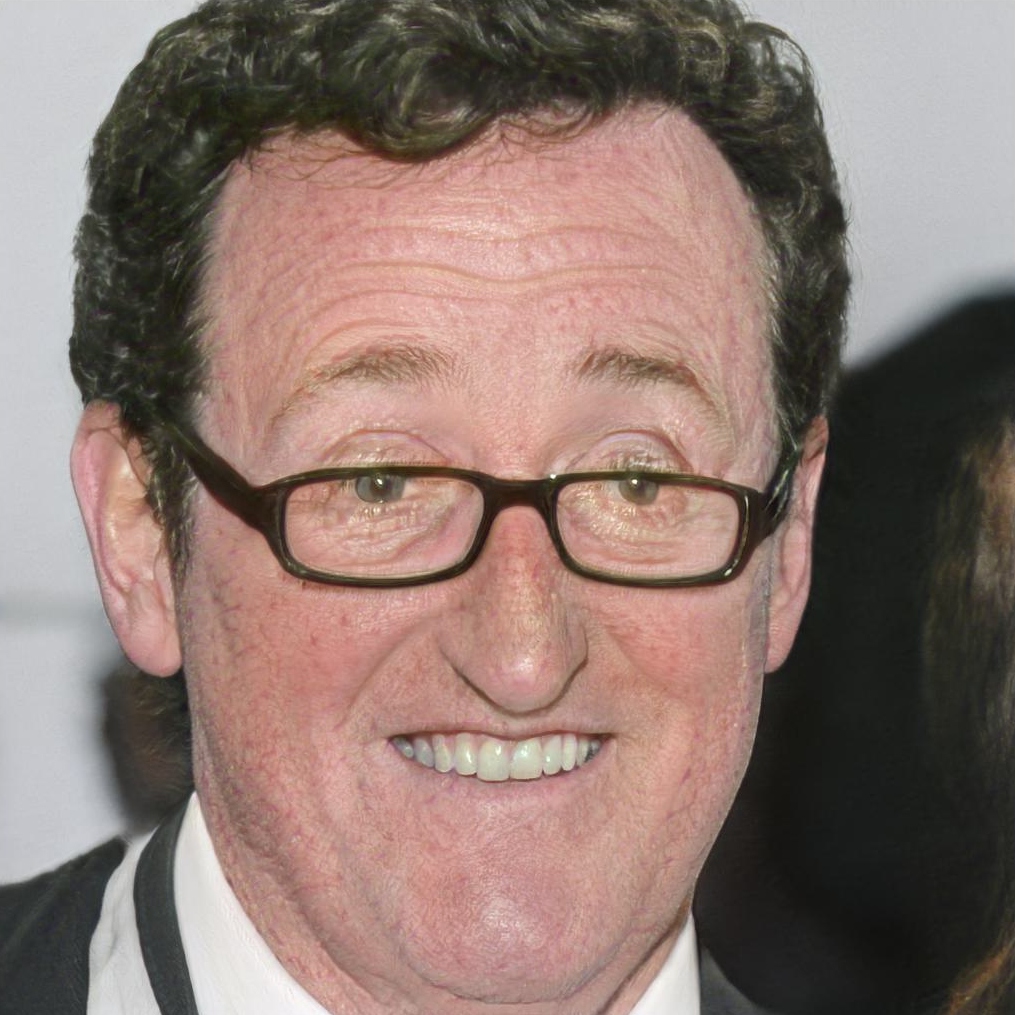} & \centeredthree{\includegraphics[width=12cm]{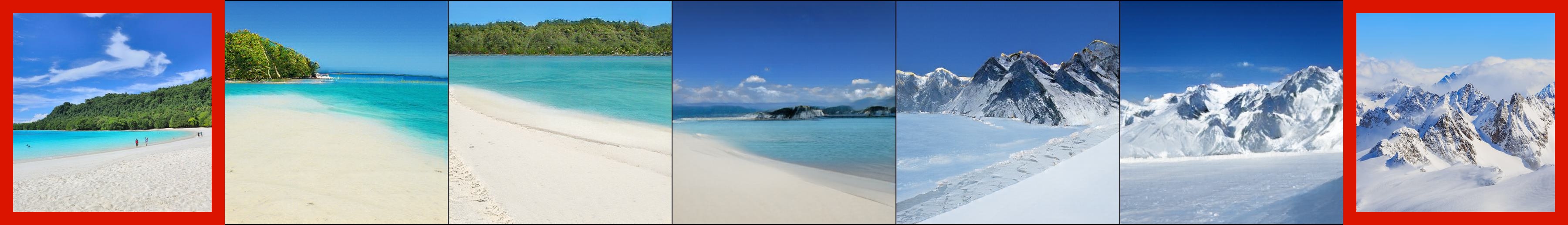}} \\[-0.12cm]
   \hspace{-0.1cm}\includegraphics[width=1.85cm]{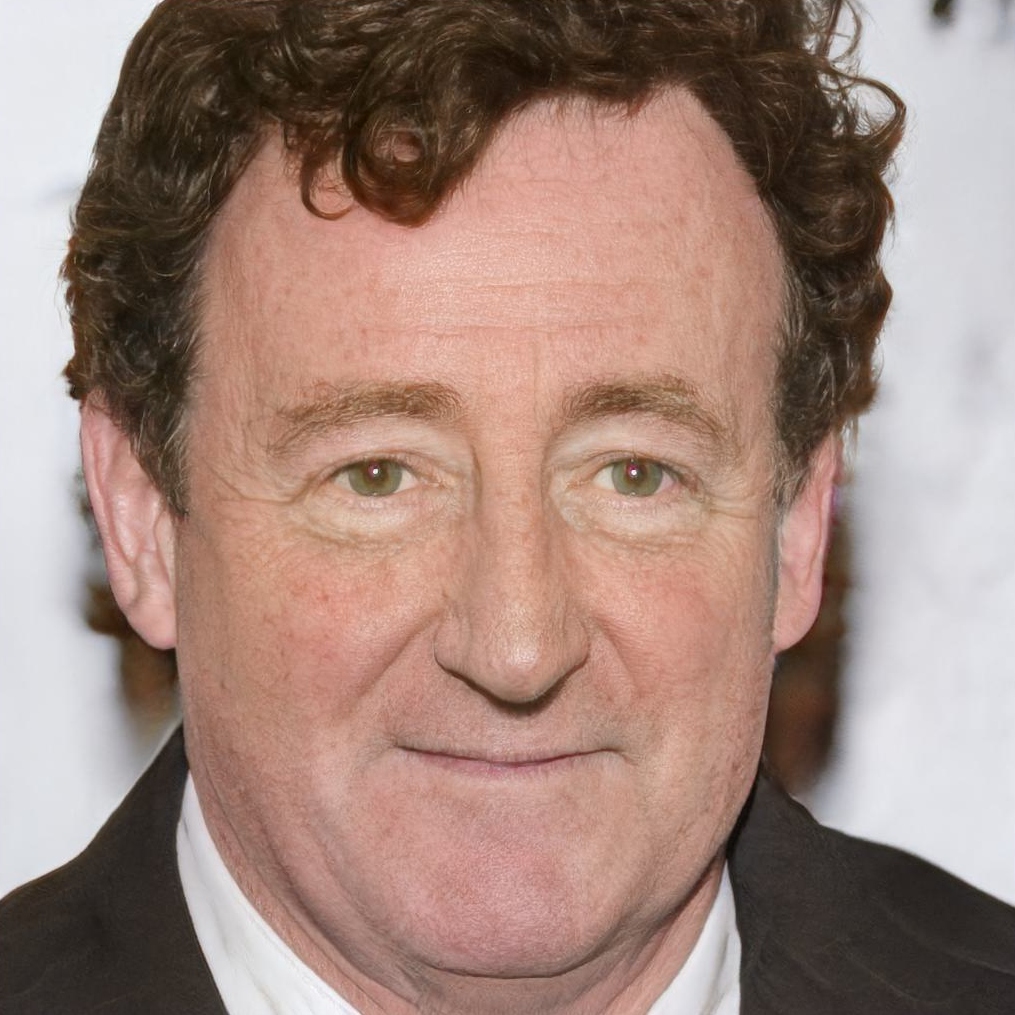} & \hspace{-0.64cm}\includegraphics[width=1.85cm]{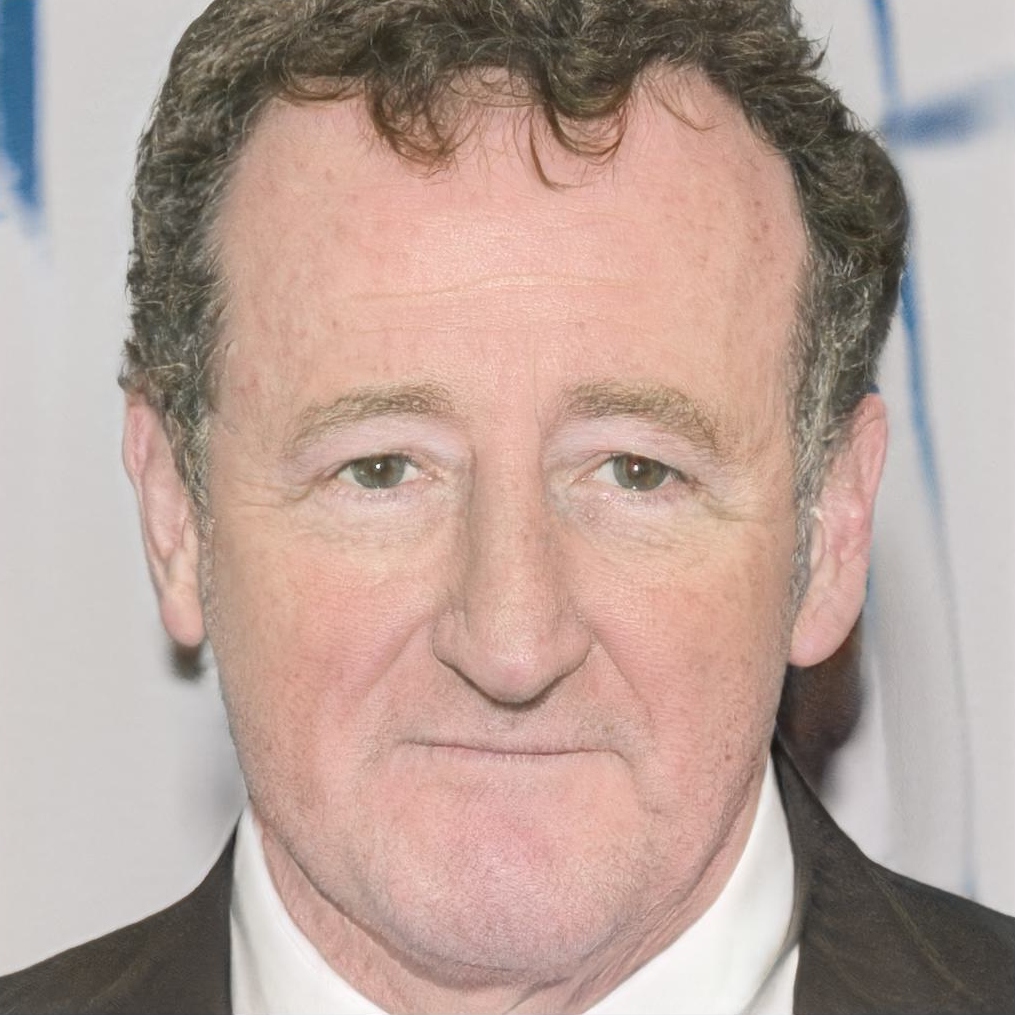} & \centeredfour{\includegraphics[width=12cm]{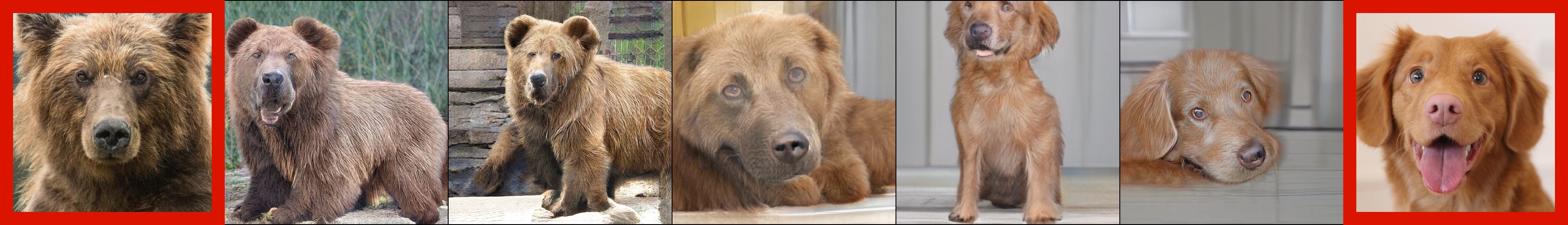}}
\end{tabular}
\caption{TR0N samples conditioning on image semantics with $G$ as a StyleGAN2 (\textbf{left panels}, $x'$ is highlighted in red); and interpolations (\textbf{right panels}, $x_1'$ and $x_2'$ are highlighted in red) with $G$ as a StyleGAN2 (\textbf{top right panels}), and with $G$ as a BigGAN (\textbf{bottom right panels}).}
\label{fig:image_semantics_app}
\vspace{128in}
\end{figure*}

\autoref{fig:image_semantics_app} shows the image semantics conditioning samples mentioned in \autoref{sec:semantics}. Despite both the StyleGAN2 and the BigGAN-based models having good performance at the first task (conditioning on a single image), we find that the former is better at interpolation. Note that TR0N carries out the interpolation on $\mathcal{C}_{\text{CLIP}}$ rather than $\mathcal{Z}$, which we hypothesize might be the reason that the BigGAN-based model is not as strong at interpolation.

% You can have as much text here as you want. The main body must be at most $8$ pages long.
% For the final version, one more page can be added.
% If you want, you can use an appendix like this one, even using the one-column format.
%%%%%%%%%%%%%%%%%%%%%%%%%%%%%%%%%%%%%%%%%%%%%%%%%%%%%%%%%%%%%%%%%%%%%%%%%%%%%%%
%%%%%%%%%%%%%%%%%%%%%%%%%%%%%%%%%%%%%%%%%%%%%%%%%%%%%%%%%%%%%%%%%%%%%%%%%%%%%%%

\end{document}